\def\year{2022}\relax
\documentclass[letterpaper]{article} %
\usepackage{aaai22}  %
\usepackage{times}  %
\usepackage{helvet}  %
\usepackage{courier}  %
\usepackage[hyphens]{url}  %
\usepackage{graphicx} %
\usepackage{natbib}  %
\usepackage{caption} %
\DeclareCaptionStyle{ruled}{labelfont=normalfont,labelsep=colon,strut=off} %
\usepackage{algorithm}
\usepackage{algorithmic}

\usepackage{newfloat}
\usepackage{listings}
\floatstyle{ruled}
\newfloat{listing}{tb}{lst}{}
\floatname{listing}{Listing}

\usepackage{amssymb}

\usepackage{amsmath,amsfonts,bm}

\def\eqref#1{equation~\ref{#1}}

\def\1{\bm{1}}

\def\vh{{\bm{h}}}

\DeclareMathAlphabet{\mathsfit}{\encodingdefault}{\sfdefault}{m}{sl}
\SetMathAlphabet{\mathsfit}{bold}{\encodingdefault}{\sfdefault}{bx}{n}

\newcommand{\E}{\mathbb{E}}

\newcommand{\Var}{\mathrm{Var}}

\def\T{{\mathsf{T}}}

\usepackage{amsmath}
\usepackage[utf8]{inputenc} %
\usepackage{url}            %
\usepackage{booktabs}       %
\usepackage{amsfonts}       %
\usepackage{nicefrac}       %
\usepackage{microtype}      %
\usepackage{xcolor}         %
\usepackage{graphicx}
\usepackage{amsthm}
\newtheorem{theorem}{Theorem}
\newtheorem{corollary}{Corollary}[theorem]

\usepackage{bbm}
\usepackage{cleveref}
\usepackage{todonotes}
\usepackage{mathtools}
\usepackage[inline]{enumitem}

\usepackage{caption}
\usepackage{subcaption}

\usepackage[rl]{abaisero}
\newcommand\realset{\mathbb{R}}
\DeclareMathOperator\Exp{\mathbb{E}}
\newcommand\hs{{\bm{h}}}
\newcommand\as{{\bm{a}}}
\newcommand\os{{\bm{o}}}
\newcommand\policies{{\bm{\pi}}}
\newcommand\vpolicies{V^\policies}
\newcommand\qpolicies{Q^\policies}
\newcommand\ppolicies{P^\policies}

\title{A Deeper Understanding of State-Based Critics \\in Multi-Agent Reinforcement Learning}
\author {
    Xueguang Lyu,
    Andrea Baisero,
    Yuchen Xiao, 
    Christopher Amato 
}
\affiliations {
    
    Northeastern University \\
    \{lu.xue, baisero.a, xiao.yuch, c.amato\}@northeastern.edu
}

\begin{document}

\maketitle

\begin{abstract}

Centralized Training for Decentralized Execution, where training is done in a centralized offline fashion, has become a popular solution paradigm in Multi-Agent Reinforcement Learning.
Many such methods take the form of actor-critic with state-based critics, since centralized training allows access to the true system state, which can be useful during training despite not being available at execution time.
State-based critics have become a common empirical choice, albeit one which has had limited theoretical justification or analysis.
In this paper, we show that state-based critics can introduce bias in the policy gradient estimates, potentially undermining the asymptotic guarantees of the algorithm.
We also show that, even if the state-based critics do not introduce any bias, they can still result in a larger gradient variance, contrary to the common intuition.
Finally, we show the effects of the theories in practice by comparing different forms of centralized critics on a wide range of common benchmarks, and detail how various environmental properties are related to the effectiveness of different types of critics.

\end{abstract}

\section{Introduction}
The rising popularity of Centralized Critics in Multi-Agent Reinforcement Learning (MARL) has led to the usage of state information becoming common practice.
The rationale behind state-based critics is straightforward---centralized critics train in a centralized offline manner and usually have access to the environment state, which appears desirable compared to local observations.
Because critics can be discarded after the training of actors, they do not hinder independent execution by each agent.
As a result, state-based critics have become a convenient and popular design decision for centralized training with actor-critic methods~\cite{foerster:aaai18, SQDDPG, LIIR, LICA, VDAC, du2021learning,schroeder2019multi, hideseek, DOP}.
However, state-based critics have received little formal analysis, making the state-based critic's benefits and downsides insufficiently understood in the field.
In this paper, we point out that there exists misconceptions in the field towards state-based critics.
Our primary contribution is to fill this knowledge gap by providing both intuition and analysis of theoretical properties of state-based critics, complimented with empirical findings and suggestions.

First, we show how the state-based critic is not entirely sound in theory.
We provide bias analysis to conclude that the state-based critic may incur unbounded bias on the policy gradient compared to the asymptotically unbiased history-based critic.
Second, even assuming that the state-based critic is unbiased, we analyze policy gradient variance and show that the policy gradient variance with the state-based critic cannot be less than that of the history-based critic.
We also give trivial and intuitive toy examples that highlight the essence of our arguments and provide an intuitive understanding of the theories.

We also compare the empirical performance of state-based and history-based critics.
Supported by a wide array of experiments, we also discuss the implications of our theories in practice.
We highlight particular circumstances regarding reactive policies, information gathering, and representation learning.
We demonstrate how the effectiveness of different types of critics depends on observation models of the tasks and highlight the deficiencies of popular benchmarks.
We also test an alternative critic that combines state and history information, which demonstrates reliability across tasks.
Through linking empirical results with theory, we establish a deeper understanding of where the state-based critics work and where they fall short.

\section{Related Work}

Centralized Training for Decentralized Execution (CTDE)~\cite{OliehoekSV08, foerster2016learning} %
has shown significant benefits for learning decentralized policies by addressing environmental non-stationarity that emerges in independent learning methods.
CTDE provides access to global information during training while having decentralized execution.
Learning decentralized actors with a centralized critic has thus arisen as a direct and widespread use of the CTDE framework.
There are many variants of the information used by the centralized critic, such as the environment state, joint observation, joint action-observation history, or even mixed combinations. 

COMA~\cite{foerster:aaai18}, as the first deep multi-agent actor-critic-based algorithm with CTDE, learns a centralized critic to provide joint Q-value estimations with explicit access to ground truth state information.
MADDPG~\cite{lowe2017multi}, the other pioneer, also uses centralized critics to update decentralized actors. In MADDPG, each critic accesses the joint-observation, which implicitly results in it being a state-based critic as the considered multi-agent domains are fully observable.
We include a deeper dive into the implicit assumptions and the scope of the work of COMA and MADDPG in~\Cref{on-assumptions-of-previous-works}.

The impressive performance achieved by COMA in SMAC and MADDPG in OpenAI Particle environments caused many future methods to use state-based critics without further study.
For example, numerous other popular works such as SQDDPG~\cite{SQDDPG}, LIIR~\cite{LIIR}, LICA~\cite{LICA},VDAC-mix~\cite{VDAC}, DOP~\cite{DOP} and MACKRL~\cite{schroeder2019multi} also utilize state-based centralized critics. 
Centralized state-based critic is also used in larger scale environments for emergence tool use~\cite{hideseek}.

The decision of whether critics condition on states or observations is often considered not of algorithmic importance; thereby, recent state-of-the-art methods treat state-based critics as an engineering decision rather than an algorithmic design (e.g., in DOP~\cite{DOP} which focuses on factorization).
However, it is unclear whether state-based critics are strictly beneficial since there is little theoretical analysis about how a state-based centralized critic impacts decentralized policy optimization.
Ablation comparisons over different centralized critic designs are necessary but missing in the current literature.
In this paper, we fill this gap by providing a theoretical and practical analysis of state-based critics.

\section{Preliminaries}

\subsection{Dec-POMDPs}
Decentralized partially observable Markov decision processes (Dec-POMDPs)~\cite{Oliehoek} are multi-agent cooperative sequential decision making problems.  A Dec-POMDP is composed of
\begin{enumerate*}[label={}]
    \item a set of agents $\mathcal{I}$,
    \item a state space $\mathcal{S}$, with initial state $s_0\in\mathcal{S}$,
    \item a joint action space $\mathbf{A} \doteq \bigtimes_{i\in\mathcal{I}}\mathcal{A}_i$, one per agent,
    \item a joint observation space $\mathbf{\Omega} \doteq \bigtimes_{i\in\mathcal{I}}\Omega_i$, one per agent,
    \item a state transition function $\mathcal{T}\colon \mathcal{S} \times \mathbf{A} \to \Delta\mathcal{S}$,
    \item a joint observation function $\mathcal{O}\colon \mathcal{S} \times \mathbf{A} \rightarrow \Delta\mathbf{\Omega}$,
    \item a joint reward function $\mathcal{R}\colon \mathcal{S} \times \mathbf{A} \times \mathcal{S} \to \mathbb{R}$, and
    \item a discount factor $\gamma\in[0, 1]$.
\end{enumerate*}

Control in Dec-POMDPs is performed by a set of policies $\policies = \langle \pi_1, \ldots, \pi_{|\mathcal{I}|}\rangle$, one per agent, each representing a (possibly stochastic) mapping from the agent's past observations to its next action, $\pi_i\colon \mathcal{H}_i\to\Delta\mathcal{A}_i$.
At each timestep $t$, a joint action $\as = \langle a_{1,t}, ..., a_{|\mathcal{I}|,t} \rangle$ is taken by the agents, each using their own individual history.  As feedback from the system, a scalar reward  $r_t = \mathcal{R}(\mathbf{s}_t,\bm{a}_t, \mathbf{s}_{t+1})$ is shared by all agents, and each agent receives a local observation $\langle o_{1,t},\dots,o_{|\mathcal{I}|,t} \rangle \sim \mathcal{O}(s_t,\bm{a}_t)$.
The collective objective of all agents is that of maximizing the expected performance $J\doteq \Exp\left[ G_0 \right]$, where $G_t \doteq \sum_k \gamma^k r_{t+k}$ are the discounted sum of future rewards, also called returns.

We focus on finite horizon problems for theoretical analysis;  note, however, that the finite horizon can be arbitrarily large enough to approximate episodic and infinite-horizon problems as well.
We will consider value functions over histories, states and history-state pairs.
We first introduce the \textit{history-state value} functions~\cite{bono2018cooperative} $Q^\pi(\hs,s,\as)$, it is later served as a link between state values and history values; it is defined as the expected return given the agents being in history $\hs$ and state $s$ and taking action $\as$:
\begin{align}
    \qpolicies(\hs, s, \as) &= \Exp\left[ G \mid  \hs, s,\as \right] \,. \label{eq:vhs} 
\end{align}
\noindent The history and state values are related to the history-state values via marginalization over the conditional on-policy distributions $\rho(s\mid \hs)$ and $\rho(\hs\mid s)$, respectively~\cite{Sutton:1999} (see \Cref{appendix:on-policy-dist}):
\begin{align}
    \qpolicies(\hs, \as) &= \mathop{\E}_{s\sim \rho(s\mid \hs)}\left[ G \mid \hs, s, \as \right] = \mathop{\E}_{s\sim \rho(s\mid \hs)}\left[ \qpolicies(\hs, s, \as) \right] \,, \label{eq:vh} \\
    \qpolicies(s, \as) &= \mathop{\E}_{\vh \sim \rho(\hs\mid s)}\left[ G \mid \hs, s, \as \right] = \mathop{\E}_{\hs\sim \rho(\hs\mid s)}\left[ \qpolicies(\hs, s, \as) \right] \,. \label{eq:vs}
\end{align}

\subsection{Multi-Agent Actor Critic Methods}
\label{subsec:multi_agent_actor_critic_methods}

Actor Critic (AC)~\cite{konda2000actor,sutton2000policy} is a popular Policy Gradient (PG) method which involves the training of policy and critic models;  we consider the centralized training case~\cite{lowe2017multi,bono2018cooperative,lyu2021contrasting}, where there is one policy model per agent (each separately parameterized by $\theta_i$), and a single centralized critic model (parameterized by $\phi$). 
We will often omit the model parameterization, when implicitly clear from context.
To distinguish the critic models from the value functions that they are trained to model, we will denote them as $\vmodel$. 
The policy gradient theorem states that the policy gradient follows the expected returns provided by the joint history values $\qpolicies(\hs, \as)$:
\begin{equation}
    \nabla_i J_\hs = \Exp_{\hs\sim\rho(\hs), \as\sim\policies(\hs)} \left[ \qpolicies(\hs, \as) \nabla_{\theta_i}\log\pi_i(a_i; h_i) \right] \,. \label{eq:gradient:qh}
\end{equation}
Similarly, COMA~\cite{foerster:aaai18} and MADDPG~\cite{lowe2017multi} introduced policy gradient variants which employed state values $\qpolicies(s, \as)$:
\begin{equation}
    \nabla_i J_s = \mathop{\E}_{\hs, \as} \left[ \mathop{\E}_{s\sim \rho(s|\hs)} \qpolicies(s, \as) \nabla_{\theta_i}\log\pi_i(a_i; h_i) \right] \,. \label{eq:gradient:qs}
\end{equation}

In either case, the values $\qpolicies$ can be estimated in a number of ways;  in actor critic, it is common to use one-step returns and the critic model to estimate $\qpolicies(\hs, \as)$ as $r + \gamma\vmodel(\hs\as\os)$, and $\qpolicies(s, \as)$ as $r + \gamma\vmodel(s')$.  In advantage actor critic, the critic model is further used as baseline for variance reduction, resulting in the following estimates:
\begin{align}
    \qpolicies(\hs, \as) - \vpolicies(\hs) &\approx r + \gamma\vmodel(\hs\as\os) - \vmodel(\hs) \,, \\
    \qpolicies(s, \as) - \vpolicies(s) &\approx r + \gamma\vmodel(s') - \vmodel(s) \,.
\end{align}

\section{Bias and Variance of State Values and Gradients}\label{sec:bias_and_variance}

In this section, we analyze the bias and variance resulting from the use of state value functions $\qpolicies(s,\as)$.  
Without loss of generality, we consider single-sample Monte Carlo estimates of \Cref{eq:gradient:qh,eq:gradient:qs},
\begin{align}
    \hat\nabla_i J_\hs &= \qpolicies(\hs, \as) \nabla\log\pi_i(a_i; h_i) \,, \label{eq:gradient:qh:mc} \\
    \hat\nabla_i J_s &= \qpolicies(s, \as) \nabla\log\pi_i(a_i; h_i) \,, \label{eq:gradient:qs:mc}
\end{align}
\noindent where $\hs,s\sim\rho(\hs,s)$ and $\as\sim\policies(\hs)$.
In \Cref{sec:bias}, we show that the state-based gradient estimate $\hat\nabla_i J_s$ can be biased.  In \Cref{sec:variance}, we show that even if the state-based gradient estimate $\hat\nabla_i J_s$ is unbiased, it will have a variance equal to or higher than that of $\hat\nabla_i J_\hs$.

In a related single-agent work~\cite{baisero2021unbiased}, state values are found to be undefined due to the potential non-existent of history distributions in infinite-horizon case.
Our analysis, on the other hand, assumes finite-horizon, so that history distributions and state values can be properly defined; and we show how even when the state values are properly defined, they are still not unbiased.
In addition to bias, we will also provide comparisons on gradient variance.
Note that the Q values for our analysis is the analytical return values, not the results of learned models.
Representation learning is a separate problem discussed in the experimental section.

\subsection{Bias Analysis}\label{sec:bias}

We begin by noting that the gradient given by the history-based critic $\hat\nabla_i J_\hs$ (\Cref{eq:gradient:qh}) has already been proven to be unbiased~\cite{foerster2016learning,lyu2021contrasting} due to the joint histories are Markov states for the history-MDP~\cite{Oliehoek} in which the policy gradient theorem holds~\cite{sutton2000policy}.
On the other hand, the gradient given by the state-based critic is:
\begin{align}
    \nabla_i J_s &= \mathop{\E}_{\hs,s\sim\rho(\hs,s),\as\sim\policies(\hs)}\left[ \qpolicies(s, \as) \nabla\log\policy_i(a_i; h_i) \right] \nonumber \\
    &=\mathop{\E}_{\hs,\as}\left[ \Exp_{s\sim\rho(s\mid \hs)}\left[ \qpolicies(s, \as) \right] \nabla\log\policy_i(a_i; h_i) \right] .
\end{align}
Together with history-based gradient~(\Cref{eq:gradient:qh}), we see the state-value-based-gradient $\hat\nabla_i J_s$ is also unbiased iff
\begin{align}
    \qpolicies(\hs,\as) = \Exp_{s\sim\rho(s\mid \hs)}\left[ \qpolicies(s,\as) \right] .
    \label{eq:bias}
\end{align}

Therefore, the analysis of the bias of $\hat\nabla_i J_s$ can be performed indirectly through $\qpolicies(s,\as)$. 
We adopt the methodology employed by prior work for the single-agent control case~\cite{baisero2021unbiased}, and will treat $\qpolicies(s,\as)$ as an estimator of $\qpolicies(\hs,\as)$.  Consequently, if $\qpolicies(s,\as)$ is unbiased, then $\hat\nabla_i J_s$ is also unbiased.

Consider the tabular form of the history-state function under a specific action $\as$, $\qpolicies_\as \in \realset^{|\bm{\sset}| \times |\bm{\hset}|}$, such that $\qpolicies_{\as,ij} = \qpolicies(\hs, s, \as)$ where $i$ and $j$ are the indices respectively corresponding to the joint history $\hs$ and a state $s$ (we will use this convention for the remainder of this section);  if both the state and observation spaces are finite, $\qpolicies_\as$ will be a finite matrix.
Also consider the tabular form of the normalized discounted visitations $\ppolicies\in[0, 1]^{|\bm{\sset}| \times |\bm{\hset}|}$, such that $\ppolicies_{ij} = \rho(\hs, s)$.
Note that recovering $\rho(\hs\mid s)$ and $\rho(s\mid \hs)$ from $\ppolicies$ requires renormalizing the values in a specific row or column,
\begin{align}
    \rho(\hs\mid s) &= \frac{ \rho(\hs, s) }{ \sum_{s'} \rho(\hs, s') } = \frac{ \ppolicies_{ij} }{ \sum_{j'} \ppolicies_{ij'} } \\
    \rho(s\mid \hs) &= \frac{ \rho(\hs, s) }{ \sum_{\hs'} \rho(\hs', s) } = \frac{ \ppolicies_{ij} }{ \sum_{i'} \ppolicies_{i'j} }
\end{align}
The matrices $\qpolicies_\as$ and $\ppolicies$ contain the necessary information to determine all marginal history $\qpolicies(\hs, \as)$ and state values $\qpolicies(s,\as)$ as normalized dot products:
\begin{align}
    \qpolicies(\hs,\as) &= \sum_s \rho(s|\hs) \qpolicies(\hs s \as)  = \sum_{j'} \frac{ \ppolicies_{ij'} }{ \sum_{j'} \ppolicies_{ij'}  } \qpolicies_{\as,ij'} \,, \label{eq:tabular:vh} \\
    \qpolicies(s,\as) &= \sum_{\hs} \rho(\hs| s) \qpolicies(\hs s \as) = \sum_{i'} \frac{ \ppolicies_{i'j} }{ \sum_{i''} \ppolicies_{i''j} }
    \qpolicies_{\as,i'j} \,. \label{eq:tabular:vs}
\end{align}

On the other hand, the correct expected state value as listed in \Cref{eq:bias} is obtained as
\begin{align}
    &\phantom{{}={}} \Exp_{s\sim\rho(s\mid h)}\left[ \qpolicies(s,\as) \right] \nonumber \\
    &= \sum_s \rho(s\mid \hs) \qpolicies(s,\as) \nonumber \\
    &= \sum_s \rho(s\mid \hs) \sum_{\hs'} \rho(\hs'\mid s) \qpolicies(\hs', s, \as) \nonumber \\
    &= \sum_{j'} \frac{ \ppolicies_{ij'} }{ \sum_{j''} \ppolicies_{ij''} } \sum_{i'} \frac{ \ppolicies_{i'j'} }{ \sum_{i''} \ppolicies_{i''j'} } \qpolicies_{\as, i'j'} \,. \label{eq:tabular:Evs}
\end{align}

Note that $\qpolicies(\hs,\as)$ in \Cref{eq:tabular:vh} only involves the elements on a specific row of both $\qpolicies_\as$ and $\ppolicies$, while the expected state value in \Cref{eq:tabular:Evs} involves all elements of both $\qpolicies_\as$ and $\ppolicies$. 
While this provides an intuitive reason for the fact that the two values are not necessarily the same, the inequality proof is still incomplete;
in fact, the values in $\qpolicies_\as$ and $\ppolicies$ are not arbitrary, but are related by problem dynamics, policies, and history-state Bellman equations, which may still result in \Cref{eq:tabular:vh,eq:tabular:Evs} being numerically equivalent.
However, we prove that this is not the case.
\begin{theorem}\label{thm:bias}
     $\qpolicies(s,\as)$ may be a biased estimate of $\qpolicies(\hs,\as)$ (proof in \Cref{sec:bias:proof}).
\end{theorem}

\begin{corollary}\label{corollary:gradient-bias}
    $\nabla_i J_s$ may be biased (proof in \Cref{appendix:policy-gradient-biase-of-state-based-critic}).
\end{corollary}

The example below serves as prove-by-example for~\Cref{thm:bias}. See~\Cref{sec:bias:proof} for an alternative proof.

\paragraph{Example}
In classic Dec-POMDP domain Dec-Tiger~\cite{nair2003taming}, two agents face two doors, left and right; a tiger is randomly initialized behind one of the doors.
The agents can \textit{listen}, \textit{open-left} or \textit{open-right}.
The \textit{listen} action is used to detect the tiger location and produces either \textit{hear-left} or \textit{hear-right} which indicates the correct tiger location with probability $85\%$.
The cost for \textit{listen} is $-2$; the episode ends with a $-50$ penalty if the tiger door is opened by both agents, and $+20$ for the other door. 
If two different doors are opened by the two agents, the episode ends with a penalty of $-100$.
If only one agent opens a door (the other agent listens), the episode ends with $-101$ if the state is tiger and $+9$ otherwise.
For the purpose of this example, we use a finite horizon of $3$ for calculating the values.
We use the optimal policy in which the agents listen twice and pick the most promising door.
We focus on the history $h=\{(\textit{listen}, \textit{listen}), (\textit{hear\_right}, \textit{hear\_right})\}$ and actions for the agents $a=(\textit{listen}, \textit{listen})$ as our example.
By solving analytically (see supplementary material), we come to the follow history and state values:
\begin{align}
    Q(h, a) &\approx 13.8859 \label{dtiger_q_h_a_value}\\ 
    Q(tiger\_left, a) & = -16.175 \label{dtiger_q_s1_a_value}\\
    Q(tiger\_right, a) & = -16.175 \label{dtiger_q_s2_a_value}
\end{align}
That is, after each agent listens once and both hear the tiger on the right, listening again by both agents is has an estimated value of 13.8859. In contrast, the state-based estimates just represent the value achieved after visiting the state and taking the corresponding action. 
It is obvious that no matter with what probability our state values in \Cref{dtiger_q_s1_a_value,dtiger_q_s2_a_value} are combined, we can only get a value of $-16.175$ which is a biased estimator of our history value in \Cref{dtiger_q_h_a_value}.
Note that our $h$ contains a promising observation $(\textit{hear\_right}, \textit{hear\_right})$ which brings our confidence that the tiger is behind the right to very high ($96.98\%$).
However, as illustrated in \Cref{eq:tabular:Evs}, the state values are  averaged over different histories, good \emph{and} bad. 
The reason why $Q(\textit{tiger\_right}, a)$ is negative is because it has to consider situations where agents get less promising observations (like $(\textit{hear\_left}, \textit{hear\_right})$) or no observations (at the beginning of the episode) while being in the corresponding state.
That is, \emph{history critics represent the true value of the history-based policy while state critics are averaged over all the histories that visit the state}.
As a result, we see that in Dec-Tiger, the state values are biased estimators of history values.

\subsection{Variance Analysis}\label{sec:variance}

In this section, we discuss the effect of state values $\qpolicies(s, \as)$ on the variance of the policy gradient estimate $\hat\nabla_i J_s$.
Like our bias analysis, we discuss the oracle values $Q^\pi$ instead of its estimated counterpart $\hat Q^\pi$.
The policy gradient theorem for Dec-POMDPs explicitly requires the history value $Q^\pi(\hs, \as)$ to be the value used to weigh the policy's score function~\cite{lyu2021contrasting, bono2018cooperative}.
In that capacity, $Q^\pi(\hs, \as)$ is a specific scalar associated with the history, and has no variance. 
On the other hand, using state value $Q^\pi(s, \as)$ as estimator of $Q^\pi(\hs, \as)$ introduces variance, as $s$ is sampled from the history's associated belief $b(\hs)$.
However, it does not necessarily imply that the corresponding gradient $\hat\nabla_i J_s$ also has higher variance than $\hat\nabla_i J_\hs$ in general.
Instead, we show that, if $\qpolicies(s, \as)$ is unbiased for a given policy and a Dec-POMDP, then $\hat\nabla_i J_s$ has a variance greater or equal than that of $\hat\nabla_i J_\hs$.

\begin{theorem}\label{thm:variance}
    When the state value function $\qpolicies(s, \as)$ is unbiased, the state-based policy gradient estimates $\hat\nabla_i J_s$ have a variance greater or equal than that of the history-based policy gradient estimates $\hat\nabla_i J_\hs$, i.e.,
    \begin{equation}
    \begin{aligned}
        &\qpolicies(\hs, \as) = \Exp_{s\sim\rho(s\mid\hs)}\left[ \qpolicies(s, \as) \right] \\
        & \implies \Var\left[ \hat\nabla_i J_s \right] \ge \Var\left[ \hat\nabla_i J_\hs \right].
    \end{aligned}
    \end{equation}
    (proof in \Cref{sec:variance:proof}).
\end{theorem}

Although \Cref{thm:variance} alone contains a result which is conditional to an equality which does not necessarily hold for a generic Dec-POMDP, combining \Cref{thm:bias,thm:variance} results in a broader statement about the overall quality of state-based policy gradient estimates, i.e.,
\begin{corollary}
    $\hat\nabla_i J_s$ cannot be guaranteed to have strictly better bias/variance properties than $\hat\nabla_i J_\hs$, i.e., either its bias is higher (or equal), or its variance is higher (or equal), or neither is lower (or equal) (Follows directly from \Cref{corollary:gradient-bias,thm:variance}).
\end{corollary}

\paragraph{Example}  Consider a (single-agent) \emph{beverage} domain, in which the agent is a barista who serves coffee or tea to a client.  The client, who either prefers \emph{coffee} or \emph{tea}, represents the randomly sampled initial state.  The agent does not observe the client's preference (we denote this as $h=\nohistory$) and receives a reward of $1$ if it chooses to serve the correct beverage, and $-1$ if it chooses the wrong beverage.  In either case, the episode ends.  Suppose the agent chooses to serve \emph{tea}.  Then,
\begin{align}
    \qpolicy(s=\emph{coffee}, a=\emph{tea}) &= -1 \\
    \qpolicy(s=\emph{tea}, a=\emph{tea}) &= 1 \\
    \qpolicy(h=\nohistory, a=\emph{tea}) &= 0
\end{align}
While $\qpolicy(h=\nohistory, a=\emph{tea})$ is a constant with zero variance, the random variable $\qpolicy(s, a=\emph{tea})$ conditioned on $h=\nohistory$ has strictly positive variance,
\begin{align}
    & \Var_{s\mid h=\nohistory}\left[ \qpolicy(s, a=\emph{tea}) \right] \nonumber\\
    &= \Exp_{s\mid h=\nohistory}\left[ \qpolicy(s, a=\emph{tea}) ^2 \right] - \Exp_{s\mid h=\nohistory}\left[ \qpolicy(s, a=\emph{tea}) \right]^2 \nonumber \\
    &= \Exp_{s\mid h=\nohistory}\left[ 1 \right] - 0^2 \nonumber \\
    &= 1 \,.
\end{align}
Therefore, the policy gradient estimates will have a higher variance with the state-based critic.  
For more intuition on the variance of state-based critic, refer to the Dec-Tiger example in \Cref{variance-intuition}.

\section{Experiments}
\label{sec:experiments}

To understand the performance of centralized critics in practice, we test state-based critics and history-based critics using vanilla Advantage Actor-Critic with a centralized critic.
We implement state or history value functions $V(s), V(\vh)$ and $V(\vh,s)$ instead of Q functions; and the advantages are calculated using one-step differences as mentioned in \Cref{subsec:multi_agent_actor_critic_methods}.
We highlight some of the interesting results and discuss the potential reasons behind performance differences.
Furthermore, we introduce and test history-state-based Critics (HSC), where the concatenation of state and history is used as the input of the critic, analogous to Unbiased Asymmetric Actor-Critic~\cite{baisero2021unbiased} in single-agent settings.
HSC uses $V(\vh,s)$ directly as the critic, which can be shown to have no bias but larger variance in theory compared to the history-based critic (\Cref{appendix:state_history_based_critic}).

\paragraph{Experiment Setup}
The figures shown are mean-aggregation of 20 runs per method; standard deviation is drawn as shaded bands around the lines.
The experiments were conducted on compute clusters with nodes equipped with Dual Intel Xeon E5-2650 CPUs and 128GB of RAM. 
Hyperparameters are individually tuned while fixing other hyperparameters.

\subsection{Observation Information Sufficiency}
\label{subsec:observation-information-sufficienty}
Some environments give partial yet "sufficient" local information in the sense that the optimal policy does not depend on the entire history. That is, the history-MDP (or even observation-MDP) is (close to) value-equivalent to the true underlying MDP; in some cases, reactive policies (policies that only condition on the last observation) can achieve optimal performance (\Cref{fig:exps1,fig:grid_small}).
For example, in the Meeting-in-a-Grid domains~\cite{bernstein2005bounded,amato2009incremental}, agents observe their own location but not the teammate while trying to meet in a grid-world.
Optimally, agents would navigate to a predetermined spot (e.g., the center) and wait.
Another example is Find Treasure~\cite{shuo2019maenvs} in which one agent has to step onto a trigger location to open a door while the other agent goes through the door to find treasure. 
Again, even though the agent only observes local information, the optimal policy does not require remembering history because policies do not benefit from additional action-observation history.
In these environments in ~\Cref{fig:exps1}, one may safely assume that a given state  will produce similar return distributions with different histories because the history information does not meaningfully affect the policy and the return distributions
since the policy conditions on the last observation, which is produced by the state. 
However, we note that in harder and more partially-observable tasks, we expect that the optimal policies are not reactive.

\subsection{Reactive Policies}
\label{subsec_experiments_reactive_policies}
We note that using reactive policies is not sufficient to make state-based critics unbiased in the general case, because at any time step the observation produced by the state may still cause the policy to behave differently (\Cref{eq:bias-proof:1}).
Therefore, the state-based critic is unbiased for a reactive policy only when there is also a deterministic observation model.
That is, for a specific agent, a given state-action pair can only produce a certain deterministic observation.
This restriction ensures at most one non-null entry in each column of matrix $Q^\mathbf{\pi}_\as $, thus warranting unbiased expected state-values $Q(o,a) = \E_{s\sim \phi(s\mid h)}Q(s,a)$.
It is precisely the reason why MADDPG is not biased in their particle environments~\cite{lowe2017multi}.
This situation also is not uncommon, which we see in numerous benchmarks including the classic task Recycling as well as environments with radius-based observation models such as StarCraft Multi-Agent Challenge (SMAC)~\cite{samvelyan19smac} which we discuss in~\Cref{subsec:state_representation}.
We hence note that the benchmarks used in recent state-of-the-art works usually has the aforementioned deterministic observation model, which is in fact a special case of imperfect information.

\begin{figure*}[t]
    \begin{subfigure}{.30\textwidth}
    \centering
      \includegraphics[width=1\textwidth]{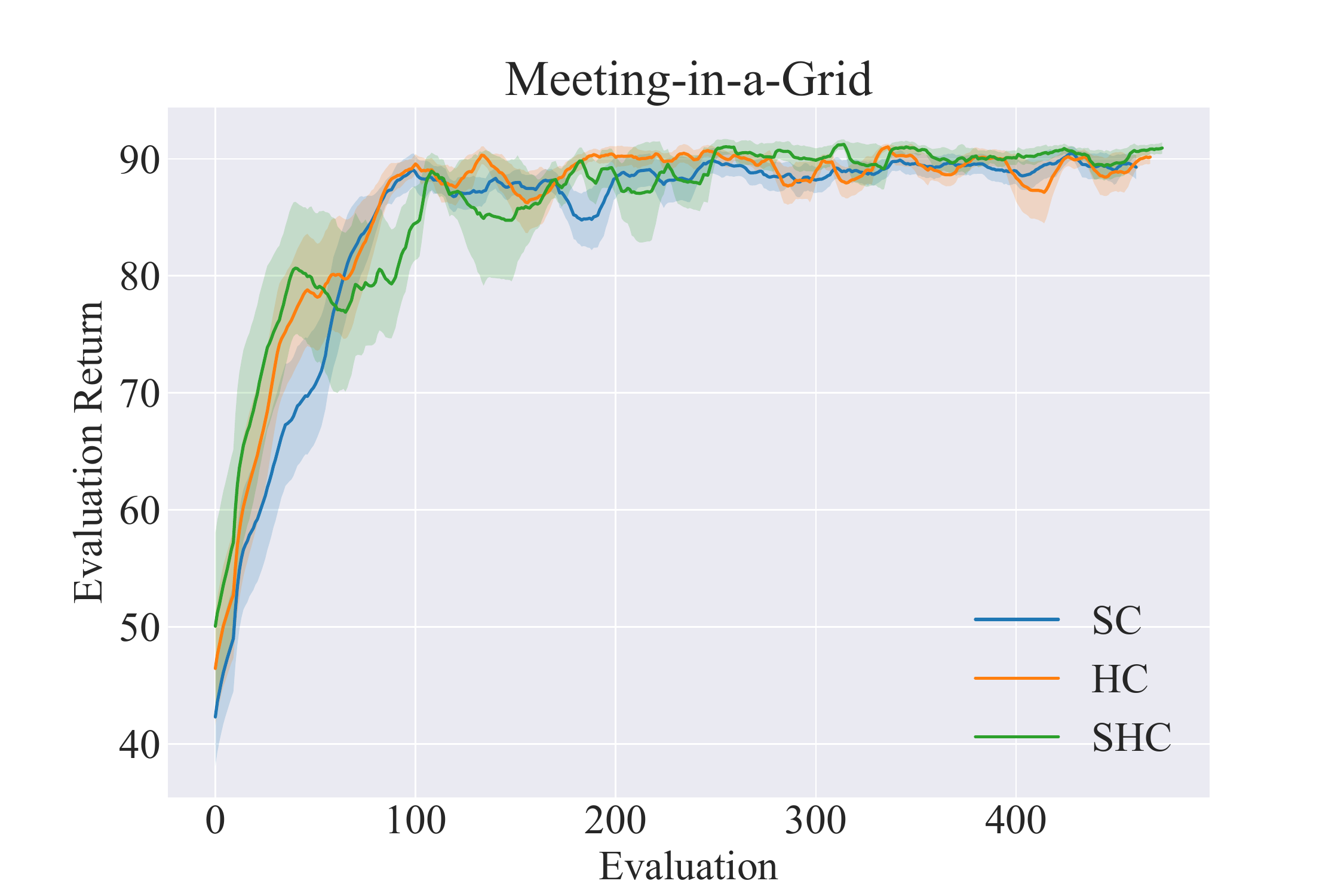}
    \end{subfigure} \hfill
    \begin{subfigure}{.30\textwidth}
      \centering
        \includegraphics[width=1\textwidth]{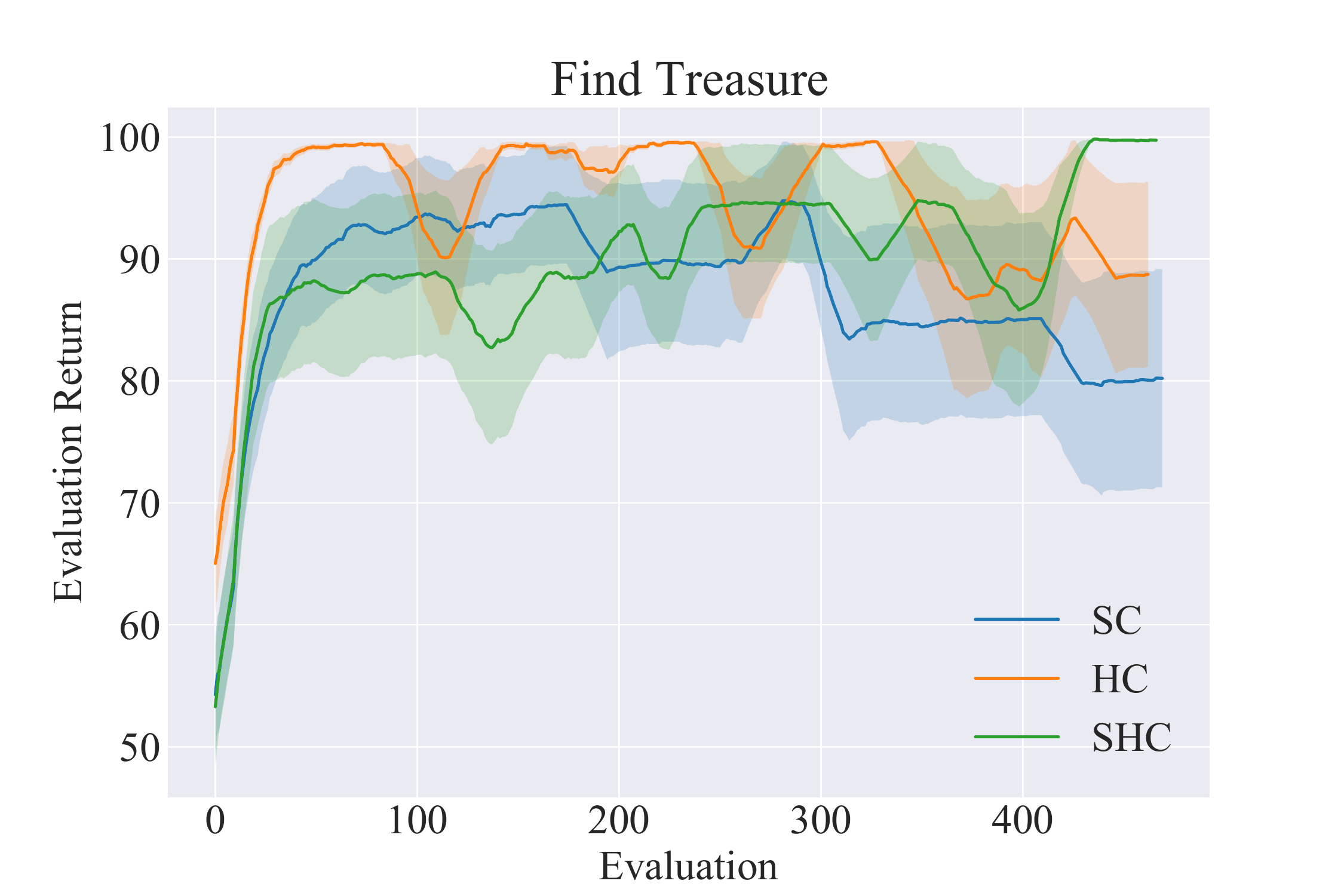}
      \end{subfigure}\hfill
    \begin{subfigure}{.30\textwidth}
    \centering
      \includegraphics[width=1\textwidth]{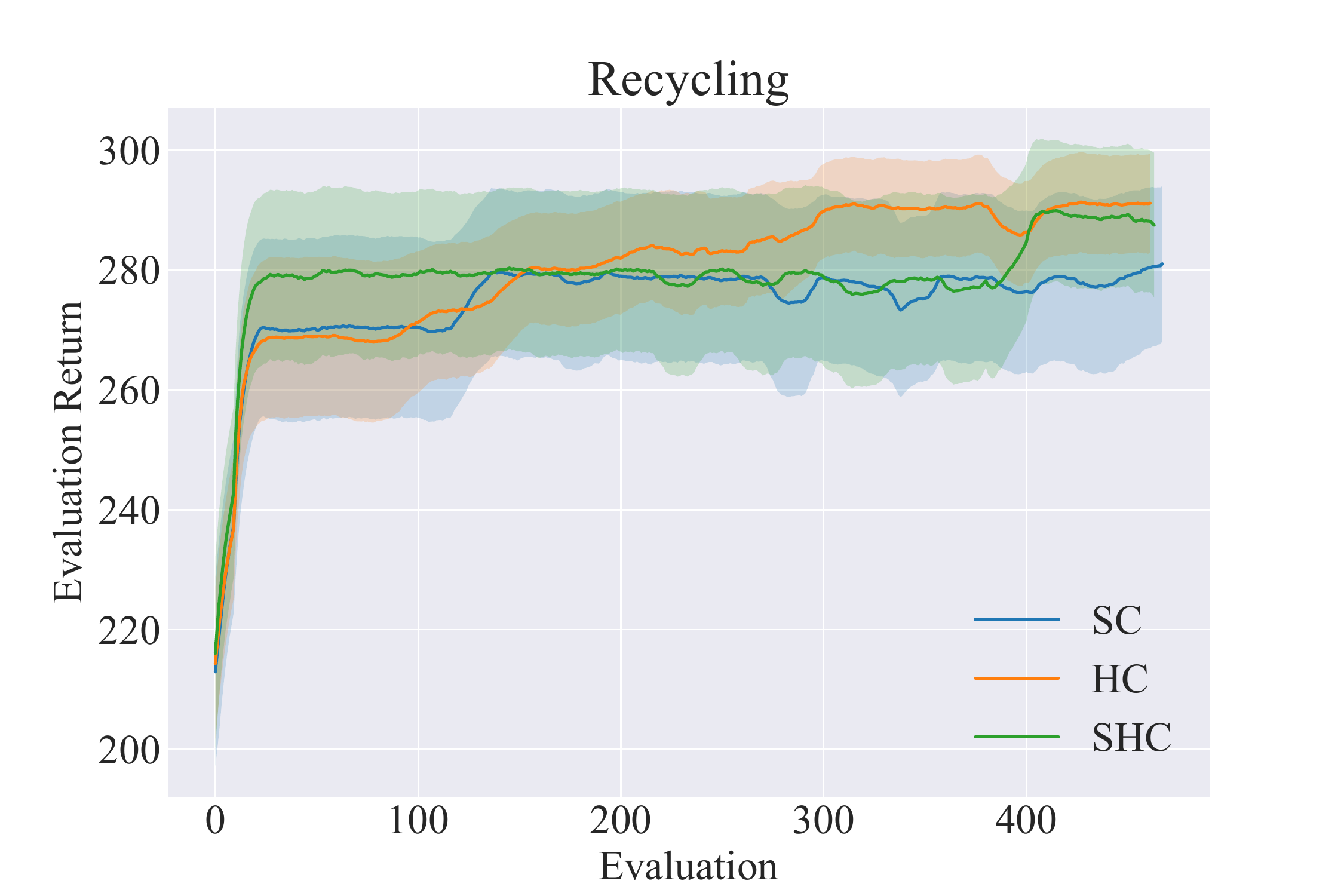}
    \end{subfigure}
    \caption{Performance evaluation of state-based critics (SC), history-based critics (HC) and history-state-based critics (HSC) in cooperative multi-agent partially observable environments: Meeting-in-a-Grid~\cite{amato2009incremental}, Find Treasure~\cite{shuo2019maenvs}, Multi-agent Recycling~\cite{amato2007optimizing}}\label{fig:exps1}
\end{figure*}

\paragraph{Cooperation} The situation where reactive policies are not optimal is captured in the multi-agent recycling task for which policies with different types of critics all converged to a suboptimal solution.
In the multi-agent recycling task, we expect the agents to work together while maintaining battery levels.
The task contains recyclable small and large targets, and the large ones require two agents to act simultaneously.
The agents only observe their own battery levels.
For policies that recycle large targets to be competitive against small-target-policies (reactive), agents must estimate their teammate's battery level based on observation history (non-reactive).
Learning to cooperate in this case is especially difficult because the agents suffer the problem of shadowed equilibrium~\cite{matignon2012independent} despite the usage of a centralized critic~\cite{lyu2021contrasting}.
\Cref{fig:exps1} shows performance for Recycling agents with various critics, in which none of the methods learns to recycle large targets (See \Cref{reactive_policy_figures} for performance of reactive policies).
Therefore,  reactive policies are learned that are not significantly biased with state-based critics.
Note that being unbiased means that the shadowed equilibrium also applies for policies trained with state-based critics.

\begin{figure*}[t]
  \begin{subfigure}{.31\textwidth}
  \centering
    \includegraphics[width=1\textwidth]{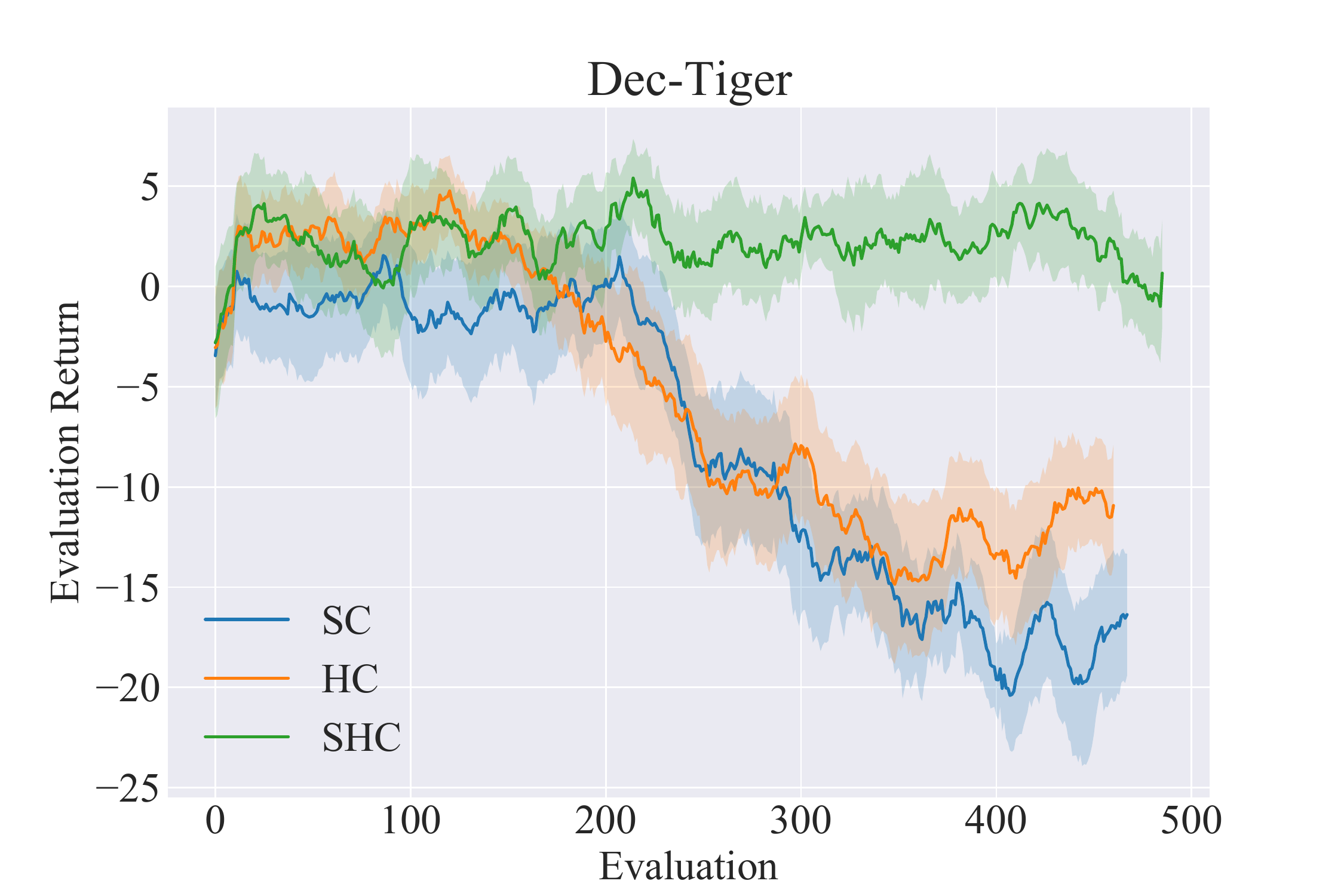}
  \end{subfigure}\hfill
  \begin{subfigure}{.31\textwidth}
  \centering
    \includegraphics[width=1\textwidth]{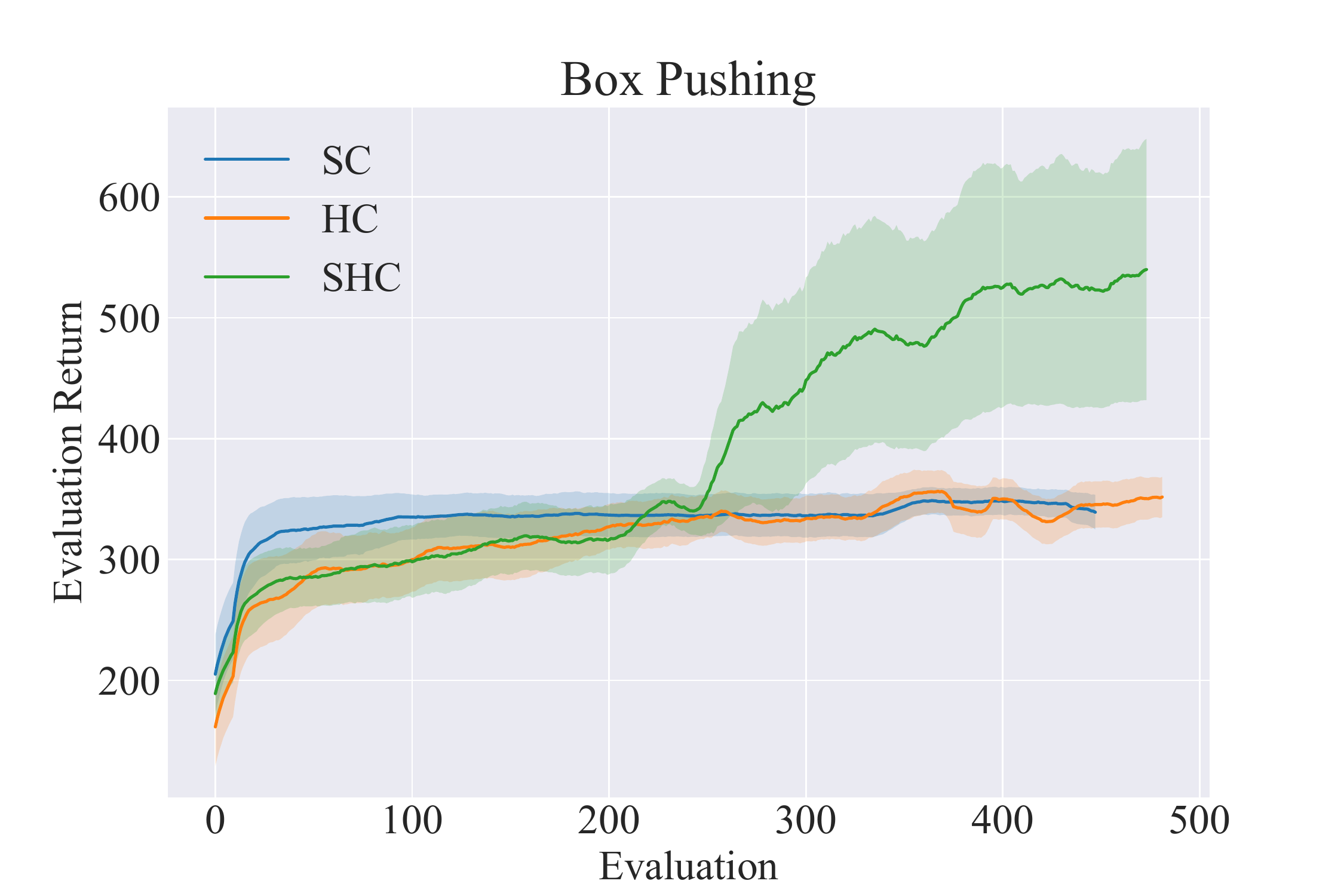}
  \end{subfigure}\hfill
  \begin{subfigure}{.31\textwidth}
  \centering
    \includegraphics[width=1\textwidth]{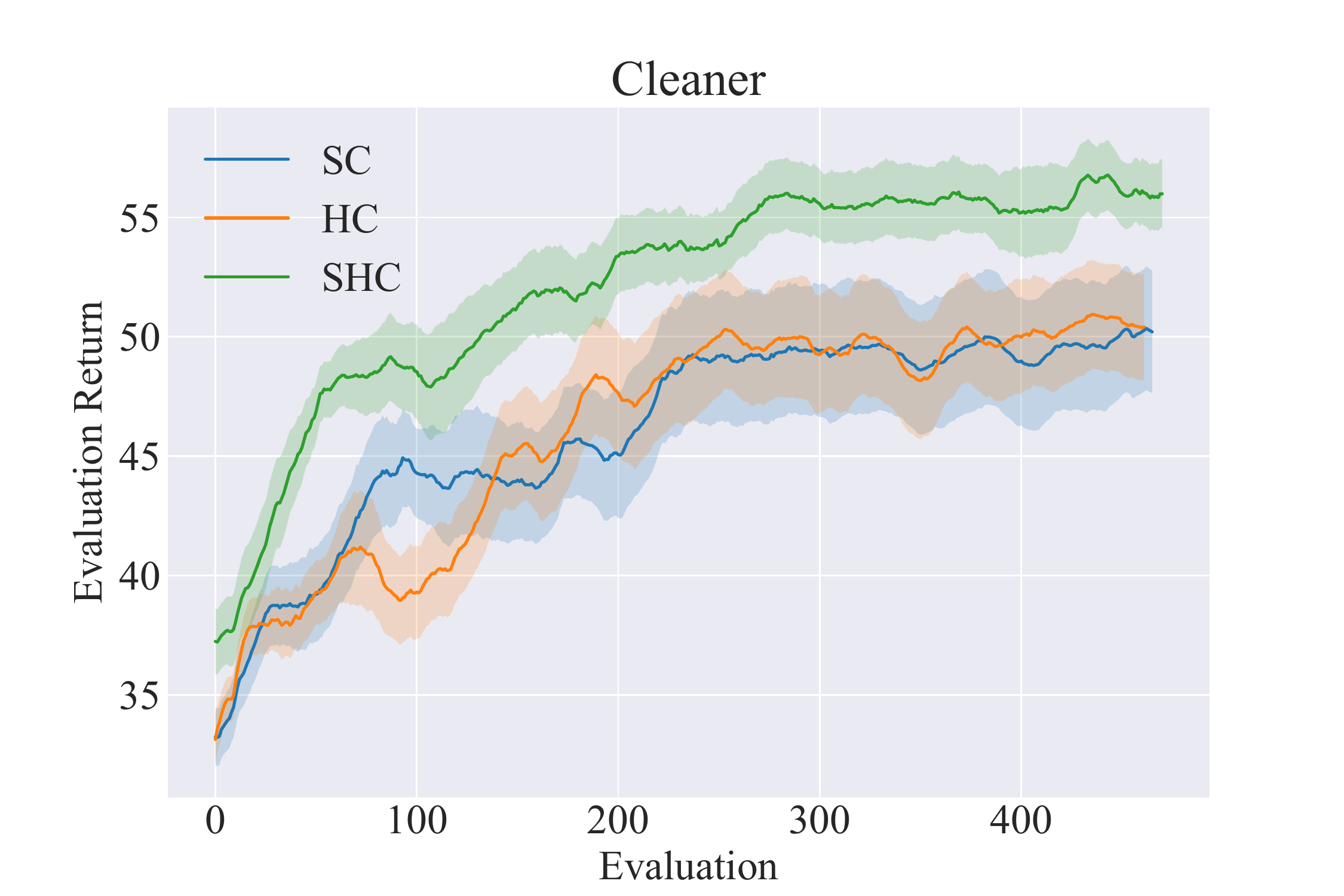}
  \end{subfigure}
  \caption{Performance evaluation of SC, HC and HSC on Dec-Tiger~\cite{nair2003taming}, Box Pushing~\cite{seuken2007improved} and Cleaner~\cite{shuo2019maenvs}.}\label{fig:exps2}
\end{figure*}

\subsection{History-State-Based Critics}
\label{subsec:state_history_based_critics}
We also test history-state-based critics (HSC), which estimate the value of a state and history pair.
For experiments, we concatenate the state and history without changing other aspects of implementation.
Note that the results shown in \Cref{fig:exps2} are not explicitly tuned but use the set of parameters tuned for the history-based critic (HC).
In the Dec-Tiger~\cite{bernstein2005bounded}, Box Pushing and Cleaner~\cite{shuo2019maenvs} domains in \Cref{fig:exps2} we see that the history-state-based critics have a clear advantage over critics that use state or history information alone, especially in the later stages of training.
It suggests that there exist situations where the history-state-based critic can benefit from the advantages of both the state as well as the history as discussed in \Cref{subsec-advantage_of_state_based_critics} and \Cref{subsec-advantage_of_history_based_critics}.
We refer readers to the appendix for the details on those environments,
but generally, both Box Pushing and Cleaner are grid world tasks with observations local to the cells around the agent.
The agent needs to estimate their teammates' locations or paths to act optimally.
Information gathering is even more crucial in Dec-Tiger, where we see history-based critics already distinctly outperform state-based ones, with HSC achieving the best results.
By listening, the agents move from histories with little or no information to histories with more information in the same state;
hence history and history-state critics can accurately represent the different values.
The state-based critic, on the other hand, regard different histories with the same value, hence not incentivizing the policy to gather information, as we have shown in~\Cref{sec:bias}.

\subsection{State Representation}\label{subsec:state_representation}

\begin{figure*}[t]
  \centering
  \includegraphics[width=0.3\linewidth]{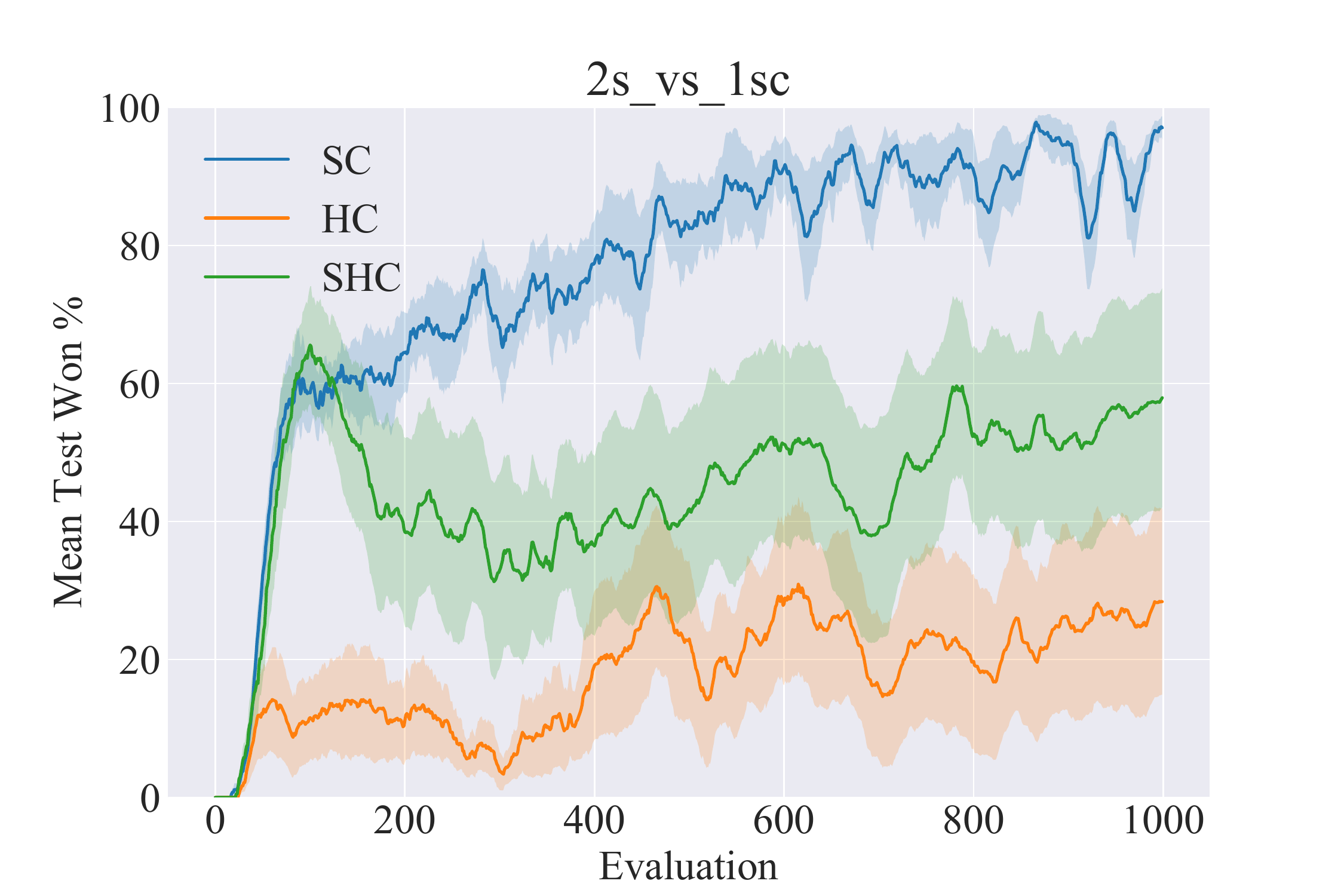}
  \includegraphics[width=0.3\linewidth]{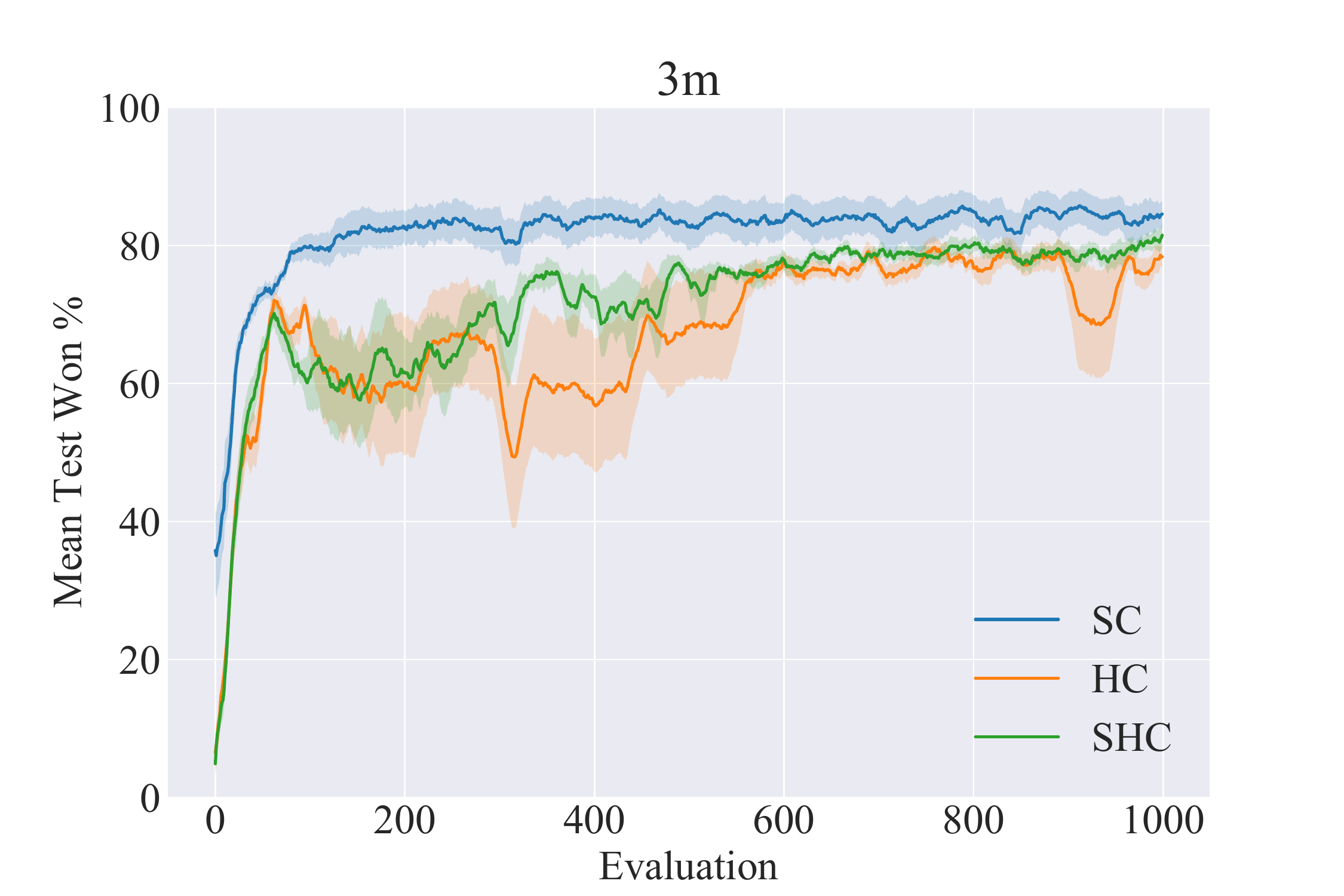} 
  \includegraphics[width=0.3\linewidth]{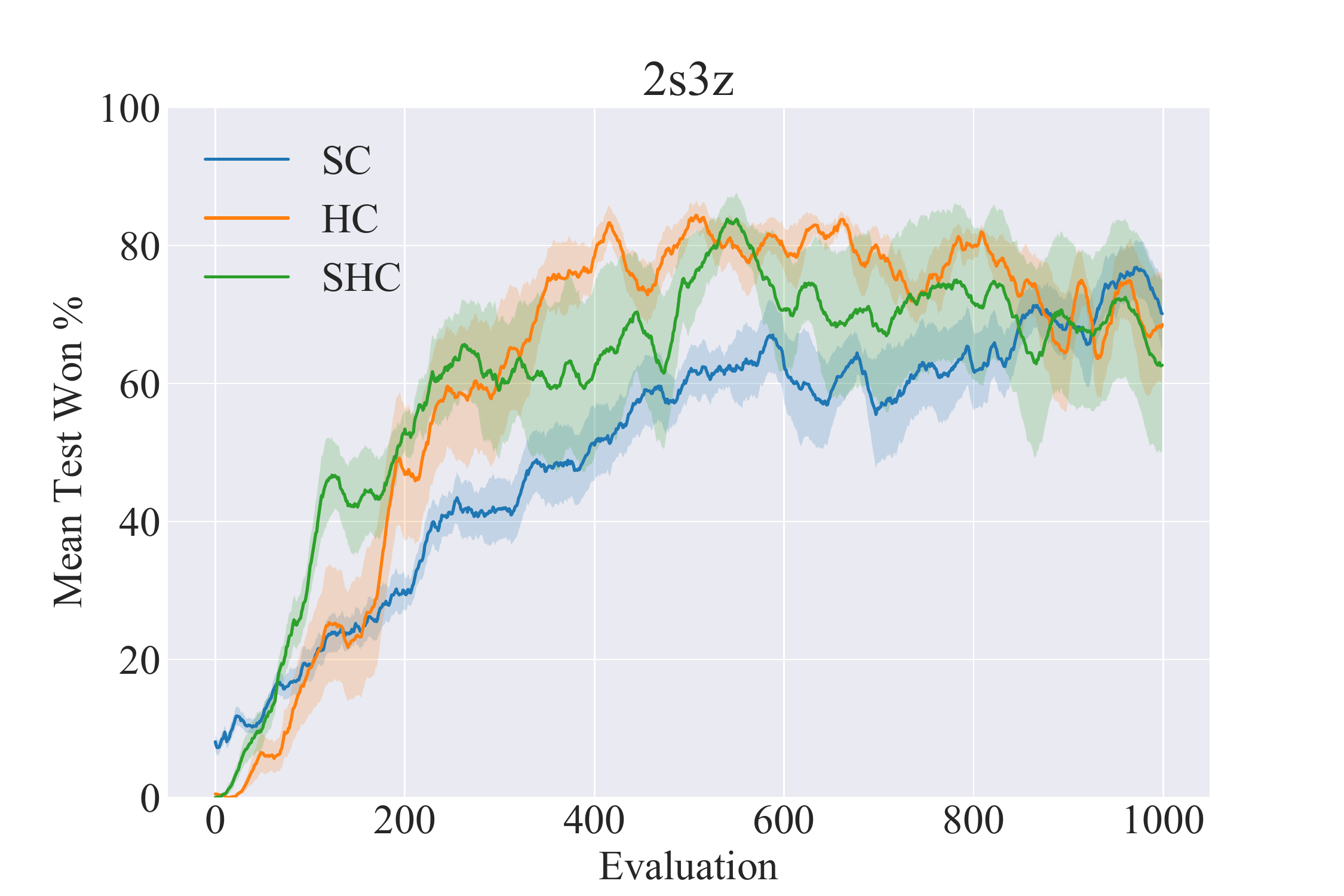}
  \includegraphics[width=0.3\linewidth]{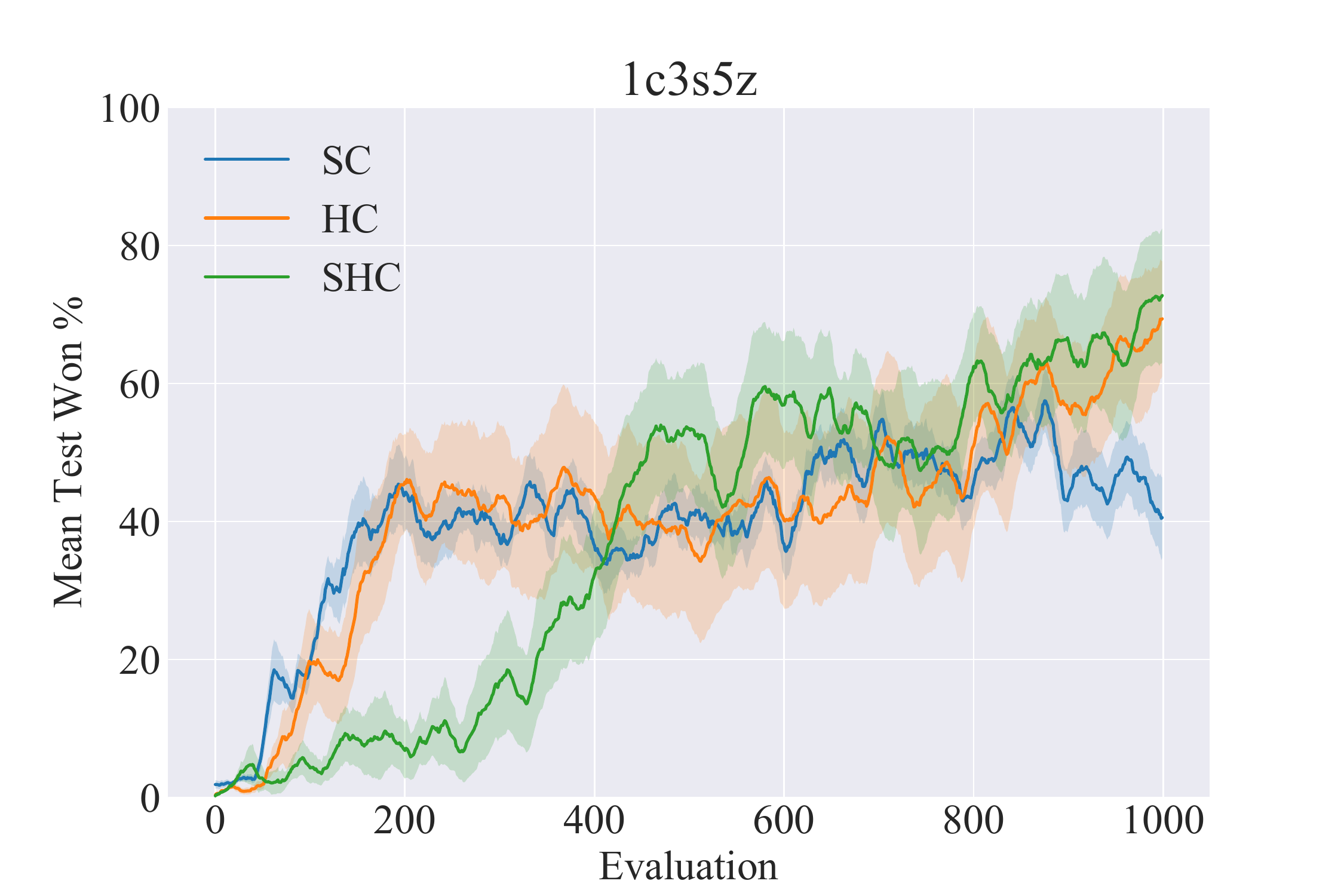}
  \includegraphics[width=0.3\linewidth]{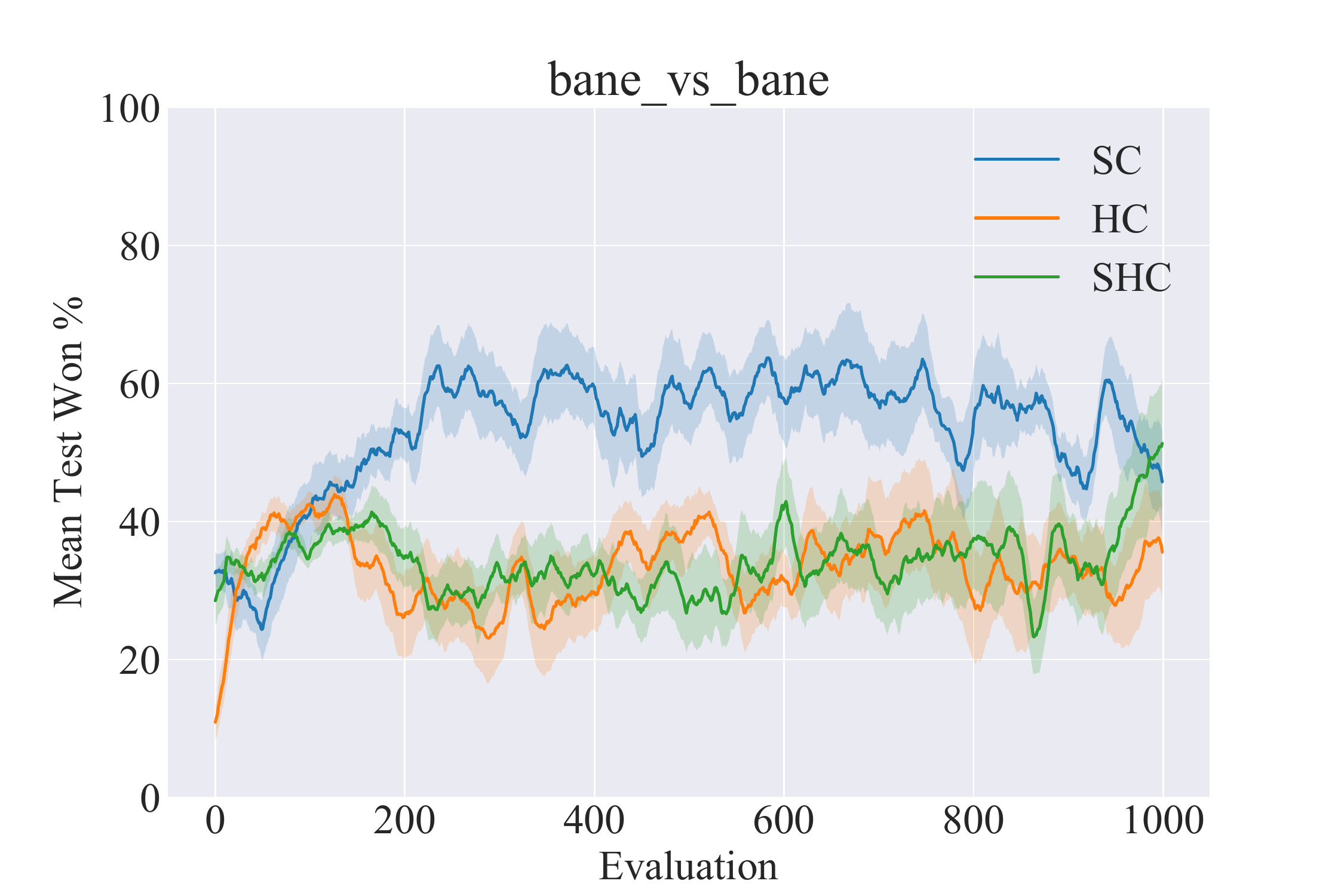}
  \caption{Performance comparison of COMA with different critics in multiple scenarios of SMAC.}
  \label{fig:smac}
\end{figure*}
We also see tasks where it is acceptable or even preferred to use state-based critics.
A typical example is the StarCraft Multi-Agent Challenge (SMAC) benchmark. As seen in \Cref{fig:smac}, most scenarios show that state-based critics exhibit the best overall performance.
In SMAC, each agent controls a combat unit to engage enemy units, and the state includes the statuses of friendly agents and opponent units.
We find SMAC agents usually have self-centered acting limits (e.g., firing ranges).
We find far-away units rarely affect an agent's decisions.
This limit causes the occluded information to rarely affect expected values.
Thus, the history values associated with a state are usually similar; as a result, using a state-based critic does not introduce significant bias in this case.
It is related to results also discussed in ~\Cref{subsec_experiments_reactive_policies}.
In specific scenarios, we see that state-based critics perform better than history-based ones.
Note that the observation for an agent is a masked version of the state, in which the information regarding far-away units is unobservable.
Therefore, both the state and the observation have similar and concise representations, while the history will contain redundant and outdated information. Learning to ignore this information can be difficult and take time. 
Together with a deterministic observation model, and history carrying semi-redundant information,
it is reasonable to believe that state-based critics may give a overall more reliable estimate for policy training.

\section{Discussion}\label{sec:discussion}
We discuss the tasks and scenarios where we can see the advantages of each type of critics and the underlying rationale.

\subsection{Advantage of History-Based Critics}
\label{subsec-advantage_of_history_based_critics}

It is shown that the history-based critic is unbiased in theory, so does it translate to increased performance?
Toy examples aside, some of the more realistic environments shown above may also favor history-based critics due to its ability to evaluate observed information---for a given history, the ability to distinguish how much information was gathered by the policy in the form of observation-action histories and reflect the information in terms of values.
The information gained from observations helps in reducing uncertainty and with such histories the policy can achieve better values compared to ones with less information, and history values encode this value difference explicitly.
At a high level, history-based policies and critics have a simple one-to-one update relationship, but state-based critics have history-based policies that may have a many-to-many update relationship.
Often, you will have two or more histories being mapped to the same state-value when they would have different history values.
This mapping can cause aliasing in values, which is the source of the bias. 

Being able to represent the value of information is the key strength we see in history-based critics and the key weakness of state-based critics.
For example, suppose the agents can gather information regarding the distribution of the world state.
In that case, histories with more information have an advantage compared to less informed histories~\cite{kaelbling1998planning}.
Using the notation in our bias analysis, that is, for a certain state $s$, its history-state values (a column of $Q_a^\pi$) vary depending on the history. %
Each (s,h) pair, therefore, represents a distinct condition in which the agent can have different behaviors (and thus returns), but share the same state value.
Since the state-based value function is the marginalization of the values of these more or less informed history values, state values remain incapable of evaluating how information effects future returns.
On the other hand, for a specific state, the history values distinguish histories with different levels of uncertainty, which results in different payoffs; 
Therefore, in environments where gathering information is critical, we suggest using history-state critics that can take information gathering into account and benefit from state information.

\paragraph{Using State Value as Baseline}
The gradient estimates of implementations which use Monte Carlo returns  with value functions as baselines, such as in DICG~\cite{li2020deep} or MAGIC~\cite{niu2021multi}, are guaranteed to be unbiased.  As a baseline, the bias of the state value function does not affect the bias of the overall policy gradient.  However, it may affect its variance:  although the history value baselines are widely considered to have good variance reduction properties, it is not clear whether the same can be said of state value baselines due to its potential bias.

\subsection{Advantage of State-Based Critics}
\label{subsec-advantage_of_state_based_critics}

On the other hand, state-based critics have proven helpful in recent works~\cite{SQDDPG, LIIR, du2021learning}.
We speculate that there are three main reasons behind state-based critics being able to provide signals that give good performance despite their theoretical shortcomings. 
First, observe that in some environments, using the state as input more easily allows the value function to extract meaningful features compared to extracting features from histories.
It is especially the case when the state representation is concise or when the history feature extraction requires a considerable amount of training.
For example, some grid-world environments' states involve agent locations, while observations stem from on-screen pixels that are of much higher dimension.
This happen to be not unusual in current multi-agent reinforcement learning benchmarks (see \Cref{subsec:state_history_based_critics,subsec:state_representation}).

Second, we observe that the state representation is usually a complete version of the observation information in multi-agent benchmarks used by recent state-of-the-art works.
As discussed in \Cref{subsec:state_representation}, those environments generate observations either by masking the underlying state representation or by outputting state fragments.
As a result, the state information is not drastically different from observations, implying that the state-based value functions are well aligned with history-based value functions. 
This alignment is amplified in situations where the information omitted in observations (e.g., information regarding far-away teammates or opponents) has zero or negligible effect on the expected values, which is applied in popular benchmarks such as SMAC~\cite{samvelyan19smac} and partially observable particle environments~\cite{lowe2017multi}.
Shown in \Cref{fig:smac,fig:speaker_and_listener,fig:cooperative_navigation},
as long as the information gathering issue mentioned above is either non-existent or not effecting the value estimation, the state-based critics can give good return signals for the purpose of policy learning.

\section{Conclusion}
In this paper, we take a close look at state-based critics in multi-agent actor-critic methods.
We show how state-based critic values may incur bias in training decentralized history-based policies.
We also show that, in theory, that state-based critics exhibit more variance in the policy gradient.
We also suggest and evaluate an alternative using state information in conjunction with the history information to train critics which have seen reliable results empirically.
We discuss task-specific conditions in which the bias will prominently occur, explain the empirical performance of these critics on various environments, summarizing where and why state and history-based critics should be effective.
This work fill in the gap of theoretical understanding of state-based critics popular in multi-agent reinforcement learning, providing a principled foundation for future works on centralized critics in multi-agent reinforcement learning.

\section*{Acknowledgments}
This research is supported
by the U.S.~Office of Naval Research  award 
N00014-19-1-2131, Army Research Office award W911NF20-1-0265 and NSF awards 1816382 and 2044993.

\bibliography{references}

\clearpage 

\appendix
\onecolumn

\section{On Assumptions of Previous Works}
\label{on-assumptions-of-previous-works}
In this section, we discuss the assumptions or the limited scope of previous work, which may have contributed to the recent neglect of the precise effects of state-based critics.
\subsection{MADDPG}
The theories given by MADDPG~\cite{lowe2017multi} is only correct under certain implicit conditions.
In particular, it assumes reactive policies and the environment has a deterministic observation model.
In MADDPG, both criteria are met during testing but are not particularly emphasized in the main paper.
Therefore their theories should not be considered true without the above restrictions.
Without emphasizing the scope, MADDPG also states that it is also applicable to recurrent policies,
however, the centralized critic bias depends on the information included in the $x$ in their Eq.5.
Understandably, some works make a very straightforward extension of MADDPG such as learning history-based policy with a state-based critic without further investigating the internal interaction between the critic and the actor during updates (Eq.3);
For example, RMADDPG~\cite{wang2019r} tried to investigate the effect of putting a recurrent layer in actor, critic or both, but without any theoretical justification.
Issues of this nature are what we aim to resolve with this work with a theoretical and empirical analysis.

\subsection{COMA}
Similarly with COMA~\cite{foerster2016learning}, some may regard Eq. 15 as incorrect, or argue that there is an assumption being made implicitly.
In the work of COMA, without deeper discussion, Eq. 15 assumes that individual history-based actors can be combined into a centralized state-based policy.
Indeed, under such an assumption, it seems using a state-critic with history-based actors won’t lead to extra bias.
However, this assumption cannot hold in general and cannot hold even with COMA on SMAC: the centralized behavior conditioned on state is highly likely to be different from decentralized behavior based on each agent’s local history,
\begin{equation}
    \mathbf{\pi}(\mathbf{u}\mid s) = \prod_a\pi^a(u^a\mid s) \neq \prod_a\pi^a(u^a\mid\tau^a)
\end{equation}
Consider the following example: let two local histories lead to a same state, the decentralized/history-based policy can produce two different action distributions, while the centralized state-based policy can only produce one action distribution.

\section{The On-Policy history-state Distributions}
\label{appendix:on-policy-dist}
Here, we define the ``on-policy'' history-state distribution in the context of finite-horizon Dec-POMDPs. 
These definitions are analogous to those by Sutton and Barto~\cite{Sutton1998}, which we extend to the multi-agent case.

Let $\eta(\hs,s)$ be the expected number of time-steps that the agents spend in state $s$ while having observed the joint histories $\hs$. 
Denoting the marginal distribution of states and joint histories at a specific time-step as $\Pr_t(\hs,s)$, we have that $\eta$ satisfies
\begin{equation}
    \eta(\hs, s) = \Pr{}_0(\hs, s) + \sum_{\bar\hs,\bar s} \eta(\bar\hs,\bar s) \sum_\as \policies(\as\mid \bar\hs) \Pr(\hs, s\mid \bar\hs, \bar s, \as) \,.
\end{equation}
which is solved by $\eta(\hs, s) \doteq \sum_t \Pr_t(\hs, s)$. 
Note that it is solvable because the set of histories are finite due to the finite-horizon assumption.
The on-policy history-state distribution $\rho(\hs, s)$ is defined by normalizing the visitation counts,
\begin{equation}
    \rho(\hs, s) \doteq \frac{ \eta(\hs, s) }{ \sum_{\hs',s'} \eta(\hs, s) } \,.
\end{equation}
From the joint history-state distribution $\rho(\hs, s)$, we can also define the respective marginal and conditional distributions,
\begin{align}
    \rho(\hs) &\doteq \sum_s \rho(\hs, s) \,, &
    \rho(s) &\doteq \sum_\hs \rho(\hs, s) \,, \\
    \rho(\hs\mid s) &\doteq \frac{ \rho(\hs, s) }{ \rho(s) } \,, &
    \rho(s\mid \hs) &\doteq \frac{ \rho(\hs, s) }{ \rho(\hs) } \,.
\end{align}

\section{Proofs}\label{appendix:proofs}

\subsection{Bias of State Values}\label{sec:bias:proof}

Here we prove \Cref{thm:bias} using an argument similar to that made in \cite{baisero2021unbiased}, with the key differences that our case involves arbitrary finite-horizon environments and policies (not reactive policies), $Q$ values (not $V$ values), and the on-policy history/state distributions (rather than the belief distribution).

\begin{proof}
For a generic environment, consider an arbitrary joint action $\as$, and two different joint histories $\hs\neq\hs'$ which are associated with the same empirical conditional state distribution $\rho(s\mid \hs) = \rho(s\mid \hs')$ (a fairly common occurrence in partially observable environments).  Because the joint histories are different, the future behaviors of the agents are generally going to differ, generally resulting in different history values,
\begin{equation}
    \qpolicies(\hs, \as) \neq \qpolicies(\hs', \as) \,. \label{eq:bias-proof:1}
\end{equation}

On the other hand, because both the conditional state distributions and the individual state values are the same for every state, the expected state values must be the same,
\begin{equation}
    \E_{s\sim\rho(s\mid \hs)}\left[ \qpolicies(s, \as) \right] = \E_{s\sim\rho(s\mid \hs')} \left[ \qpolicies(s, \as) \right] \,. \label{eq:bias-proof:2}
\end{equation}
For these specific histories, the equality $\qpolicies(\hs, a) = \Exp_{s\sim\rho(s\mid \hs)}\left[ \qpolicies(s, \as) \right]$, combined with \cref{eq:bias-proof:1}, leads to a contradiction of \cref{eq:bias-proof:2}.
\begin{equation}
    \Exp_{s\sim\rho(s\mid \hs)}\left[ \qpolicies(s, \as) \right] = \qpolicies(\hs, \as) \neq \qpolicies(\hs', \as) = \Exp_{s\sim\rho(s\mid \hs')}\left[ \qpolicies(s, \as) \right] \,.
\end{equation}
\end{proof}

\subsection{Policy Gradient Bias of State-Based Critic}
\label{appendix:policy-gradient-biase-of-state-based-critic}

\subsubsection{Joint Policies} 

Here we prove \Cref{corollary:gradient-bias} with the case of joint policies.  We will be considering ``tabular'' agent policies, i.e., policies which are parameterized by a separate parameter for each possible joint history-action pair, $\policy(\as; \hs) = \theta^\top \mathbbm{1}_{\hs,\as}$.
We first note that any non-tabular policy can be converted into an equivalent tabular policy which behaves the same way and chooses the same actions under the same histories. 
Therefore, because \Cref{thm:bias} says that there is at least one policy for which the state values are biased, then there is also at least one tabular policy for which the state values are biased.

\begin{proof}
We prove that the state-based critic's policy gradient is biased if the values are biased for tabular joint policies.
We prove by contradiction.
We first assume that $\nabla_i J_\hs = \nabla_i J_s$, and will show that under this assumption we have to conclude that the policies' state values are unbiased, which is a contradiction with \Cref{thm:bias} (which, as noted above, is also valid on the sub-family of tabular policies).

For tabular policies, we have
\begin{align}
    \nabla_\theta \log\pi(\as; \hs) &= \frac{ \nabla_\theta \pi(\as; \hs) }{ \pi(\as; \hs) } \\
    &= \frac{ \nabla_\theta \left( \theta^\top \mathbbm{1}_{\hs,\as} \right) }{ \pi(\as; \hs) } \\
    &= \frac{ \mathbbm{1}_{\hs,\as} }{ \pi(\as; \hs) } \,.
\end{align}

Therefore, 
\begin{align}
    \nabla_i J_\hs &= \E_{\hs\sim\rho(\hs), \as\sim\policies(\hs)}\left[ \qpolicies(\hs, \as) \nabla_{\theta_i} \log\pi(\as; \hs) \right] \\
    &= \E_{\hs\sim\rho(\hs), \as\sim\policies(\hs)}\left[ \qpolicies(\hs, \as) \frac{ \mathbbm{1}_{\hs,\as} }{ \pi(\as; \hs) } \right] \,, \\
    \nabla_i J_s &= \E_{\hs, s\sim\rho(\hs, s), \as\sim\policies(\hs)}\left[ \qpolicies(s, \as) \nabla_{\theta_i} \log\pi(\as; \hs) \right] \\
    &= \E_{\hs\sim\rho(\hs), \as\sim\policies(\hs)}\left[ \E_{s\sim\rho(s\mid \hs)}\left[ \qpolicies(s, \as) \right] \frac{ \mathbbm{1}_{\hs,\as} }{ \pi(\as; \hs) } \right] \,.
\end{align}
Using the assumption of $\nabla_i J_\hs = \nabla_i J_s$, we have 
\begin{align}
    \E_{\hs\sim\rho(\hs), \as\sim\policies(\hs)}\left[ \qpolicies(\hs, \as) \frac{ \mathbbm{1}_{\hs,\as} }{ \pi(\as; \hs) } \right] 
    &= \E_{\hs\sim\rho(\hs), \as\sim\policies(\hs)}\left[ \E_{s\sim\rho(s\mid \hs)}\left[ \qpolicies(s, \as) \right] \frac{ \mathbbm{1}_{\hs,\as} }{ \pi(\as; \hs) } \right] \\ 
    \E_{\hs\sim\rho(\hs), \as\sim\policies(\hs)}\mathbbm{1}_{\hs,\as}\left[ \qpolicies(\hs, \as) \frac{ 1 }{ \pi(\as; \hs) } \right] 
    &= \E_{\hs\sim\rho(\hs), \as\sim\policies(\hs)}\mathbbm{1}_{\hs,\as}\left[ \E_{s\sim\rho(s\mid \hs)}\left[ \qpolicies(s, \as) \right] \frac{ 1 }{ \pi(\as; \hs) } \right]
\end{align}
Due to the one-hot vector, we only have one non-zero value from both sides along each $(\hs,\as)$ dimension, therefore
\begin{align}
     \qpolicies(\hs, \as) \frac{ 1 }{ \pi(\as; \hs) }  
    &=  \E_{s\sim\rho(s\mid \hs)} [ \qpolicies(s, \as) ] \frac{ 1 }{ \pi(\as; \hs) } \label{eq44} \\ 
     \qpolicies(\hs, \as) &=  \E_{s\sim\rho(s\mid \hs)} [ \qpolicies(s, \as) ]\,,
\end{align}
which contradicts~\Cref{thm:bias}.
\end{proof}

\subsubsection{Independent Policies}
Here we prove \Cref{corollary:gradient-bias} with the case of independent policies. 
We continue to use ``tabular'' agent policies, $\policy_i(a_i; h_i) = \theta^\top \mathbbm{1}_{h_i,a_i}$ and prove by example.

\begin{proof}
Continuing from the Dec-Tiger example in~\Cref{sec:bias}.
We explicitly calculate the gradient for the case of $h_i=(\textit{listen}, \textit{hear-left}, \textit{listen}, \textit{hear-right}), a_i=\textit{listen}$.
Here we use a horizon of 4 such that $\pi(a_i;h_i) = 1$.
We denote the relevant joint histories as: 
\begin{align}
&\hs_1 = (\textit{listen}, \textit{listen}), (\textit{hear-left}, \textit{hear-left}), (\textit{listen}, \textit{listen}), (\textit{hear-right}, \textit{hear-left}) \\ &
\hs_2 = (\textit{listen}, \textit{listen}), (\textit{hear-left}, \textit{hear-left}), (\textit{listen}, \textit{listen}), (\textit{hear-right}, \textit{hear-right}) \\ &  
\hs_3 = (\textit{listen}, \textit{listen}), (\textit{hear-left}, \textit{hear-right}), (\textit{listen}, \textit{listen}), (\textit{hear-right}, \textit{hear-left}) \\ &  
\hs_4 = (\textit{listen}, \textit{listen}), (\textit{hear-left}, \textit{hear-right}), (\textit{listen}, \textit{listen}), (\textit{hear-right}, \textit{hear-right})
\end{align}
We first calculate the gradient provided by the history values:
\begin{equation}
\begin{aligned}
    & \nabla J_\hs (h_i, a_i) \\
    &= \E_{\hs,\as|h_i,a_i}Q^\pi(\hs, \as) \frac{\mathbbm{1}_{h_i,a_i}}{\pi(a_i;h_i)} \\
    &= Pr(\hs_1 \mid h_i, a_i) \cdot Q^\pi(\hs_1, (\textit{listen}, \textit{listen})) {\mathbbm{1}_{h_i,a_i}} 
    + Pr(\hs_2 \mid h_i, a_i) \cdot Q^\pi(\hs_2, (\textit{listen}, \textit{listen})) {\mathbbm{1}_{h_i,a_i}} \\  
    &+ Pr(\hs_3 \mid h_i, a_i) \cdot Q^\pi(\hs_3, (\textit{listen}, \textit{listen})) {\mathbbm{1}_{h_i,a_i}} 
    + Pr(\hs_4 \mid h_i, a_i) \cdot Q^\pi(\hs_4, (\textit{listen}, \textit{listen})) {\mathbbm{1}_{h_i,a_i}} \\  
     & = (0.3725 \cdot -6.854 + 0.1275 \cdot -18.175 + 0.1275 \cdot -18.175 + 0.3725 \cdot -6.854) \mathbbm{1}_{h_i,a_i} \\
     & = -9.74\, \mathbbm{1}_{h_i,a_i} \,.
\end{aligned}
\end{equation}
We then calculate the gradient provided by the state values:
\begin{equation}
\begin{aligned}
    & \nabla J_s (h_i, a_i) \\ &= \E_{\hs,\as|h_i,a_i}\E_{s\sim\rho(s\mid \hs)}Q(s, \as) \frac{\mathbbm{1}_{h_i,a_i}}{\pi(a_i;h_i)} \\
    &= Pr(\hs_1 \mid h_i, a_i) \cdot \E_{s\sim\rho(s\mid \hs)}Q(s, (\textit{listen}, \textit{listen})) {\mathbbm{1}_{h_i,a_i}} + Pr(\hs_2 \mid h_i, a_i) \cdot \E_{s\sim\rho(s\mid \hs)}Q(s, (\textit{listen}, \textit{listen})) {\mathbbm{1}_{h_i,a_i}}\\ & + 
     Pr(\hs_3 \mid h_i, a_i) \cdot \E_{s\sim\rho(s\mid \hs)}Q(s, (\textit{listen}, \textit{listen})) {\mathbbm{1}_{h_i,a_i}} + 
     Pr(\hs_4 \mid h_i, a_i) \cdot \E_{s\sim\rho(s\mid \hs)}Q(s, (\textit{listen}, \textit{listen})) {\mathbbm{1}_{h_i,a_i}} \\
     & = (0.3725 \cdot 0.0474+ 0.1275 \cdot 0.0474+ 0.1275 \cdot 0.0474+ 0.3725 \cdot 0.0474) \mathbbm{1}_{h_i,a_i} \\
     & = 0.0474\, \mathbbm{1}_{h_i,a_i} \,.
\end{aligned}
\end{equation}
Therefore, $\nabla J_s \neq \nabla J_h$.
\end{proof}

\subsection{Variance of State-Based Gradient Estimates}\label{sec:variance:proof}

Here, we complete the proof for \Cref{thm:variance} which states that, if $\qpolicies(s, \as)$ is unbiased, then the variance of the state-based policy gradient estimates $\hat\nabla_i J_s$ are greater or equal than that of the history-based policy gradient estimates $\hat\nabla_i J_\hs$.

First, we find that, for any given joint history $\hs$ and joint action $\as$,
\begin{align}
    \Exp_{s\mid\hs}\left[ \hat\nabla_i J_s \right] &=  \Exp_{s\mid\hs}\left[ \qpolicies(s, \as) \nabla\log\pi_i(a_i; h_i) \right] \nonumber \\
    &=  \Exp_{s\mid\hs}\left[ \qpolicies(s, \as) \right] \nabla\log\pi_i(a_i; h_i) \nonumber \\
    &=  \qpolicies(\hs, \as) \nabla\log\pi_i(a_i; h_i) \nonumber \\
    &=  \hat\nabla_i J_\hs \,.
\end{align}

We will use the fact that history-based critic policy gradient is the expectation of the state-based critic policy gradient to establish the variance relationship.
Then, using Jensen's inequality, we get,
\begin{align}
    \Var\left[ \hat\nabla_i J_s \right] &= \Exp_{\hs,s,\as}\left[ \hat\nabla_i J_s^\T \hat\nabla_i J_s \right] - \Exp_{\hs,s,\as}\left[ \hat\nabla_i J_s \right]^\T \Exp_{\hs,s,\as}\left[ \hat\nabla_i J_s \right] \nonumber \\
    &= \Exp_{\hs,\as}\left[ \Exp_{s\mid\hs}\left[ \hat\nabla_i J_s^\T \hat\nabla_i J_s \right] \right] - \Exp_{\hs,\as}\left[ \Exp_{s\mid\hs}\left[ \hat\nabla_i J_s \right] \right]^\T \Exp_{\hs,\as}\left[ \Exp_{s\mid \hs} \left[ \hat\nabla_i J_s \right] \right] \nonumber \\
    &\ge \Exp_{\hs,\as}\left[ \Exp_{s\mid\hs}\left[ \hat\nabla_i J_s \right]^\T \Exp_{s\mid\hs}\left[ \hat\nabla_i J_s \right] \right] - \Exp_{\hs,\as}\left[ \Exp_{s\mid\hs}\left[ \hat\nabla_i J_s \right] \right]^\T \Exp_{\hs,\as}\left[ \Exp_{s\mid \hs} \left[ \hat\nabla_i J_s \right] \right] \nonumber \\
    &= \Exp_{\hs,\as}\left[ \hat\nabla_i J_\hs^\T \hat\nabla_i J_\hs \right] - \Exp_{\hs,\as}\left[ \hat\nabla_i J_\hs \right]^\T \Exp_{\hs,\as}\left[ \hat\nabla_i J_\hs \right] \nonumber \\
    &= \Var\left[ \hat\nabla_i J_\hs \right] \,.
\end{align}

We will discuss the intuition behind this proof in~\Cref{variance-intuition}.

\section{history-state Values and Gradient Variances}
\label{appendix:state_history_based_critic}

Let us consider a variant of the policy gradient which uses the history-state value function $\qpolicies(\hs, s)$,
\begin{equation}
    \nabla_i J_{\hs s} = \Exp_{\hs,s\sim\rho(\hs,s),\as\sim\policies(\hs)}\left[ \qpolicies(\hs, s, \as) \nabla_{\theta_i} \log\pi_i(a_i; h_i) \right] \,.
\end{equation}

As for $\hat\nabla_i J_\hs$ and $\hat\nabla_i J_s$, we consider the Monte Carlo estimate associated with a single joint-history, state, and joint action sample,
\begin{equation}
    \hat\nabla_i J_{\hs s} = \qpolicies(\hs, s, \as) \nabla\log\pi_i(a_i; h_i) \,,
\end{equation}
\noindent where $\hs,s\sim\rho(\hs,s)$ and $\as\sim\policies(\hs)$.

Next, we show that $\hat\nabla_i J_{\hs s}$ is unbiased,
\begin{align}
    \Exp_{\hs,s\sim\rho(\hs,s)}\left[ \hat\nabla_i J_{\hs s} \right] &=  \Exp_{\hs\sim\rho(\hs)}\left[ \Exp_{s\sim\rho(s\mid \hs)}\left[ \hat\nabla_i J_{\hs s} \right] \right] \nonumber \\
    &= \Exp_{\hs\sim\rho(\hs)}\left[ \Exp_{s\sim\rho(s\mid \hs)}\left[ \qpolicies(\hs,s,\as) \nabla_i\log\pi_i(a_i; h_i) \right] \right] \nonumber \\
    &= \Exp_{\hs\sim\rho(\hs)}\left[ \Exp_{s\sim\rho(s\mid \hs)}\left[ \qpolicies(\hs,s,\as) \right] \nabla_i\log\pi_i(a_i; h_i) \right] \nonumber \\
    &= \Exp_{\hs\sim\rho(\hs)}\left[ \qpolicies(\hs,\as) \nabla_i\log\pi_i(a_i; h_i) \right] \nonumber \\
    &= \nabla_i J_\hs \,.
\end{align}

On the other hand, its variance is at least as high as that of $\hat\nabla_i J_\hs$.  This proof is similar to that in \Cref{sec:variance:proof}, except that it does not rely on an assumption of no bias, since no bias is already proven.  First, we find that, for any given joint history $\hs$ and joint action $\as$,
\begin{align}
    \Exp_{s\mid\hs}\left[ \hat\nabla_i J_{\hs s} \right] &= \Exp_{s\mid\hs}\left[ \qpolicies(\hs, s, \as) \nabla\log\pi_i(a_i; h_i) \right] \nonumber \\
    &= \Exp_{s\mid\hs}\left[ \qpolicies(\hs, s, \as) \right] \nabla\log\pi_i(a_i; h_i) \nonumber \\
    &= \qpolicies(\hs, \as) \nabla\log\pi_i(a_i; h_i) \nonumber \\
    &= \hat\nabla_i J_\hs
\end{align}

Then,
\begin{align}
    \Var\left[ \hat\nabla_i J_{\hs s} \right] &= \Exp_{\hs,s,\as}\left[ \hat\nabla_i J_{\hs s}^\T \hat\nabla_i J_{\hs s} \right] - \Exp_{\hs,s,\as}\left[ \hat\nabla_i J_{\hs s} \right]^\T \Exp_{\hs,s,\as}\left[ \hat\nabla_i J_{\hs s} \right] \nonumber \\
    &= \Exp_{\hs,\as}\left[ \Exp_{s\mid\hs}\left[ \hat\nabla_i J_{\hs s}^\T \hat\nabla_i J_{\hs s} \right] \right] - \Exp_{\hs,\as}\left[ \Exp_{s\mid\hs}\left[ \hat\nabla_i J_{\hs s} \right] \right]^\T \Exp_{\hs,\as}\left[ \Exp_{s\mid\hs} \left[ \hat\nabla_i J_{\hs s} \right] \right] \nonumber \\
    &\ge \Exp_{\hs,\as}\left[ \Exp_{s\mid\hs}\left[ \hat\nabla_i J_{\hs s} \right]^\T \Exp_{s\mid\hs}\left[ \hat\nabla_i J_{\hs s} \right] \right] - \Exp_{\hs,\as}\left[ \Exp_{s\mid\hs}\left[ \hat\nabla_i J_{\hs s} \right] \right]^\T \Exp_{\hs,\as}\left[ \Exp_{s\mid\hs} \left[ \hat\nabla_i J_{\hs s} \right] \right] \nonumber \\
    &= \Exp_{\hs,\as}\left[ \hat\nabla_i J_\hs^\T \hat\nabla_i J_\hs \right] - \Exp_{\hs,\as}\left[ \hat\nabla_i J_\hs \right]^\T \Exp_{\hs,\as}\left[ \hat\nabla_i J_\hs \right] \nonumber \\
    &= \Var\left[ \hat\nabla_i J_\hs \right] \,.
\end{align}

\section{Variance Intuition}\label{variance-intuition}
We again use the Dec-Tiger problem~\cite{nair2003taming} as a thought exercise to highlight the intuition behind the gradient variance bound of state-based critic.
We use the same setting from~\cref{sec:bias}.
We denote left door with $x$ and right with $y$.
Consider the following trajectory: $\{o_1 = \varnothing ,\ a_1 = l,\ o_2 = x,\ a_2 = l, o_3=x \}$ which we simplify as $h=lxlx$, where the agent listened twice and observed that the treasure is likely to be inside door $x$ both of the time.

We now look at the true value functions of Dec-Tiger.
Note that the history values of opening doors depend on the observations, the more information contained in the history, the more we expect the agent to obtain high return, i.e. $Q(\varnothing,l) \leq Q(lx, l) \leq Q(lxlx,l)$.
This table show three different histories' values with the same underlying state, as the agent gathers more information, the history value improves.

\begin{center}
\begin{tabular}{c|c|c|c}
$h$ & $s$ &$Q(s,a=l)$&$Q(h,a=l)$\\ \hline
$\varnothing$ &$x$ &$0.0474$ &$0.0474$ \\ \hline
$lx$ &$x$ &$0.0474$ &$7.23$ \\ \hline
$lxlx$ &$x$ &$0.0474$ & $15.9$ 
\end{tabular}
\end{center}
The values are calculated analytically, see supplementary material for implementation.
Let us now consider the variance of the values, which can be rather deceptive. On the surface, by looking at the table, it appears that a single state is related to a potentially large set of histories that vary in history values; biased or not, it \textit{seems} that state values have much less variance.
However, we are interested in the policy gradient for history-based policies.
If we employ a state-based critic, what is used to update a history-based policy is a distribution of state values in accordance to $\rho(s\mid h)$. 
To see such an update for a particular history, we next show values for a particular history in a table — the two tables below each focus on a particular history $h=lxlx$ and $h=\varnothing$ respectively, taken from multiple episodes.
Based on the distribution $\rho(s|h)$ ($\rho(x|lxlx) \approx 0.99$, $\rho(y|lxlx)\approx 0.001$), we might find our rollout look like the following table if we only show timesteps for which $h=lxlx$
\begin{center}
\begin{tabular}{c|c|c|c|c}
episode & $h$ & $s$ & $Q(s,a=or)$ & $Q(h,a=or)$ \\ \hline 
$1$ & $l-hl-l-hl$ &$l$ &$20$ & $13.9$ \\ \hline
$3$ & $l-hl-l-hl$ &$l$ &$20$ & $13.9$ \\ \hline
$8$ & $l-hl-l-hl$ &$l$ &$20$ &$13.9$ \\  \hline
$29$ & $l-hl-l-hl$ &$l$ &$20$ &$13.9$ \\  \hline
$30$ & $l-hl-l-hl$ &$l$ &$20$ &$13.9$ \\  \hline
$33$ & $l-hl-l-hl$ &$l$ &$20$ &$13.9$ \\  \hline
$60$ & $l-hl-l-hl$ &$r$ &$-50$ &$13.9$ \\ \hline
$89$ & $l-hl-l-hl$ &$l$ &$20$ &$13.9$ \\  \hline
$91$ & $l-hl-l-hl$ &$l$ &$20$ &$13.9$
\end{tabular}
\end{center}

We see that the history-value column is constant but the state-value column now varies. Because each table only looks at a specific history $\varnothing$ and $lxlx$, it is unsurprising that the history value remains constant in each table and thus have no variance. It is essentially the reason why one cannot obtain values of smaller variance than history values, because true history values essentially have no additional variance for a given history in policy update. 

\clearpage
\section{OpenAI Particle Environments with Partial Observability}

\subsection{Speaker and Listener}

\begin{figure}[h]
  \centering
  \captionsetup[subfigure]{labelformat=empty}
  \subcaptionbox{}
      [0.31\linewidth]{\includegraphics[height=3.4cm]{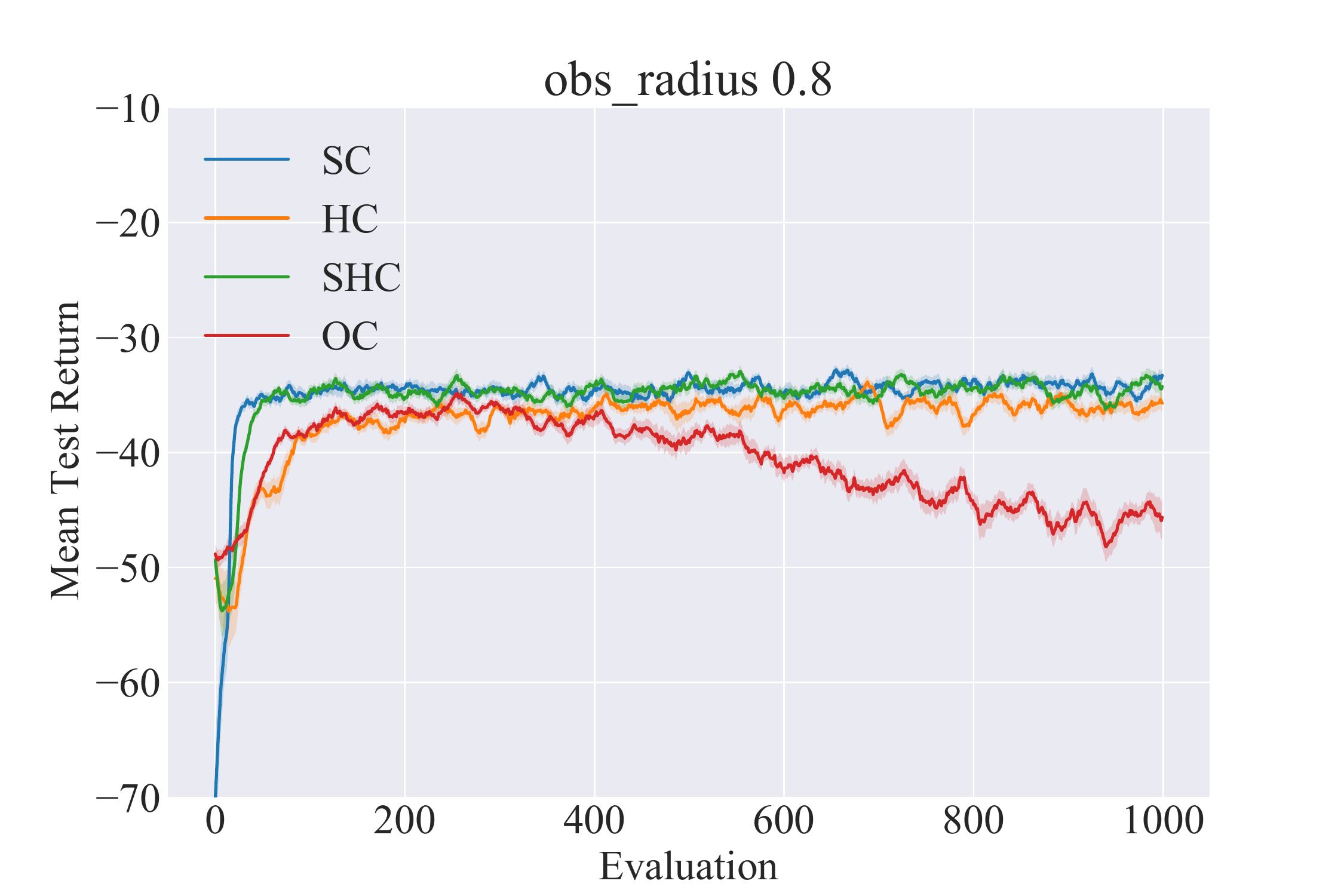}}
  ~
  \centering
  \subcaptionbox{}
      [0.31\linewidth]{\includegraphics[height=3.4cm]{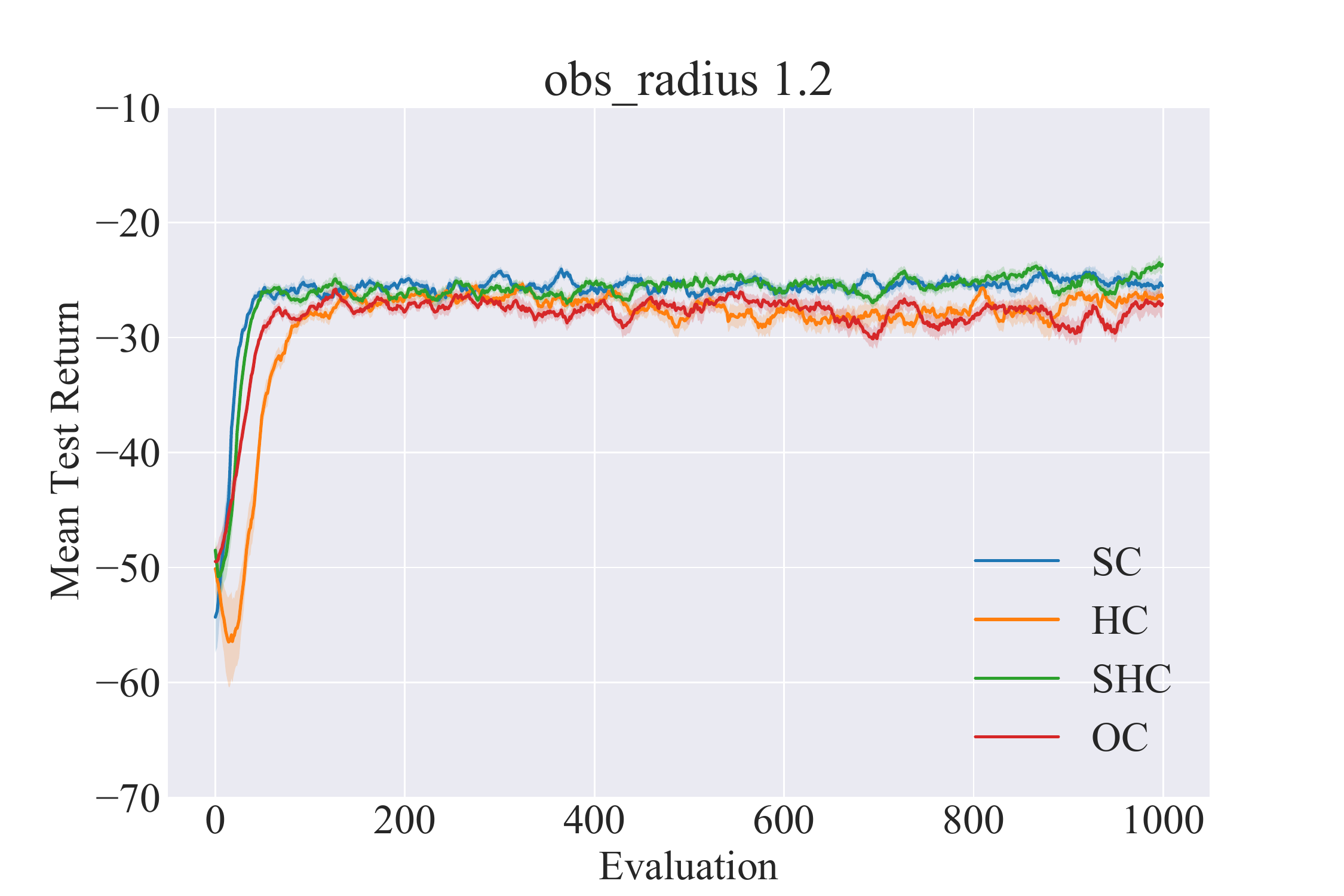}}
  ~
  \centering
  \subcaptionbox{}
      [0.31\linewidth]{\includegraphics[height=3.4cm]{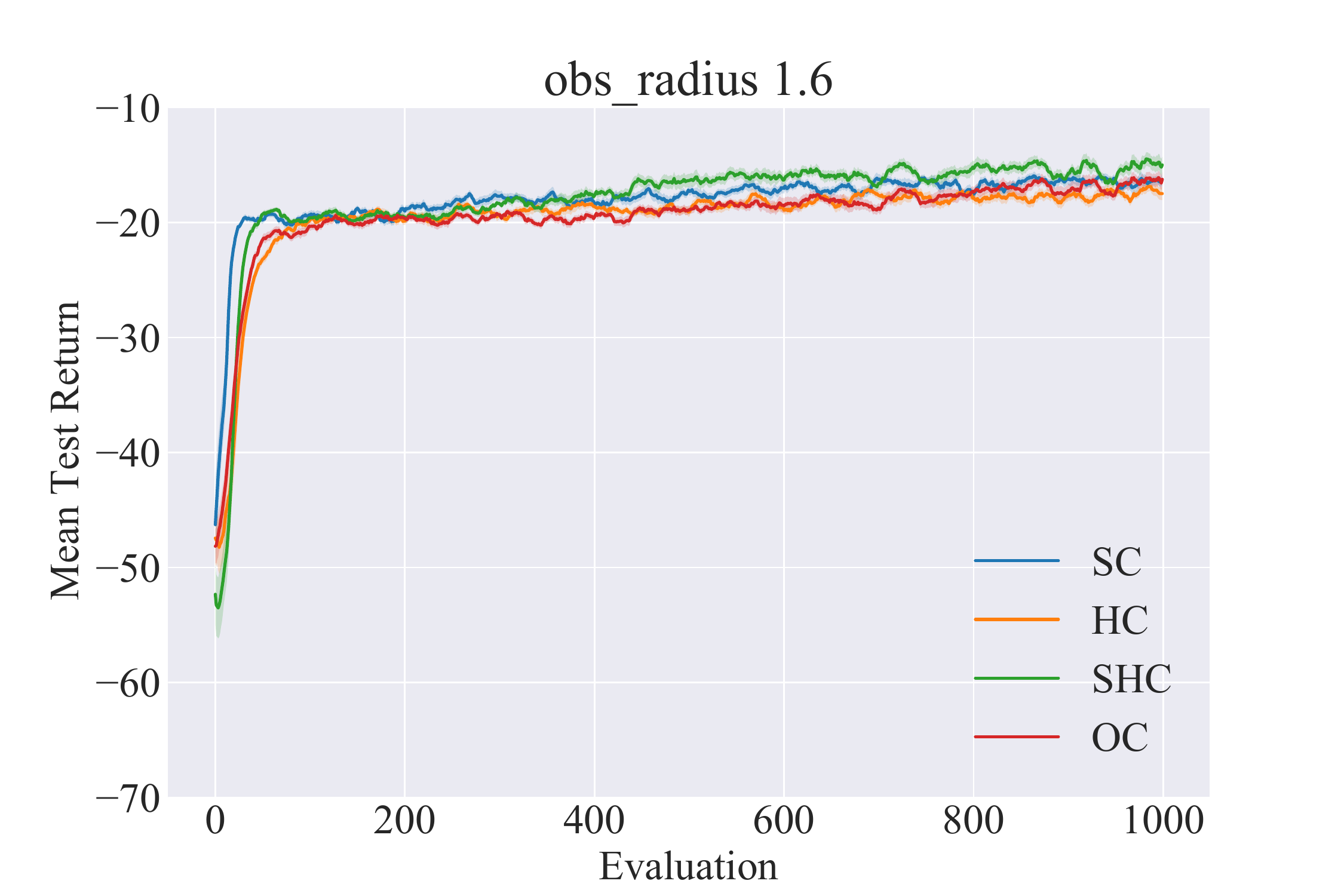}}
   
  \centering
  \subcaptionbox{}
      [0.31\linewidth]{\includegraphics[height=3.4cm]{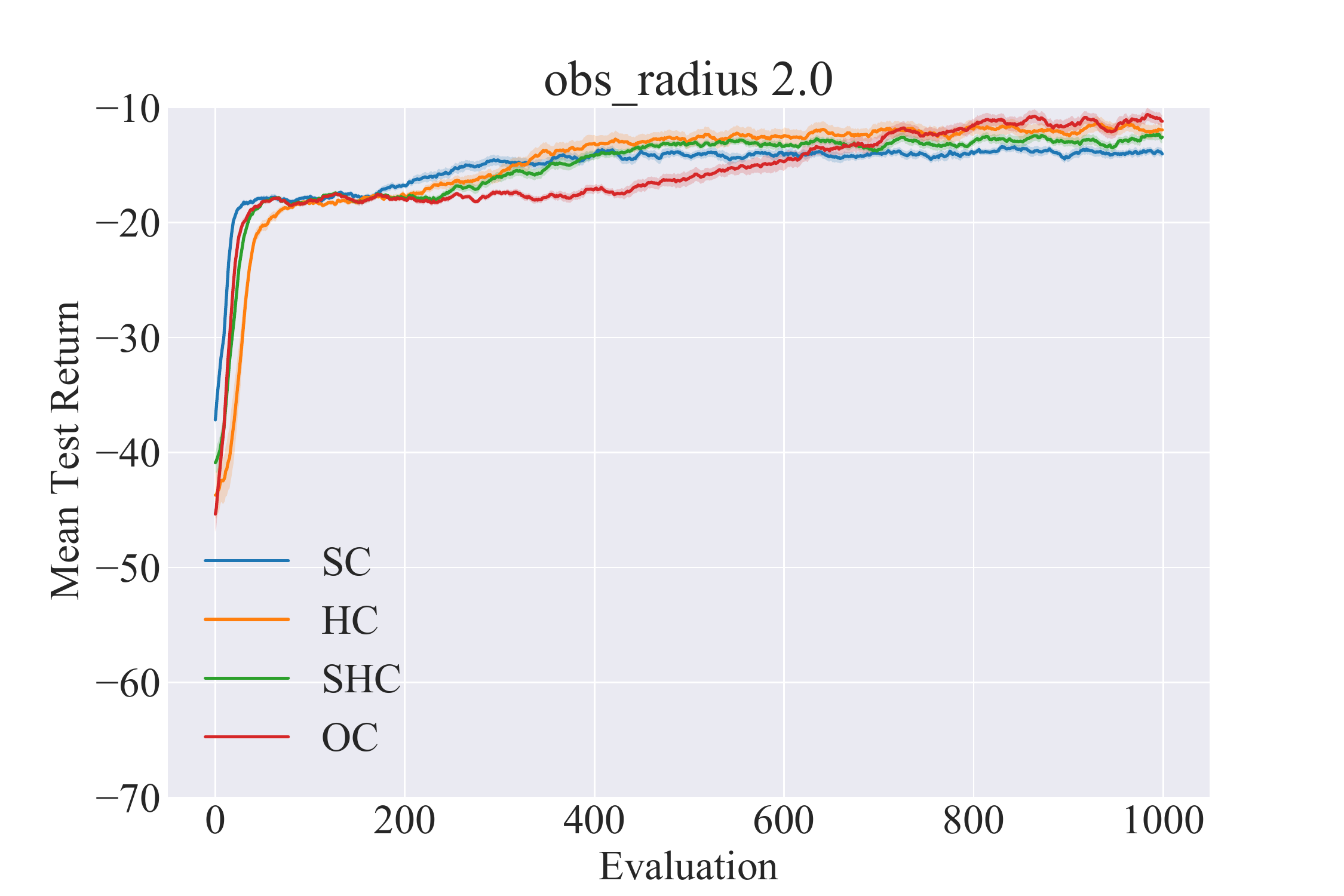}}
  ~
  \centering
  \subcaptionbox{}
      [0.31\linewidth]{\includegraphics[height=3.4cm]{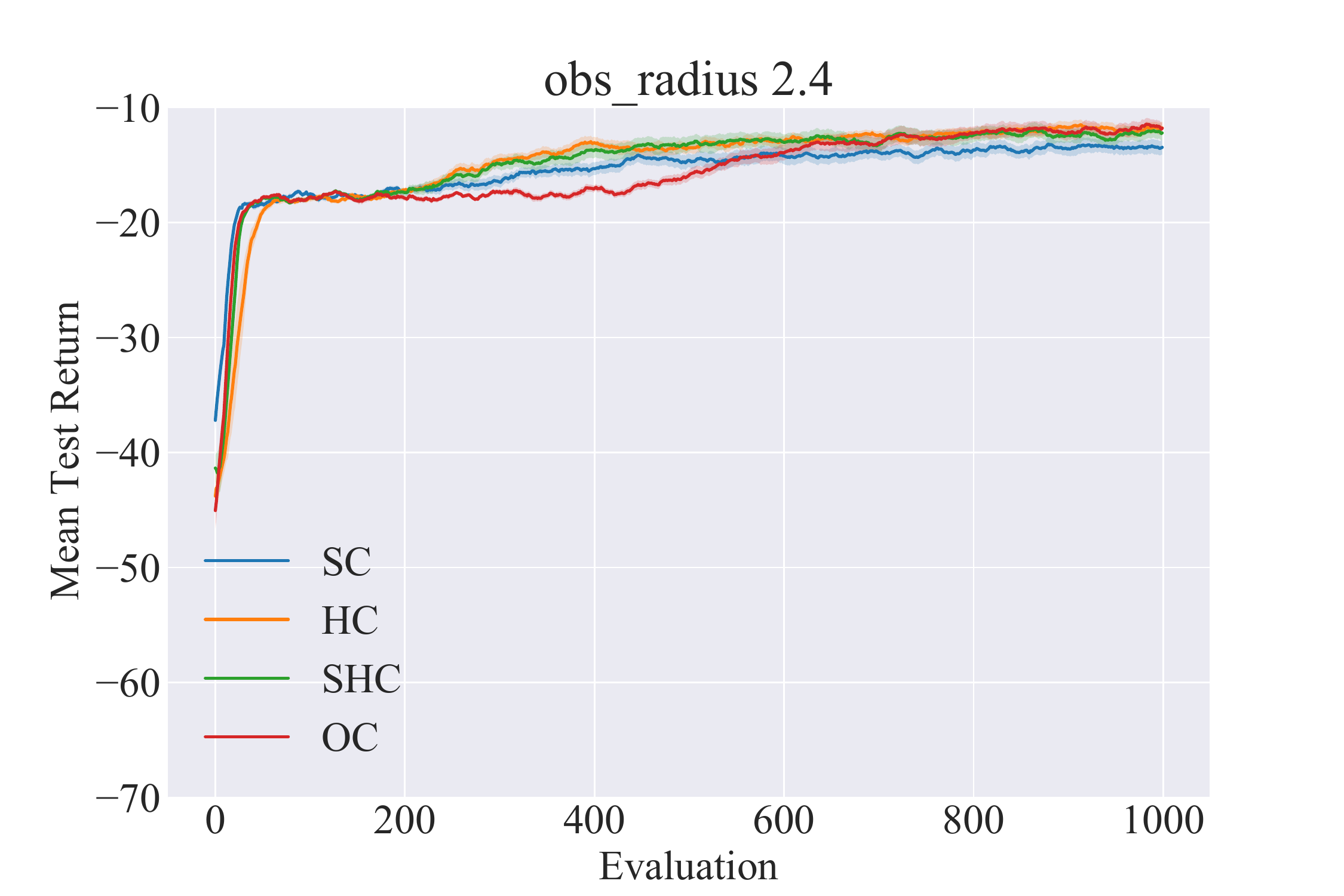}}
  \caption{Performance comparison in Speaker and Listener.}
  \label{fig:speaker_and_listener}
\end{figure}

\subsection{Predator and Prey}

\begin{figure}[h!]
     \centering
     \captionsetup[subfigure]{labelformat=empty}
     \subcaptionbox{}
         [0.4\linewidth]{\includegraphics[height=3.3cm]{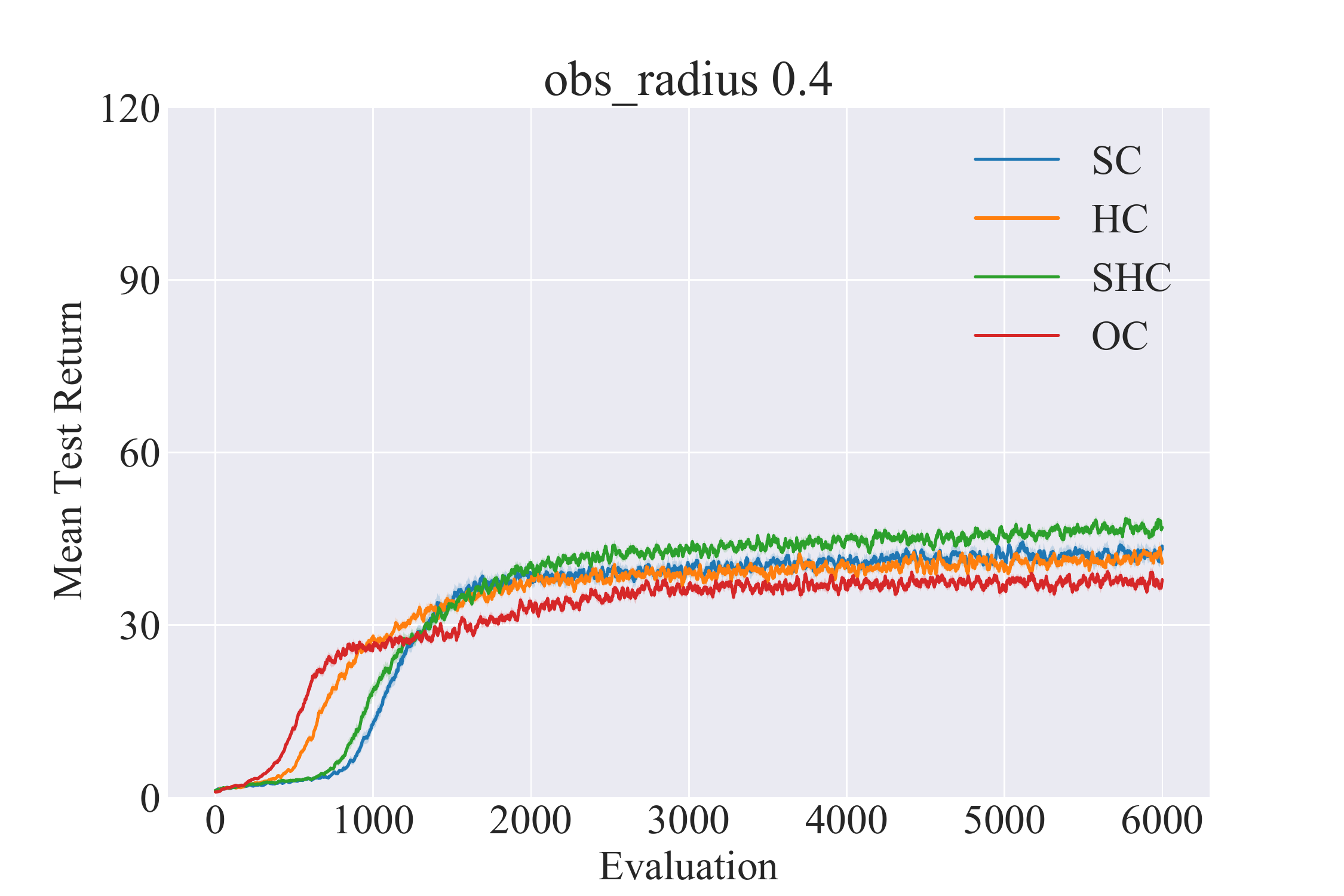}}
     ~
     \centering
     \subcaptionbox{}
         [0.4\linewidth]{\includegraphics[height=3.3cm]{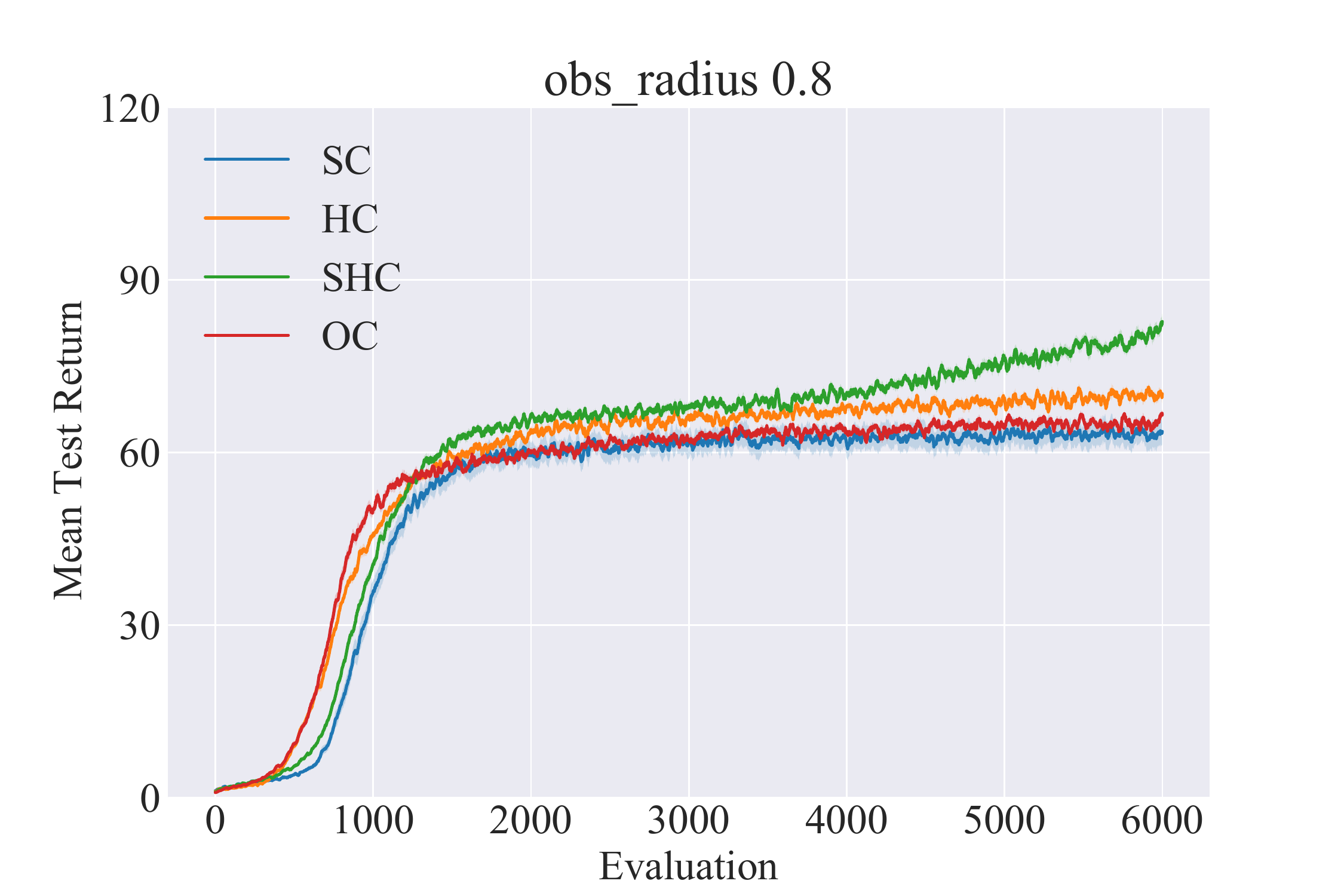}}
     ~
     \centering
     \subcaptionbox{}
         [0.4\linewidth]{\includegraphics[height=3.3cm]{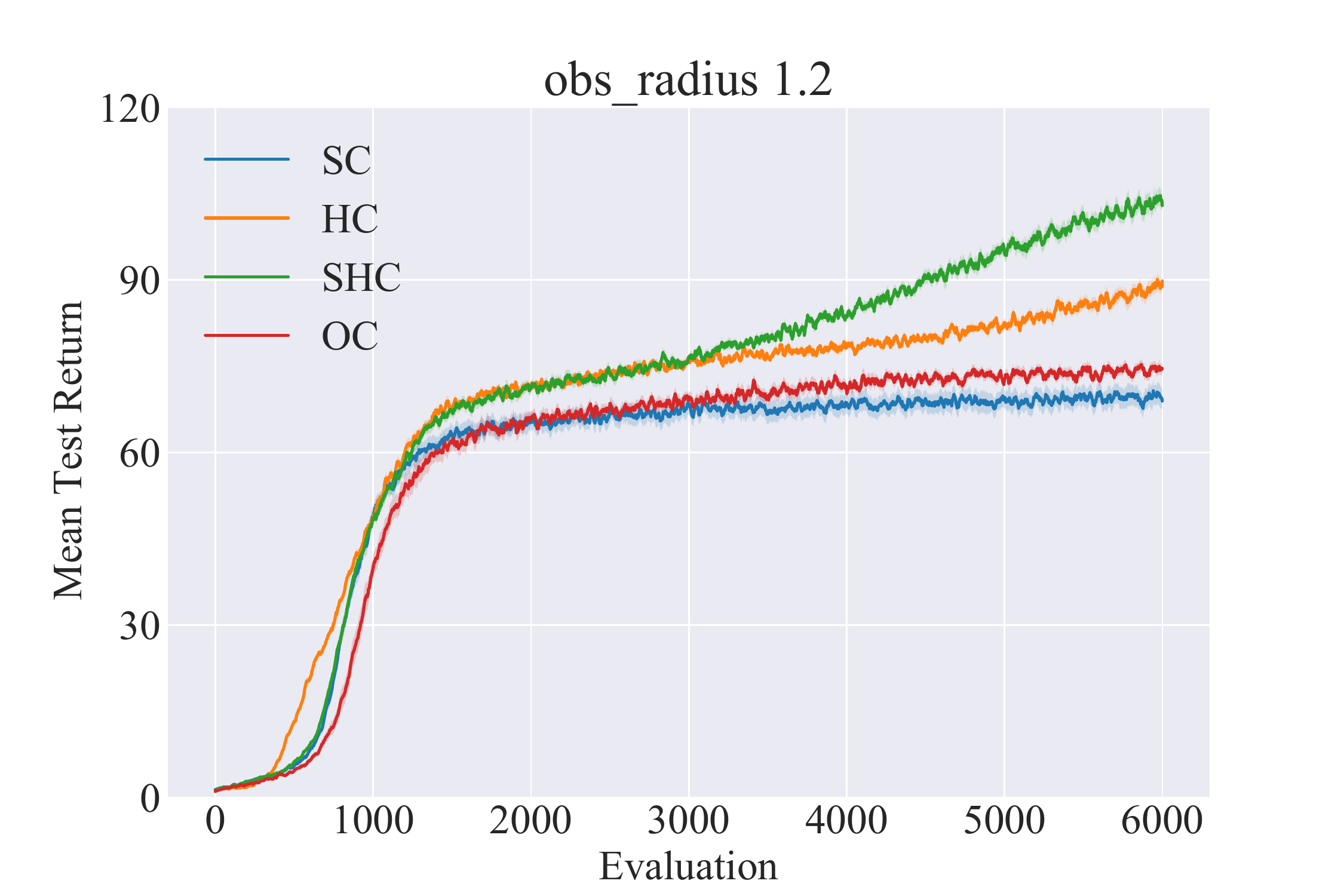}}
     ~
     \centering
     \subcaptionbox{}
         [0.4\linewidth]{\includegraphics[height=3.3cm]{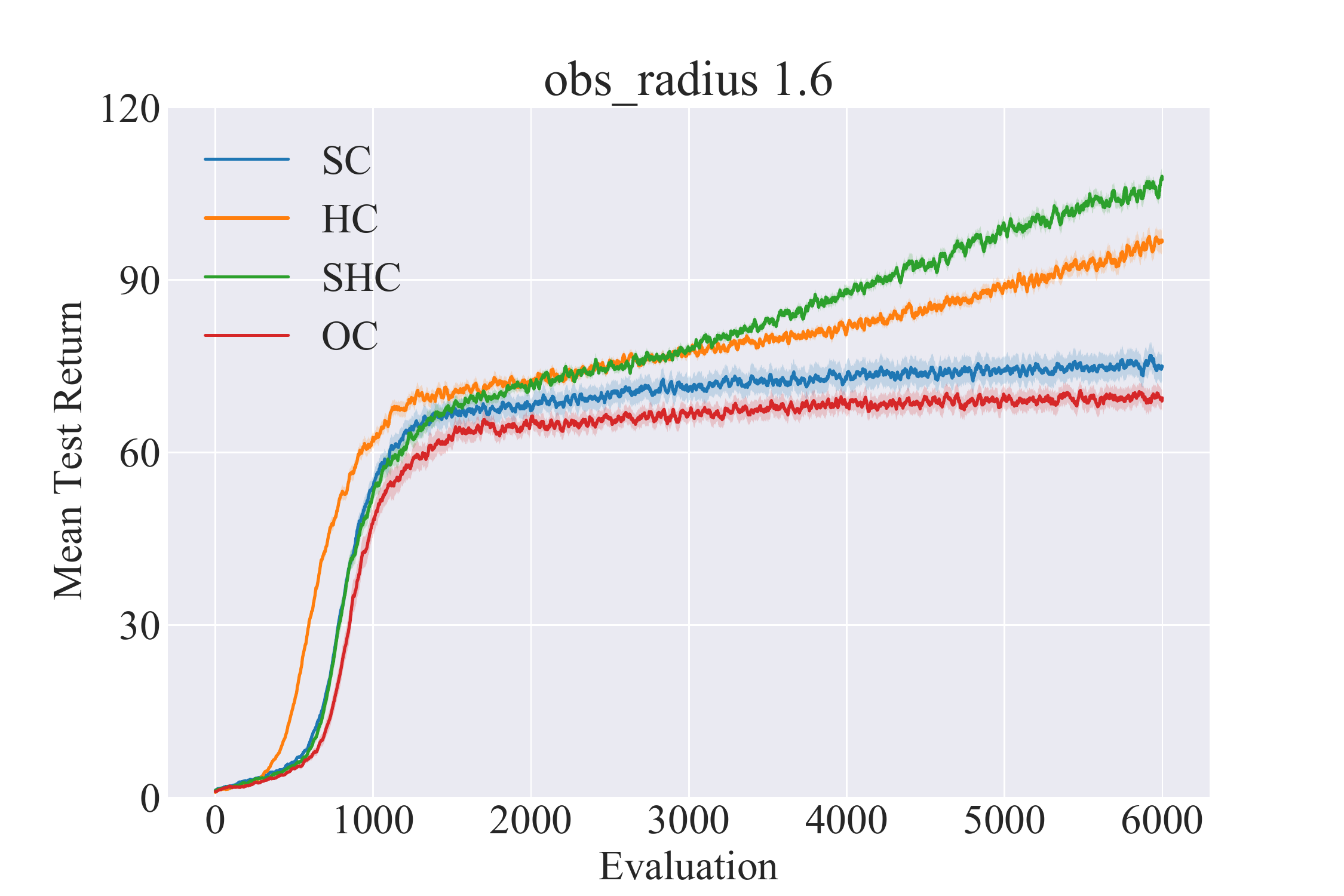}}
     \caption{Performance comparison in Predator and Prey.}
     \label{fig:predator_and_prey}
\end{figure}

\clearpage
\subsection{Cooperative Navigation}

\begin{figure}[h]
    \centering
    \captionsetup[subfigure]{labelformat=empty}
    \subcaptionbox{}
        [0.4\linewidth]{\includegraphics[height=3.5cm]{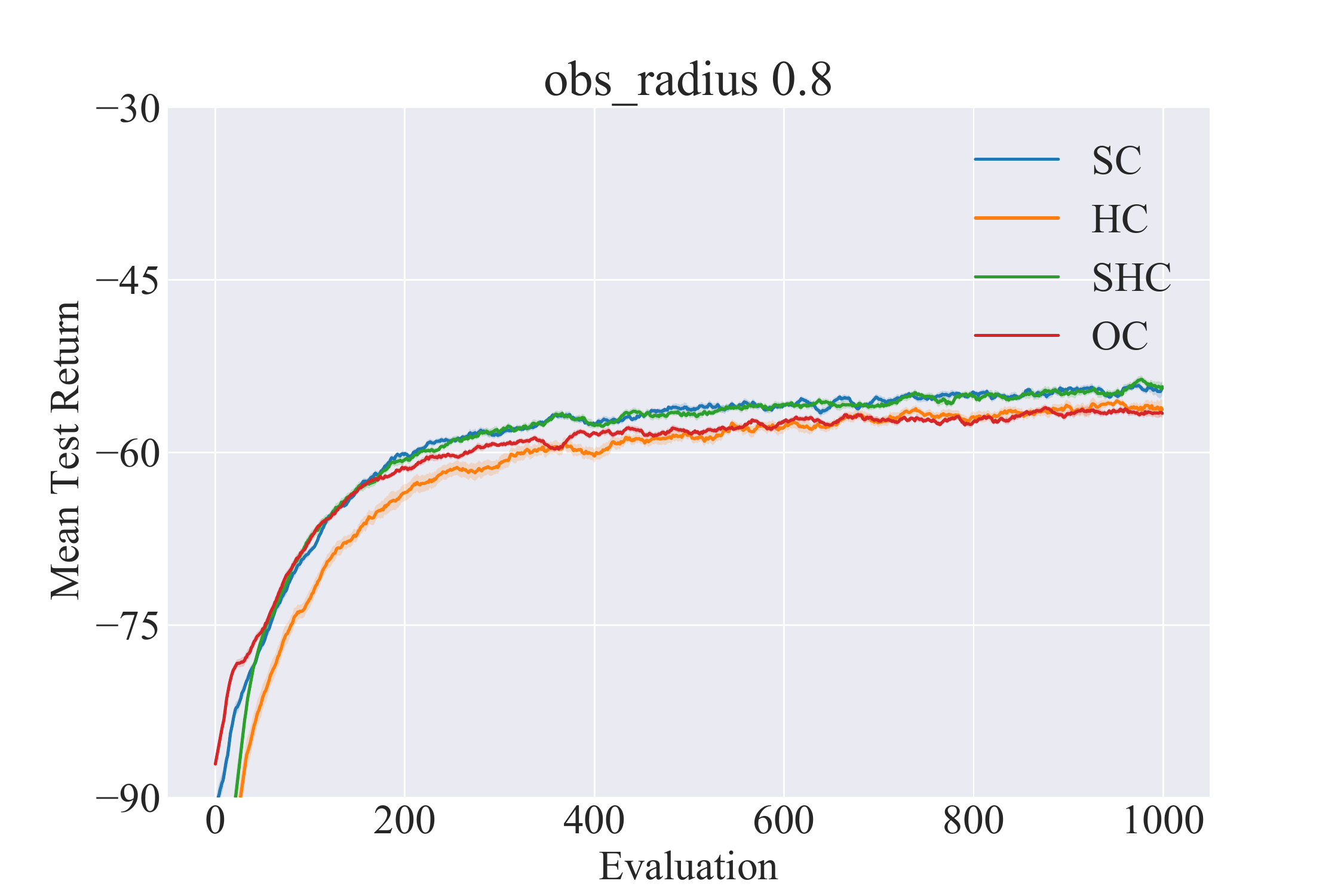}}
    ~
    \centering
    \subcaptionbox{}
        [0.4\linewidth]{\includegraphics[height=3.5cm]{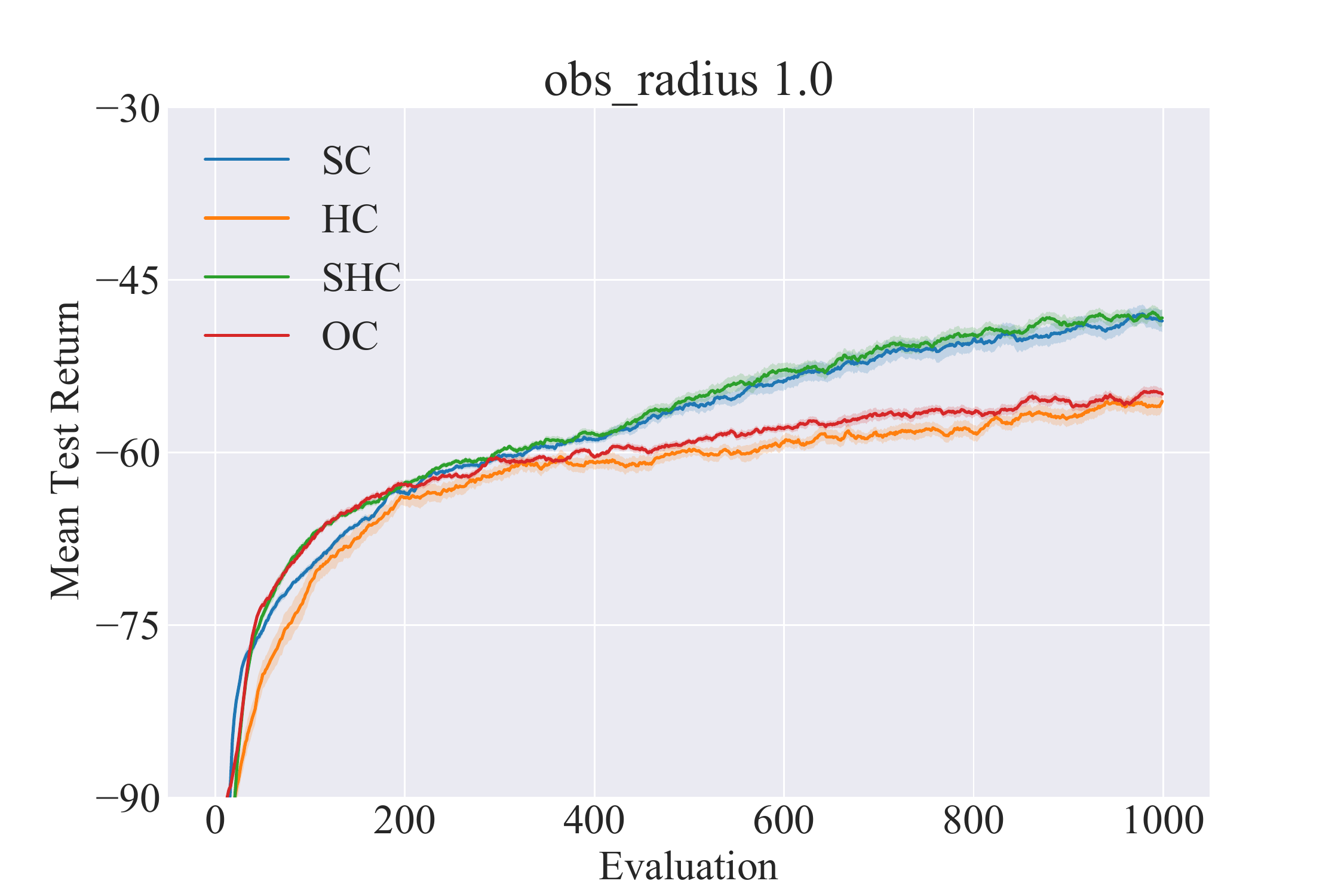}}
    ~
    \centering
    \subcaptionbox{}
        [0.4\linewidth]{\includegraphics[height=3.5cm]{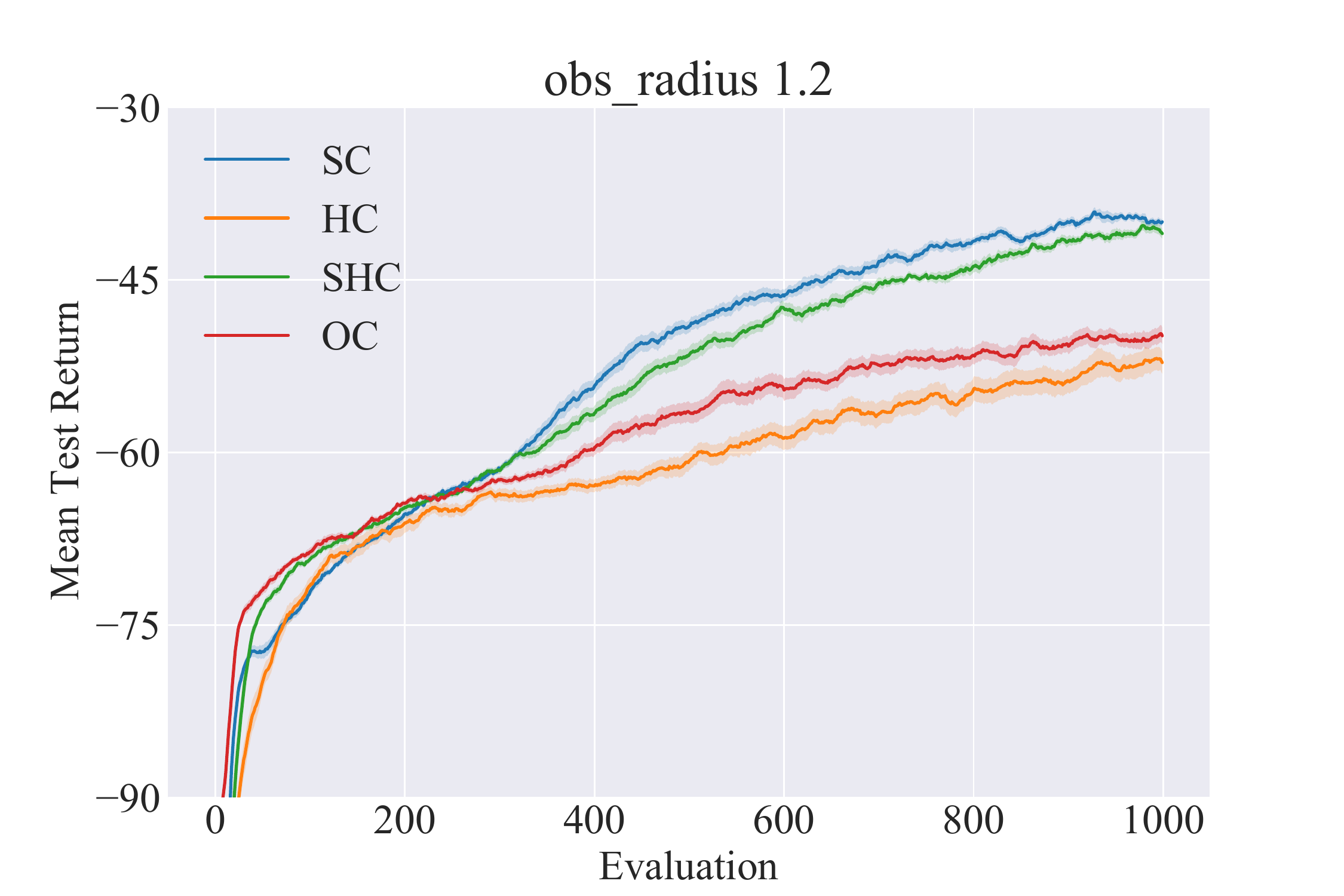}}
    ~
    \centering
    \subcaptionbox{}
        [0.4\linewidth]{\includegraphics[height=3.5cm]{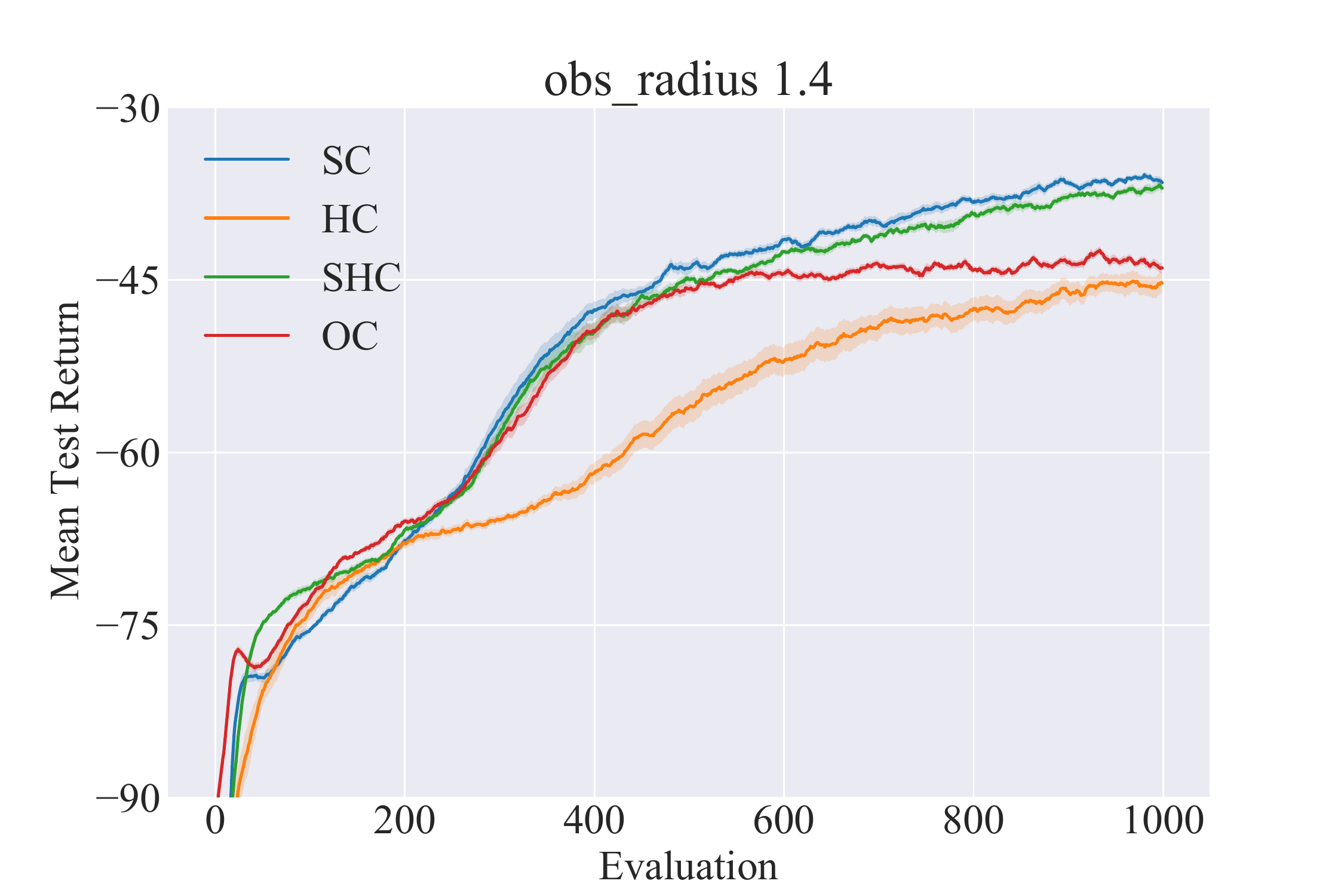}}
    \caption{Performance comparison in Cooperative Navigation.}
    \label{fig:cooperative_navigation}
\end{figure}

\subsection{Meeting-in-a-Grid Small}

\begin{figure}[ht!]
    \centering
    \includegraphics[width=0.4\textwidth]{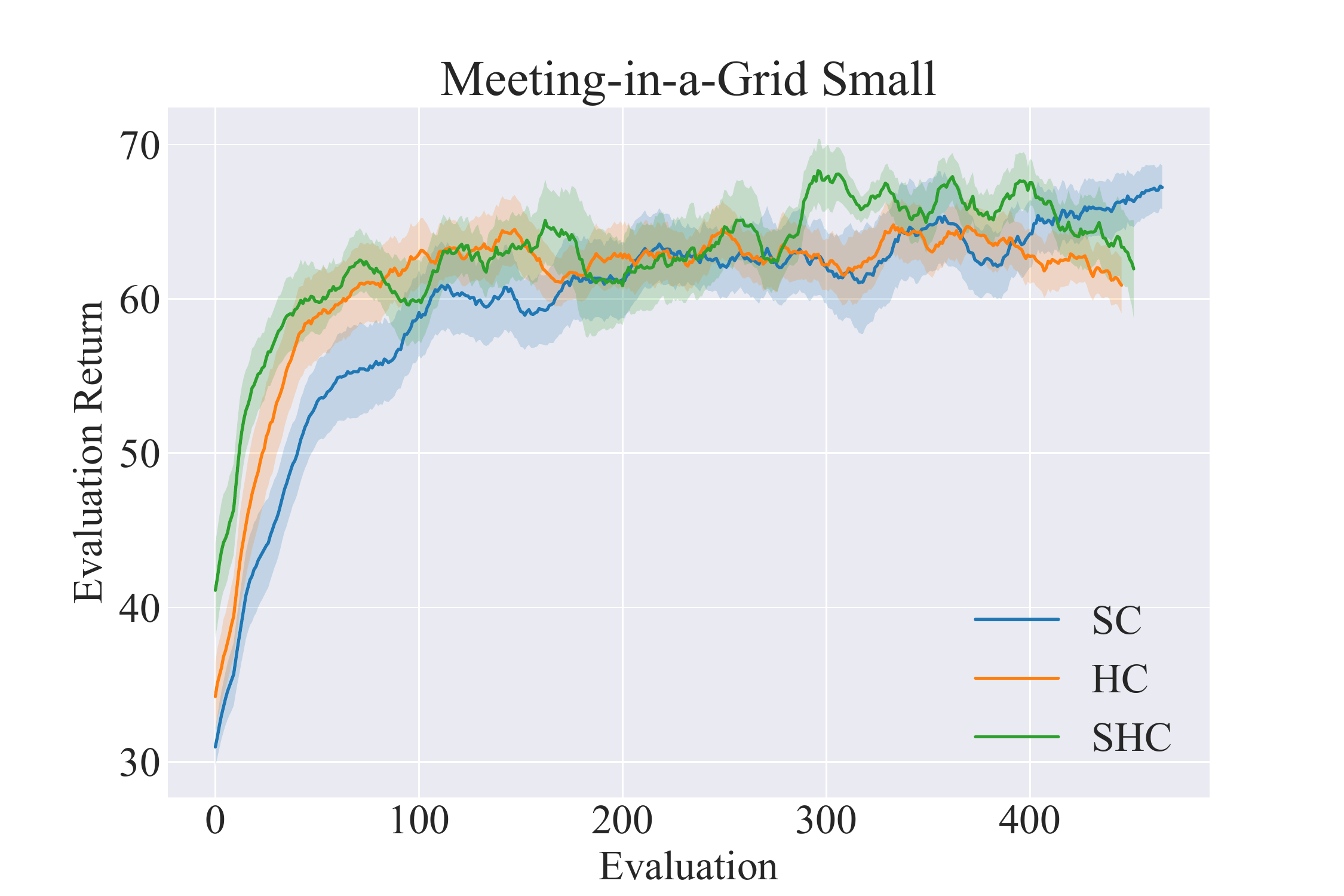}
    \caption{Meeting-in-a-Grid-Small environment~\cite{bernstein2005bounded}}
    \label{fig:grid_small}
\end{figure}

\subsection{COMA}

\begin{figure}[h]
    \centering
    \captionsetup[subfigure]{labelformat=empty}
    \subcaptionbox{}
        [0.31\linewidth]{\includegraphics[height=3.5cm]{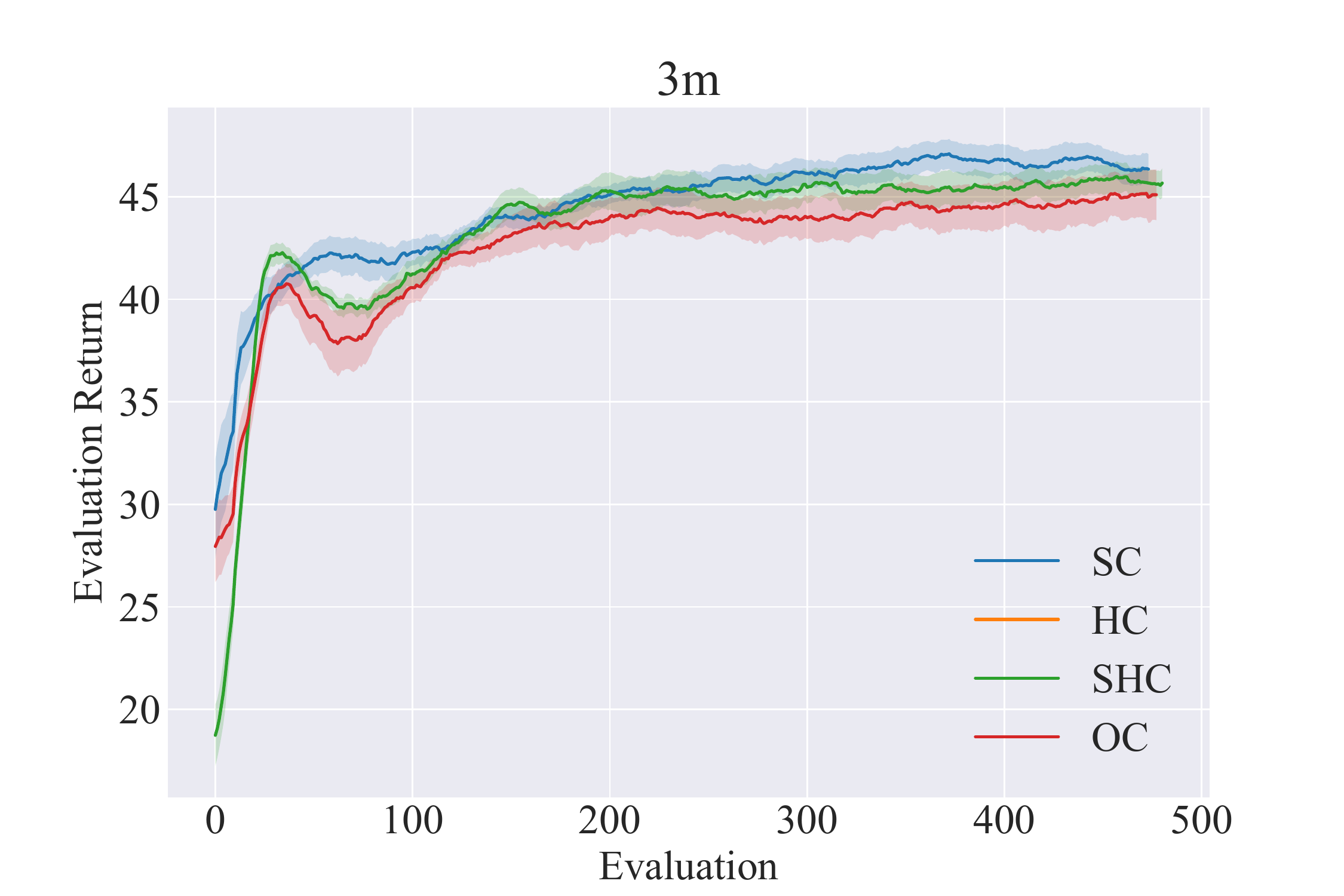}}
    ~
    \centering
    \subcaptionbox{}
        [0.31\linewidth]{\includegraphics[height=3.5cm]{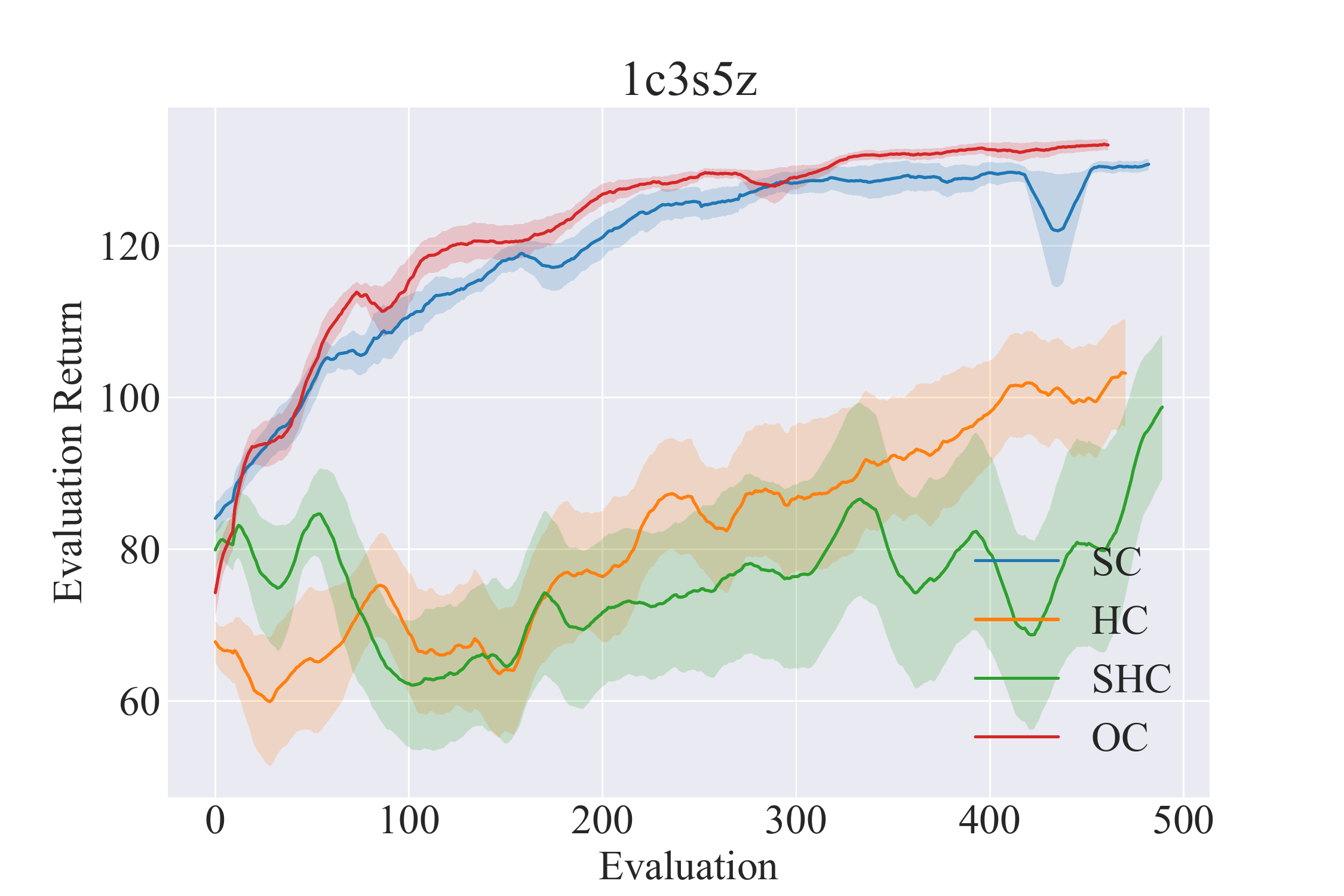}}
    ~
    \centering
    \subcaptionbox{}
        [0.31\linewidth]{\includegraphics[height=3.5cm]{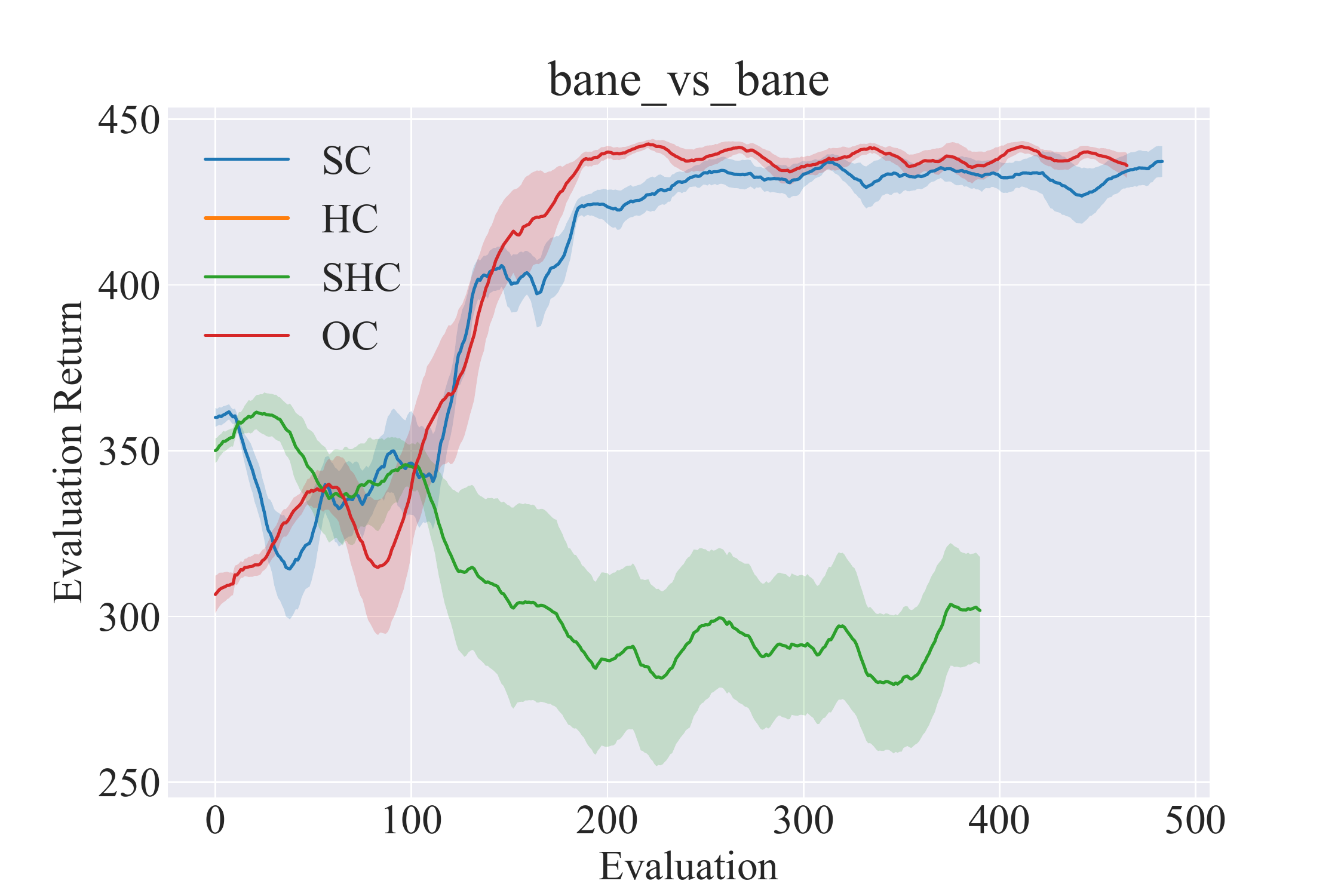}}
    \caption{Performance comparison in scenarios of SMAC using various critics, sum of reward is plotted.}
    \label{fig:smac_coma}
  \end{figure}

\clearpage
\subsection{Reactive Policies}\label{reactive_policy_figures}

\begin{figure}[h]
    \begin{subfigure}{.31\textwidth}
    \centering
      \includegraphics[width=1.1\textwidth]{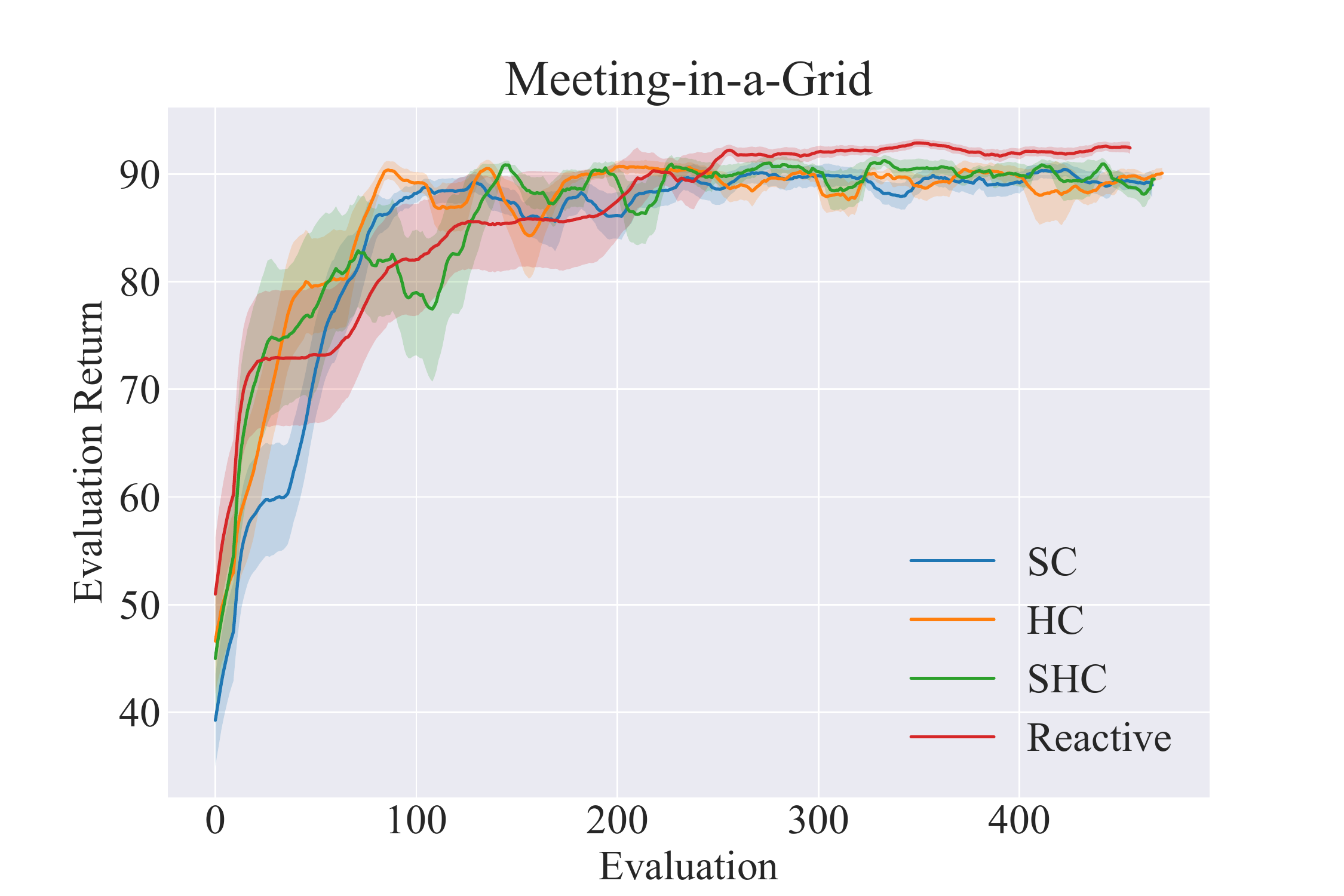}
    \end{subfigure} \hfill
    \begin{subfigure}{.31\textwidth}
      \centering
        \includegraphics[width=1.1\textwidth]{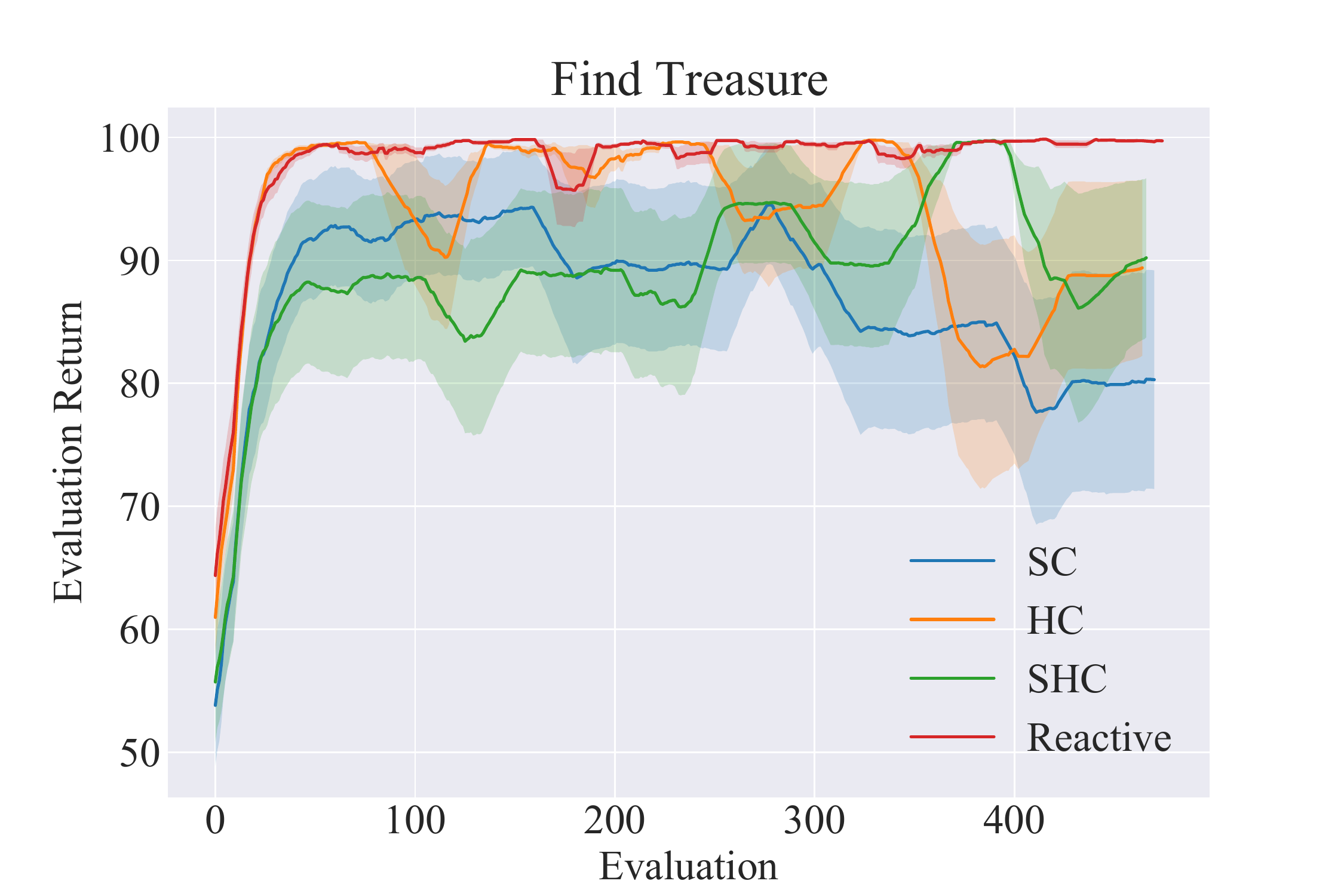}
      \end{subfigure}\hfill
    \begin{subfigure}{.31\textwidth}
    \centering
      \includegraphics[width=1.1\textwidth]{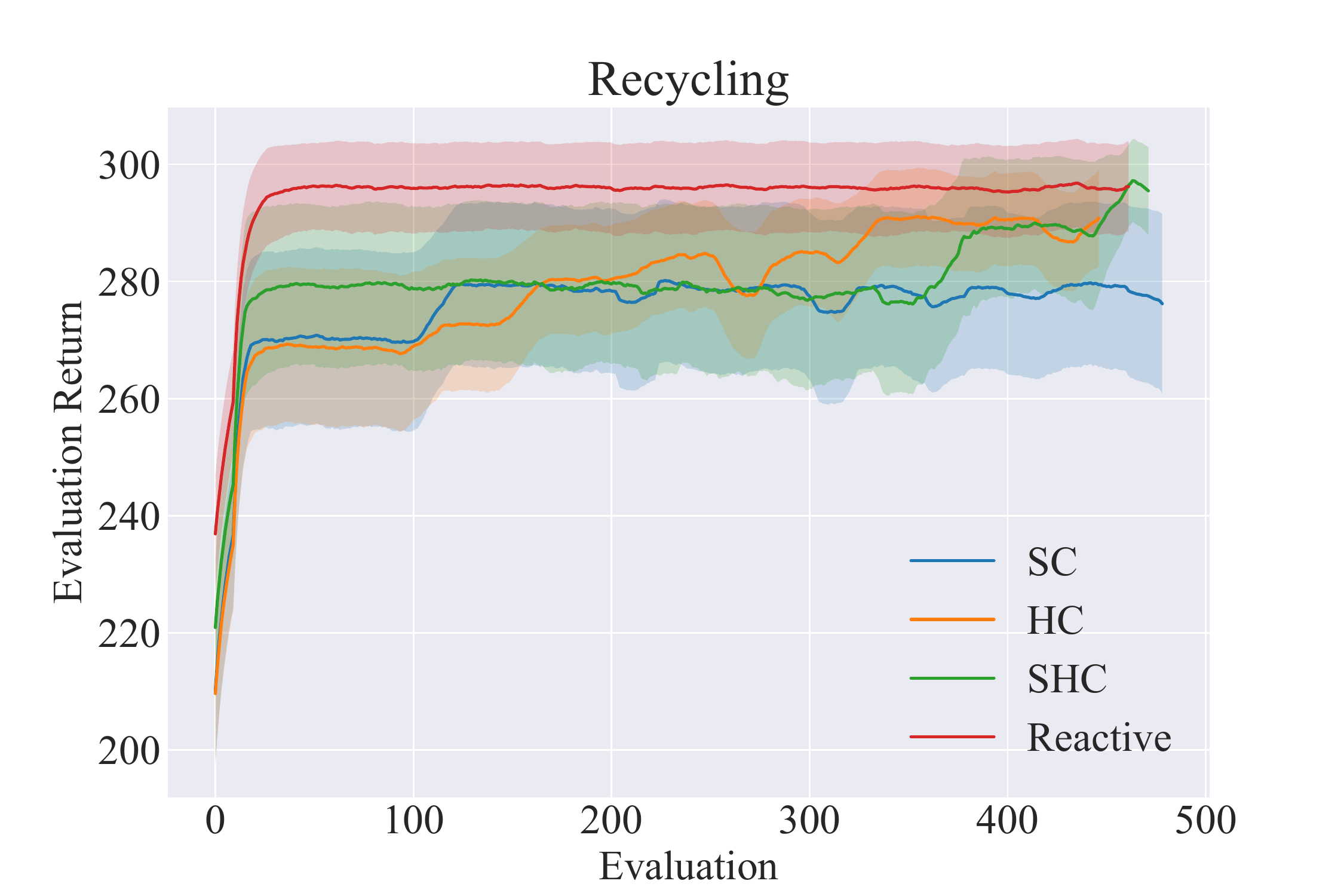}
    \end{subfigure}
    \caption{Performance evaluation including reactive policies}
    \label{fig:reactive_policies_normal}
\end{figure}
\vspace{1cm}
\clearpage

\subsection{MAAC}

\begin{figure}[h]
    \begin{subfigure}{.31\textwidth}
    \centering
      \includegraphics[width=1.1\textwidth]{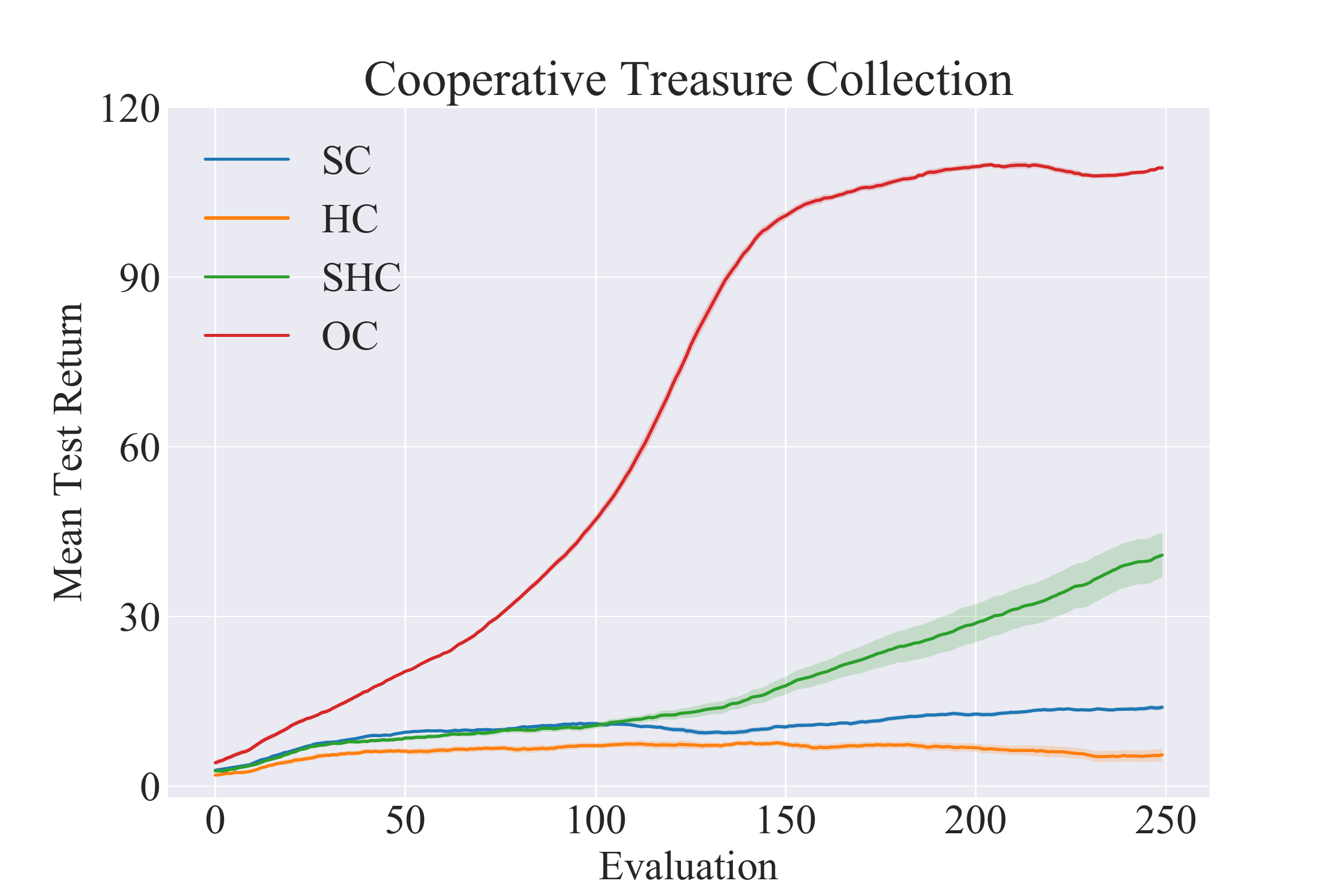}
    \end{subfigure} \hfill
    \begin{subfigure}{.31\textwidth}
      \centering
        \includegraphics[width=1.1\textwidth]{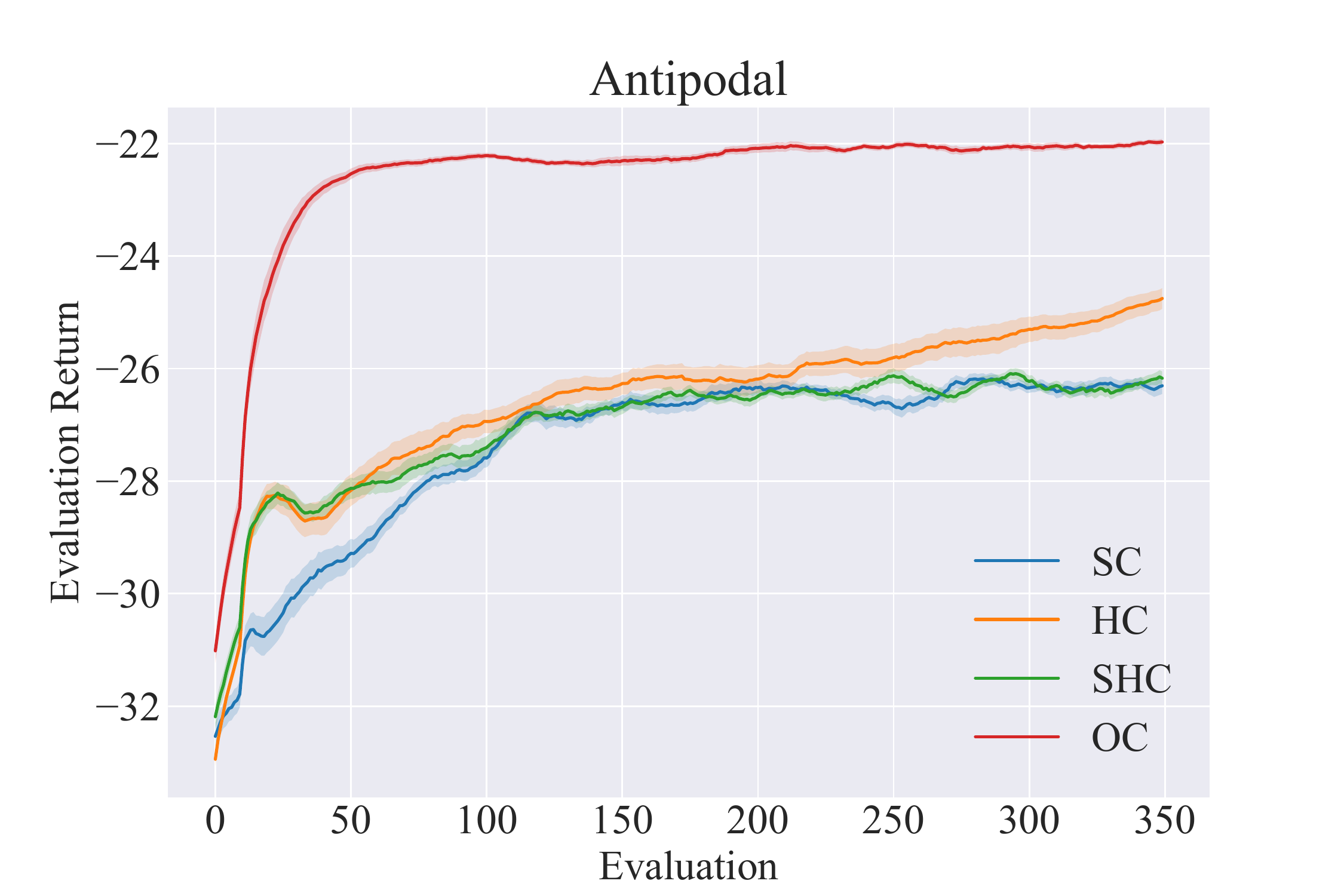}
      \end{subfigure}\hfill
    \begin{subfigure}{.31\textwidth}
    \centering
      \includegraphics[width=1.1\textwidth]{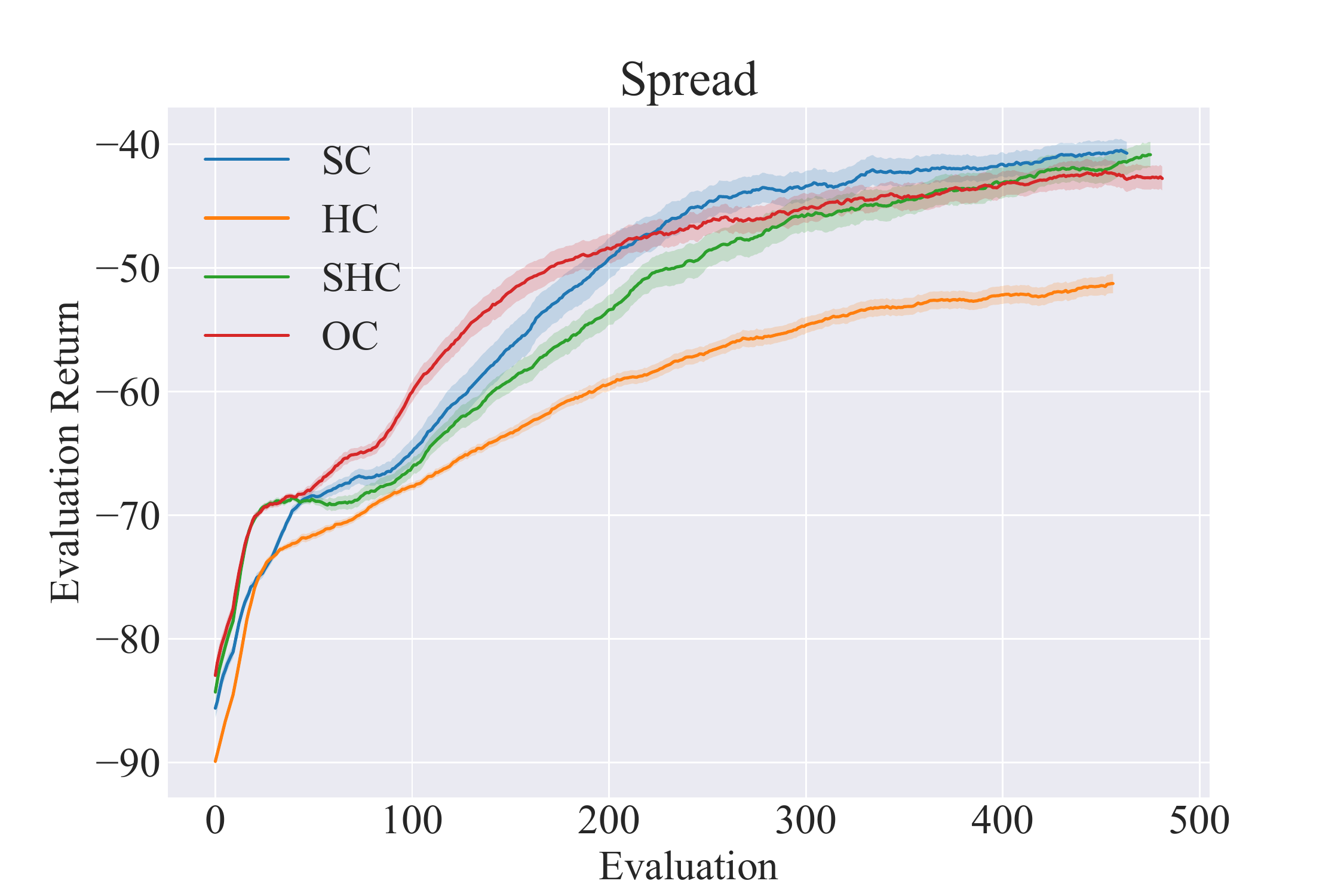}
    \end{subfigure}
    \begin{subfigure}{.31\textwidth}
        \centering
          \includegraphics[width=1.1\textwidth]{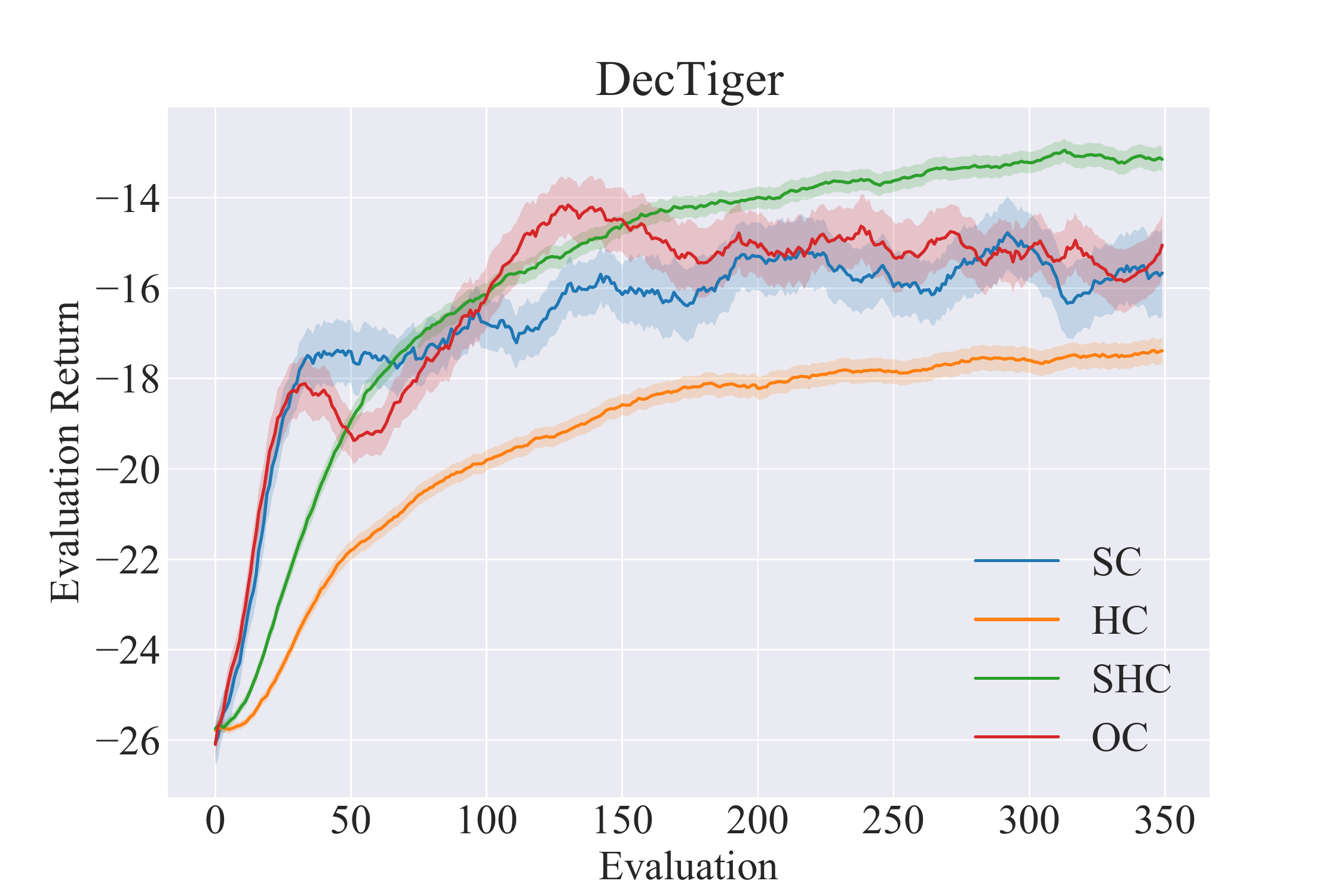}
        \end{subfigure}
    \caption{Performance evaluation using MAAC with different types of critics}
\end{figure}

\begin{figure}[h]
  \centering
  \captionsetup[subfigure]{labelformat=empty}
  \subcaptionbox{}
      [0.31\linewidth]{\includegraphics[height=3.4cm]{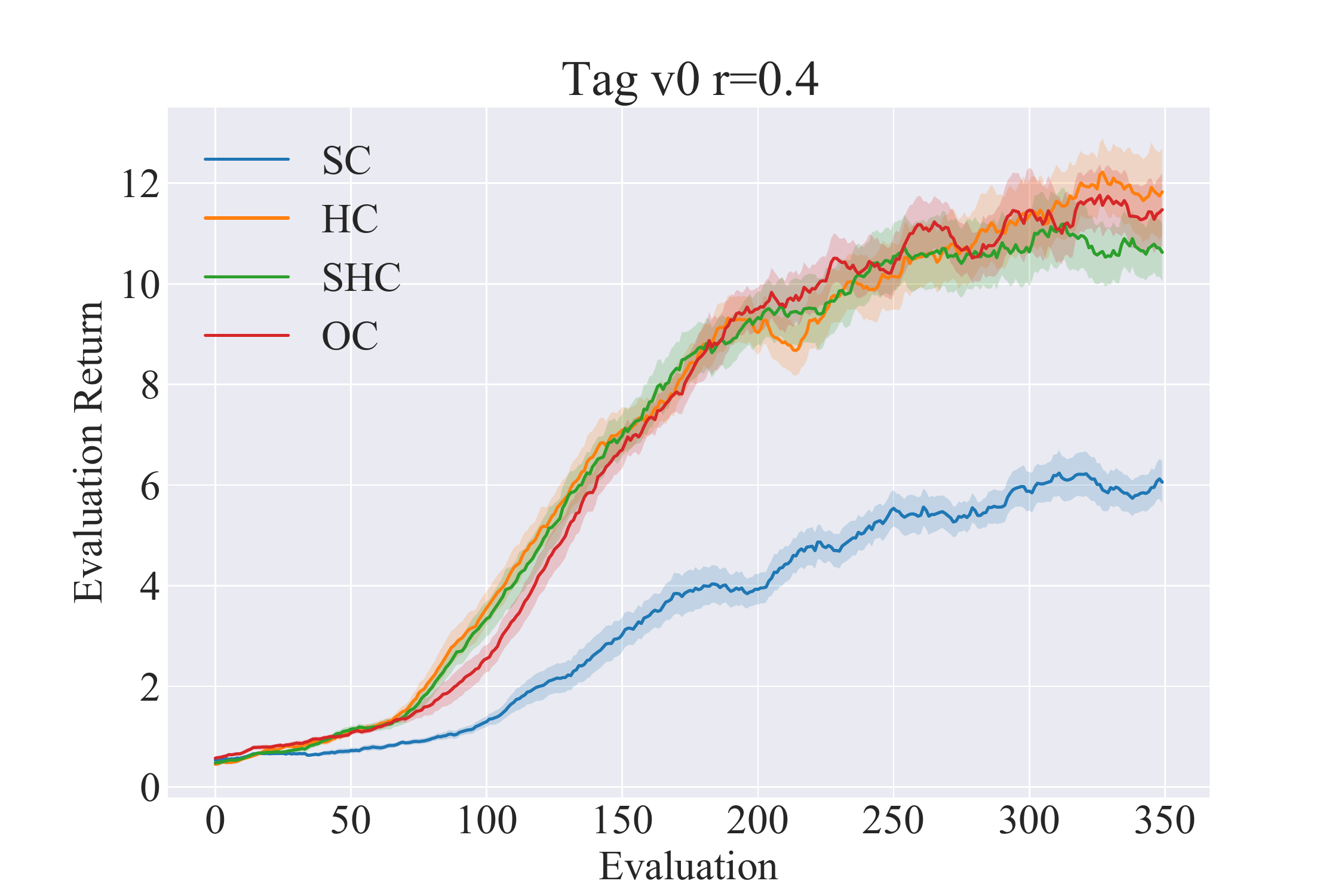}}
  ~
  \centering
  \subcaptionbox{}
      [0.31\linewidth]{\includegraphics[height=3.4cm]{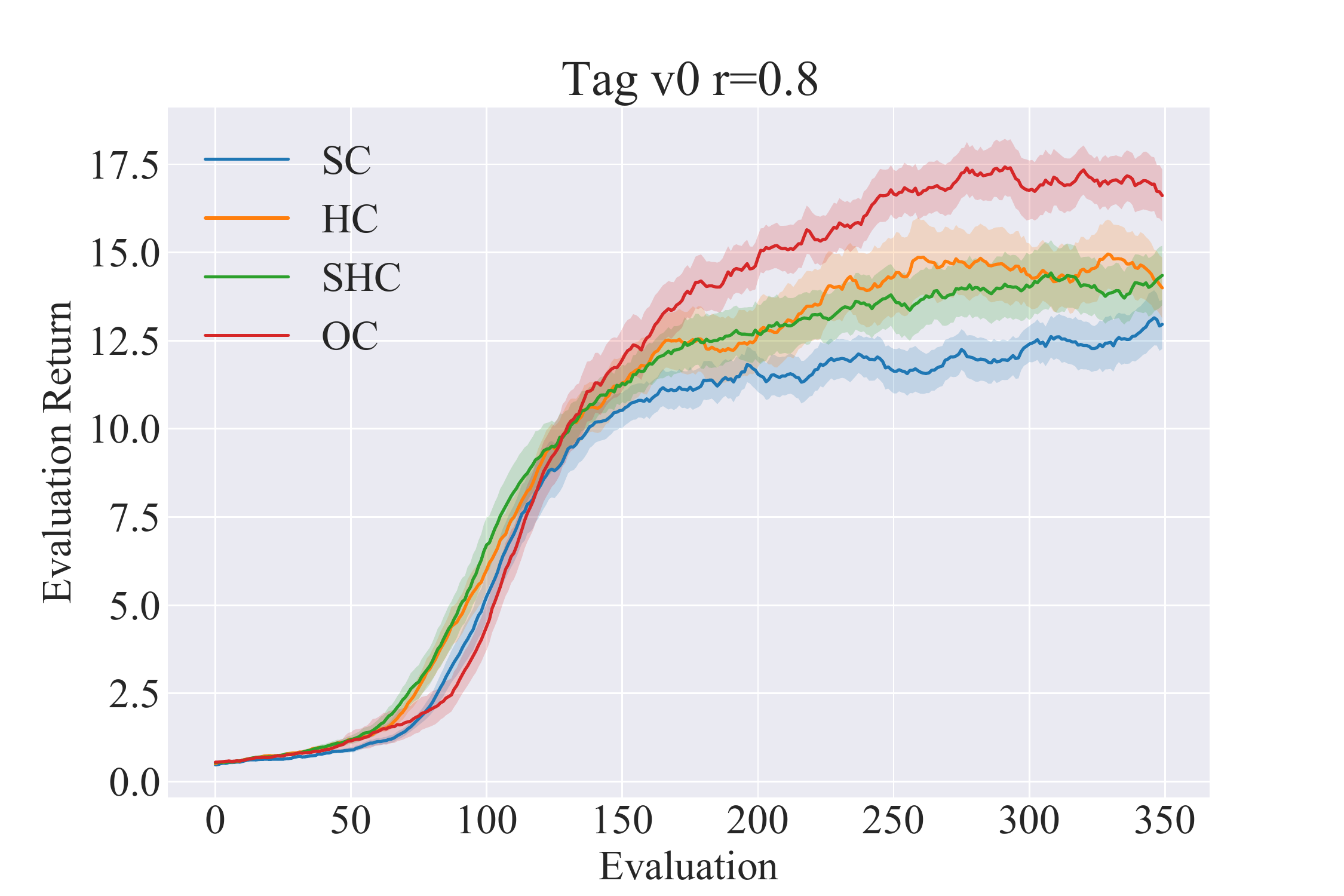}}
  ~
  \centering
  \subcaptionbox{}
      [0.31\linewidth]{\includegraphics[height=3.4cm]{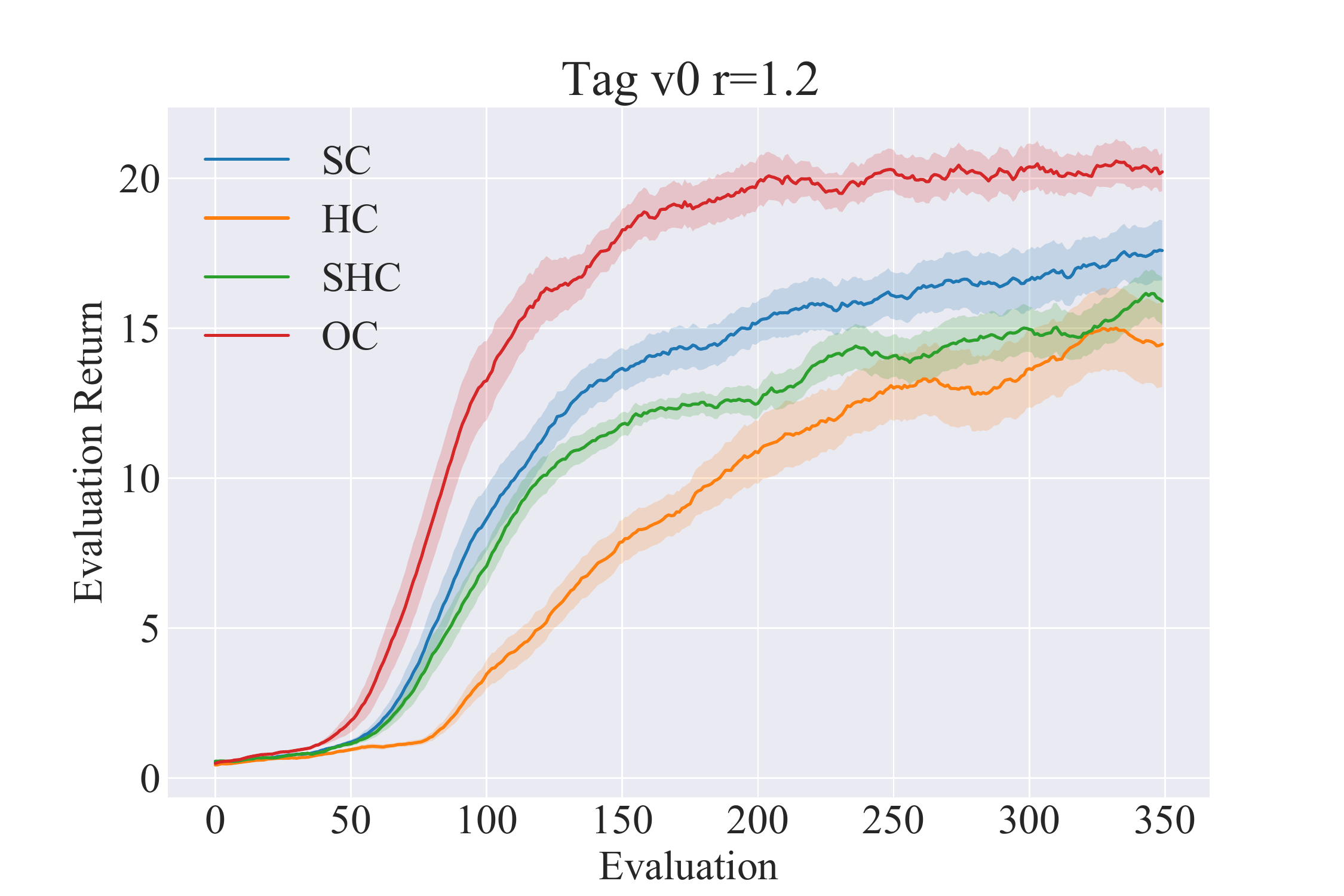}}
   
  \centering
  \subcaptionbox{}
      [0.31\linewidth]{\includegraphics[height=3.4cm]{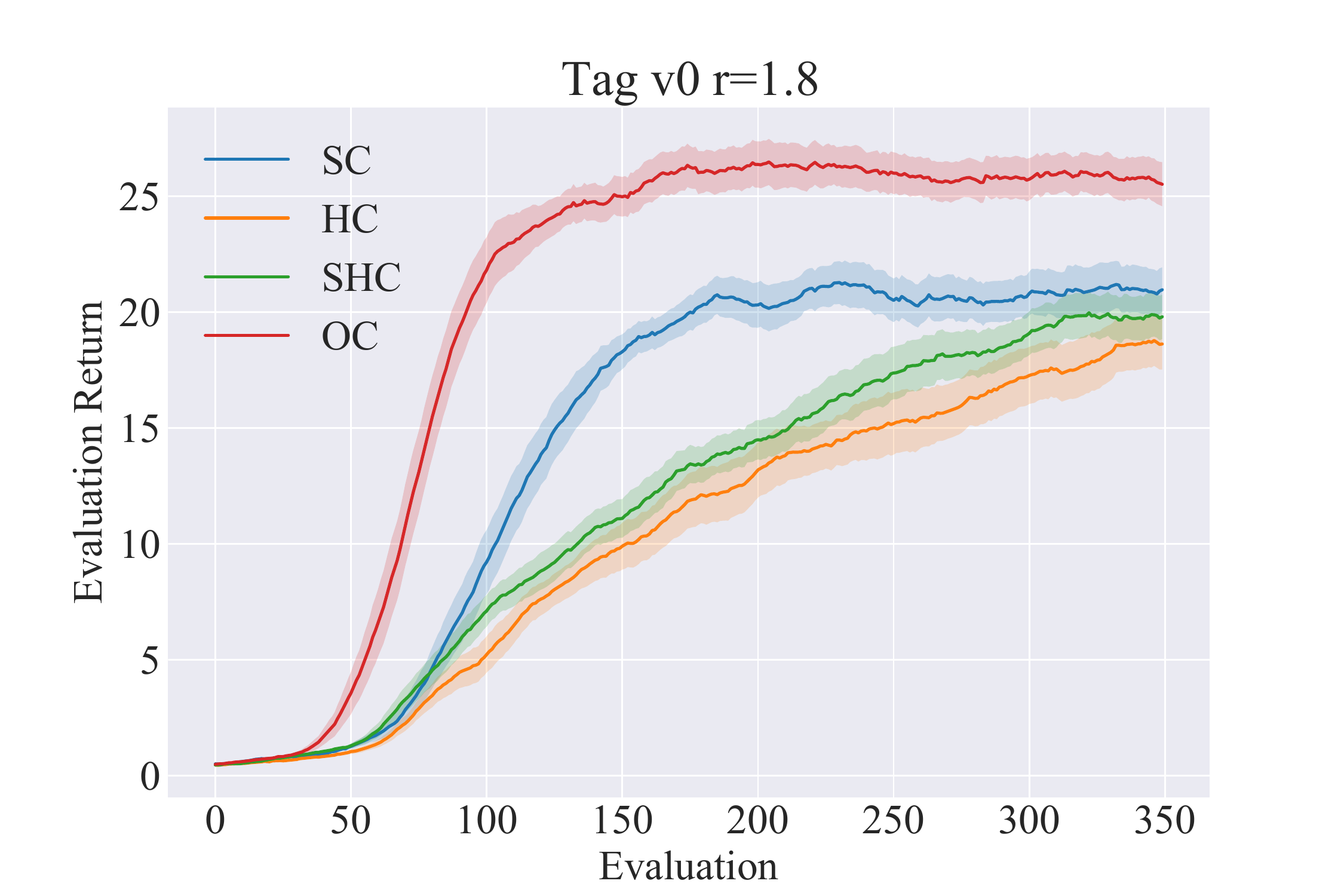}}
  ~
  \centering
  \subcaptionbox{}
      [0.31\linewidth]{\includegraphics[height=3.4cm]{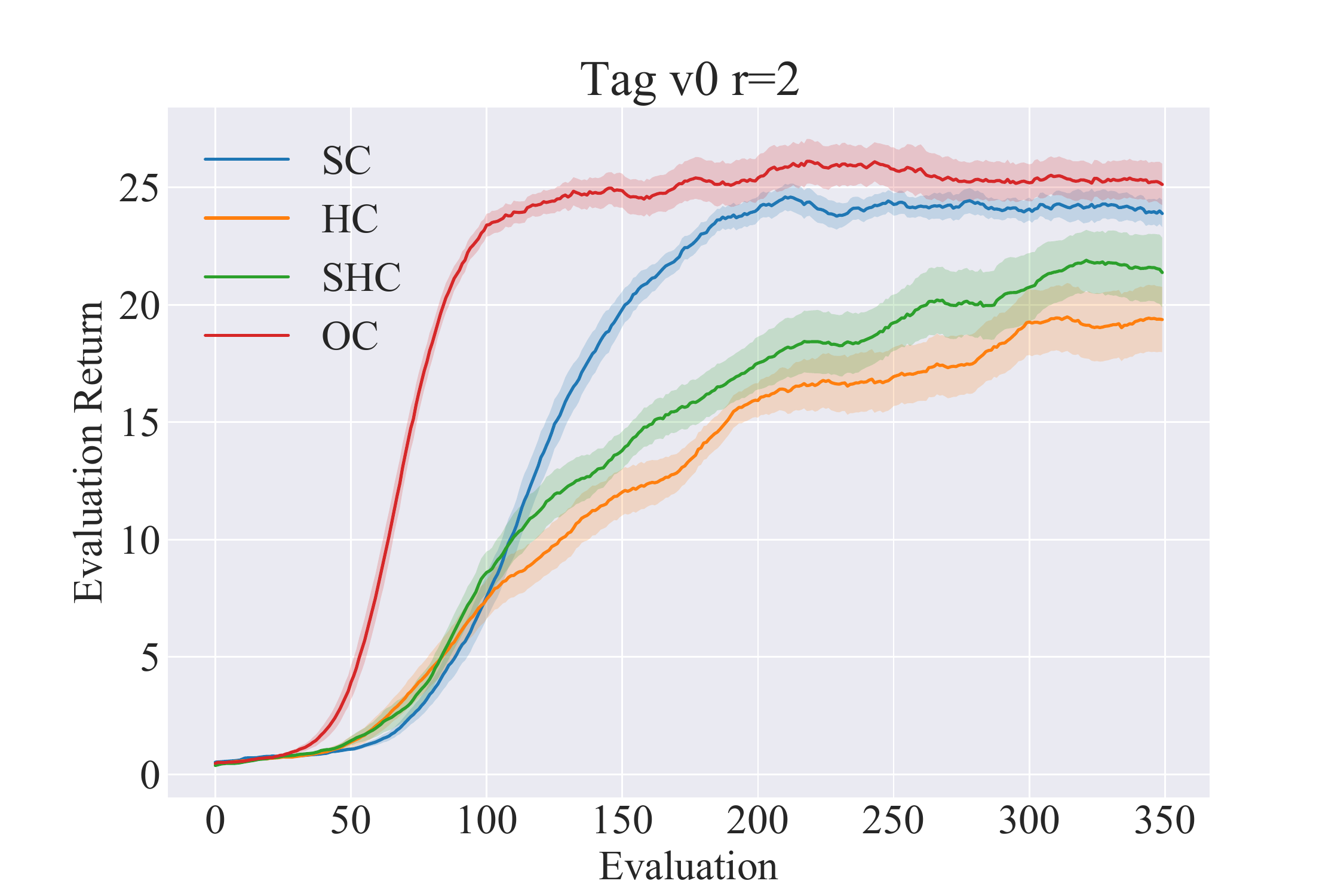}}
  \caption{MAAC Performance comparison in Cooperative Tag V0.}
  \label{fig:maax_tag_v0}
\end{figure}

\begin{figure}[h]
  \centering
  \captionsetup[subfigure]{labelformat=empty}
  \subcaptionbox{}
      [0.31\linewidth]{\includegraphics[height=3.4cm]{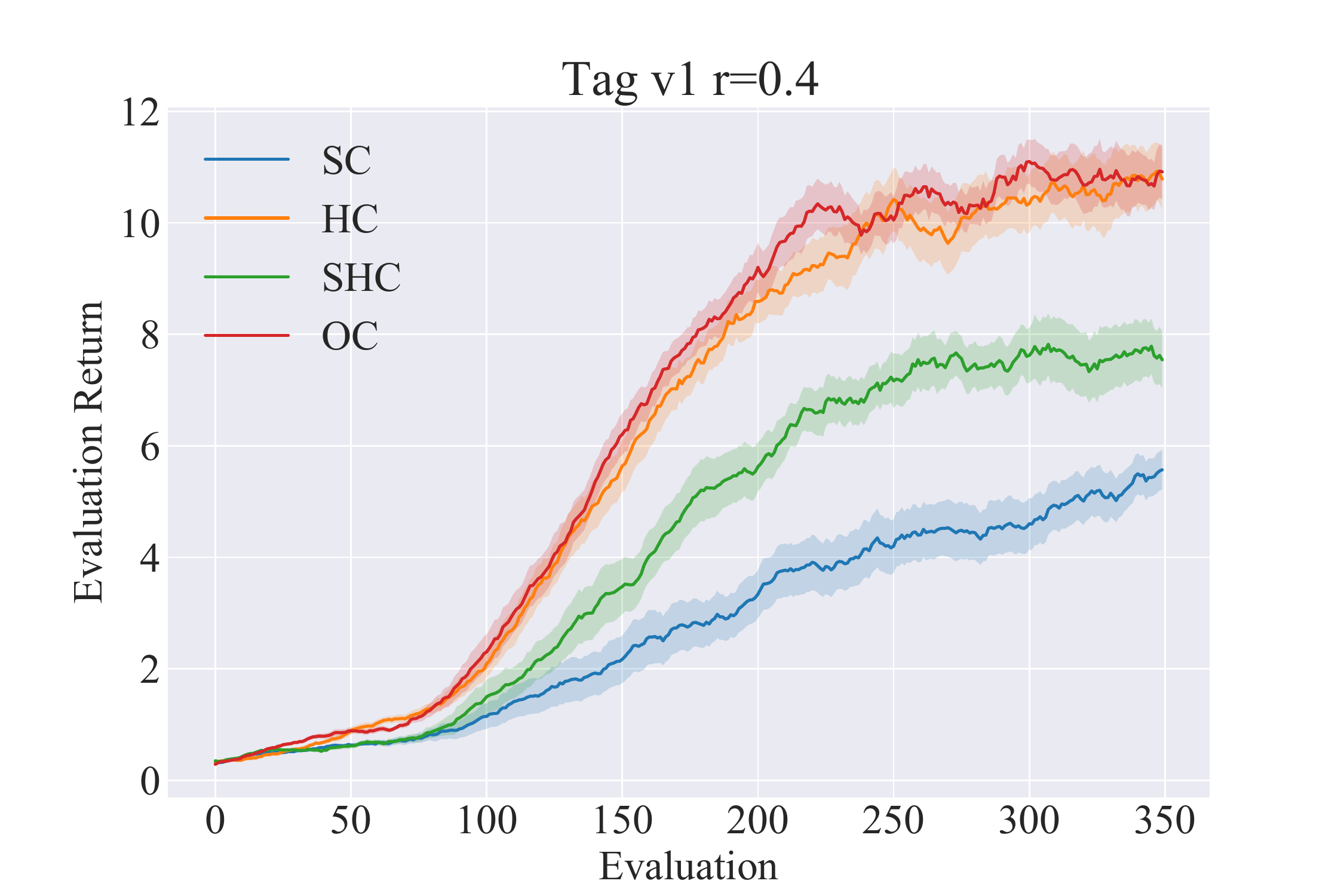}}
  ~
  \centering
  \subcaptionbox{}
      [0.31\linewidth]{\includegraphics[height=3.4cm]{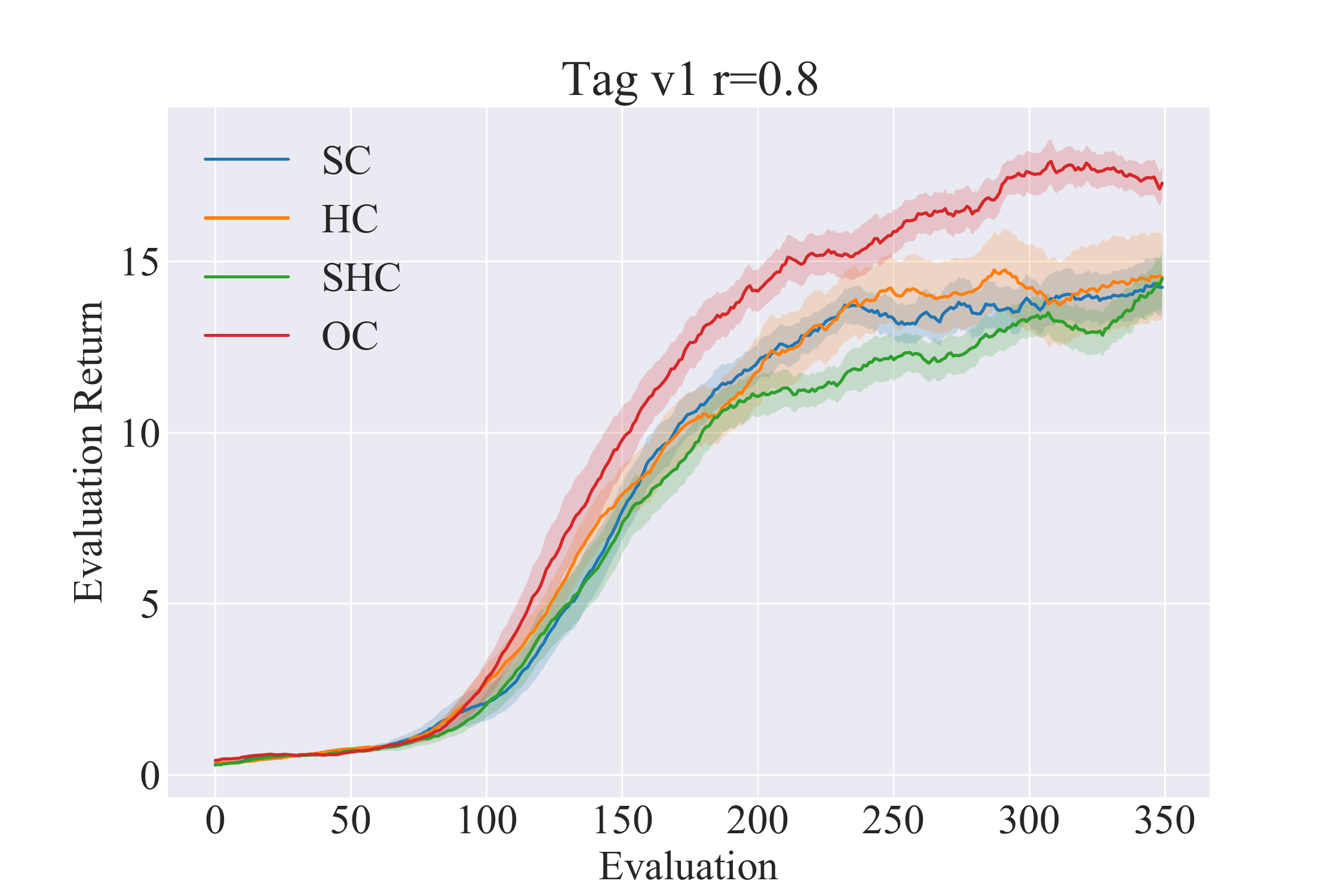}}
  ~
  \centering
  \subcaptionbox{}
      [0.31\linewidth]{\includegraphics[height=3.4cm]{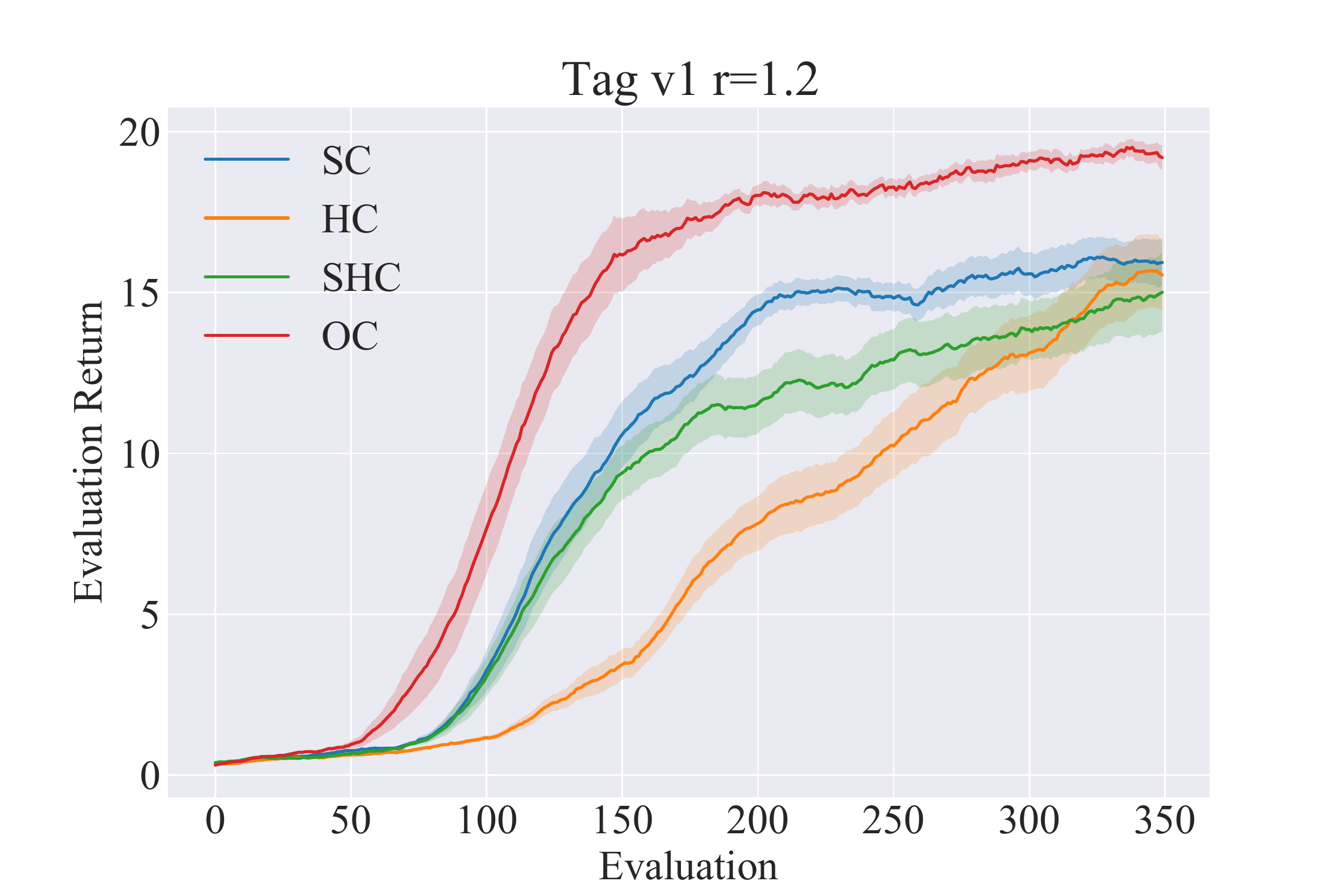}}
   
  \centering
  \subcaptionbox{}
      [0.31\linewidth]{\includegraphics[height=3.4cm]{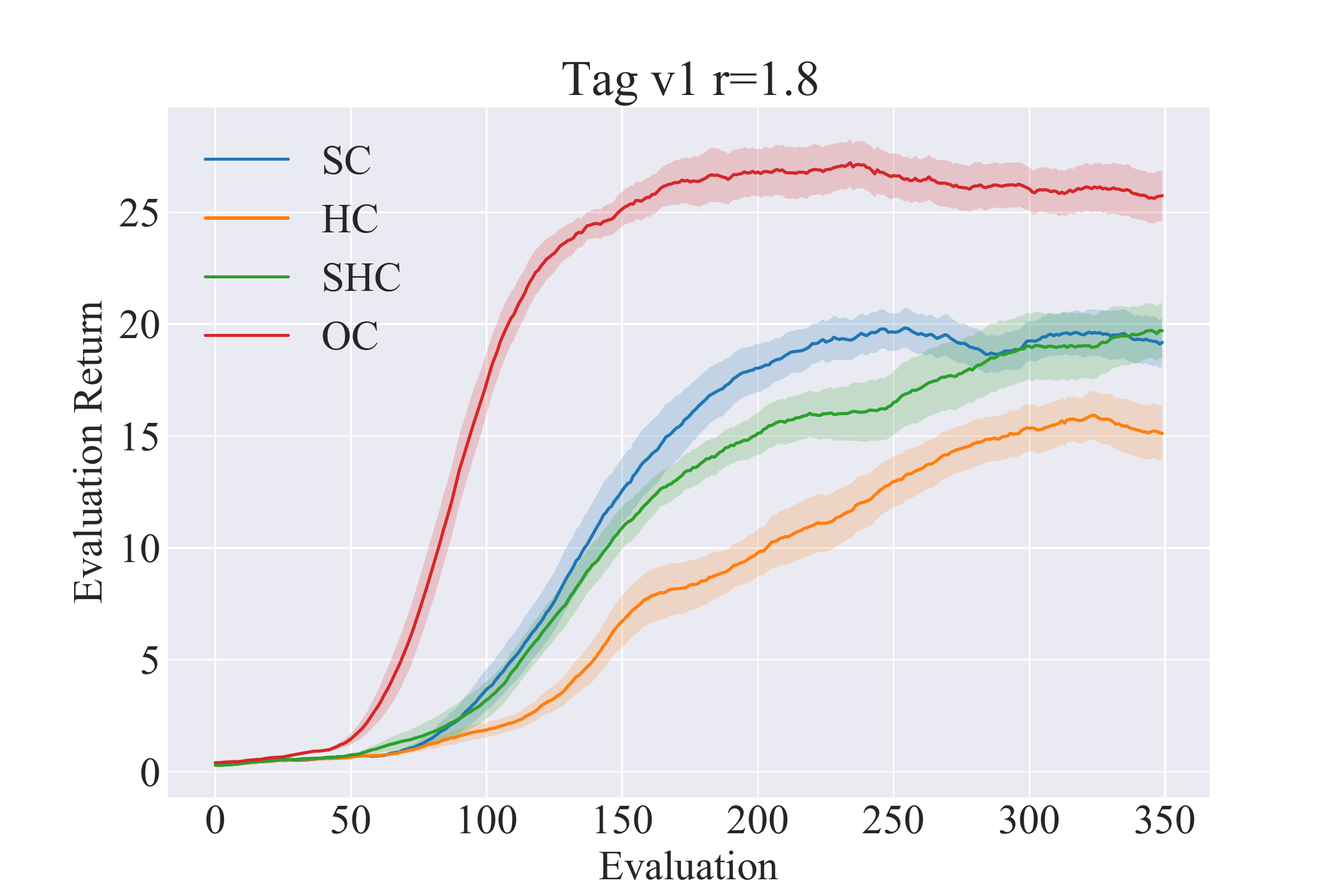}}
  ~
  \centering
  \subcaptionbox{}
      [0.31\linewidth]{\includegraphics[height=3.4cm]{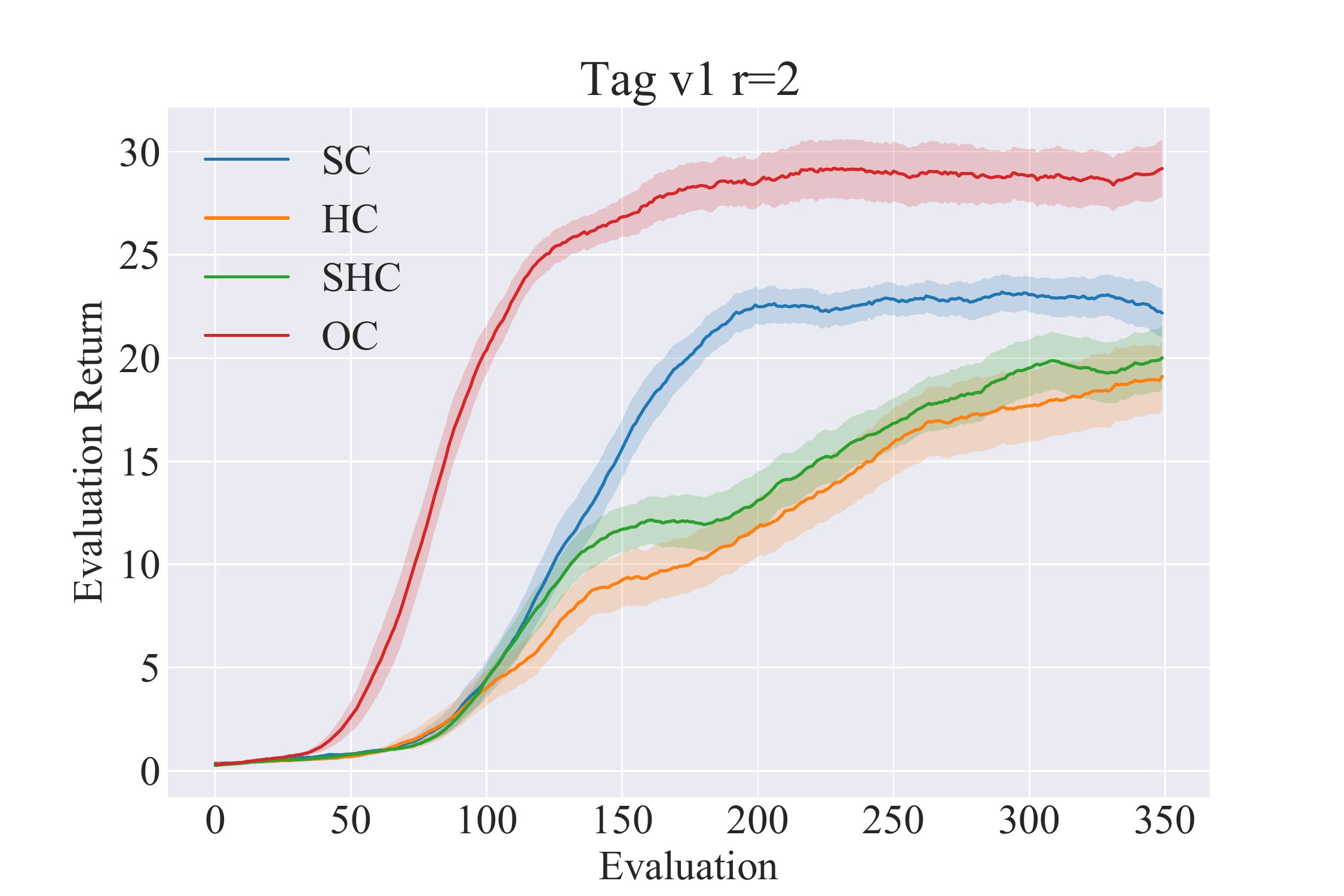}}
  \caption{MAAC Performance comparison in Cooperative Tag V1.}
  \label{fig:maax_tag_v1}
\end{figure}

\end{document}